\documentclass{article}

\usepackage{arxiv}

\usepackage[utf8]{inputenc} 
\usepackage[T1]{fontenc}    
\usepackage{hyperref}       
\usepackage{url}            
\usepackage{booktabs}       
\usepackage{amsfonts}       
\usepackage{nicefrac}       
\usepackage{microtype}      

\usepackage[english]{babel}
\usepackage{amsthm}
\usepackage{stackengine}[2013-10-15]
\newcommand\textss[1]{\stackengine{.9ex}{}{\scriptsize#1}{O}{l}{F}{F}{L}}
\usepackage{amsmath,amssymb}
\DeclareMathOperator*{\argmax}{argmax} 

\usepackage[thinlines]{easytable}
\usepackage{csquotes}

\usepackage{natbib}
 \bibpunct[, ]{(}{)}{,}{a}{}{,}%

\usepackage{graphicx}
\graphicspath{ {./images/} }

\usepackage{chngcntr}
\usepackage{apptools}
\AtAppendix{\counterwithin{lemma}{section}}
\AtAppendix{\counterwithin{proposition}{section}}
\AtAppendix{\counterwithin{corollary}{section}}

\newtheorem{proposition}{Proposition}

\newtheorem{lemma}{Lemma}

\title{Deep Q-Learning for Same-Day Delivery with Vehicles and Drones}

\author{
  Xinwei Chen \\
  Applied Mathematical and Computational Sciences\\
  University of Iowa\\
  Iowa City, United States \\
  \texttt{xinwei-chen-1@uiowa.edu} \\
   \And
 Marlin W. Ulmer \\
  Technische Universität Braunschweig\\
  Carl-Friedrich-Gauß-Fakultät\\
  Braunschweig, Germany \\
  \texttt{m.ulmer@tu-braunschweig.de} \\
   \AND
   Barrett W. Thomas \\
   Department of Business Analytics \\
   University of Iowa\\
   Iowa City, United States \\
   \texttt{barrett-thomas@uiowa.edu} \\
}

\date{}
\begin{document}

\maketitle
\begin{abstract}
In this paper, we consider same-day delivery with vehicles and drones. Customers make delivery requests over the course of the day, and the dispatcher dynamically dispatches vehicles and drones to deliver the goods to customers before their delivery deadline. Vehicles can deliver multiple packages in one route but travel relatively slowly due to the urban traffic. Drones travel faster, but they have limited capacity and require charging or battery swaps. To exploit the different strengths of the fleets, we propose a deep Q-learning approach. Our method learns the value of assigning a new customer to either drones or vehicles as well as the option to not offer service at all. In a systematic computational analysis, we show the superiority of our policy compared to benchmark policies and the effectiveness of our deep Q-learning approach. We also show that our policy can maintain effectiveness when the fleet size changes moderately. Experiments on data drawn from varied spatial/temporal distributions demonstrate that our trained policies can cope with changes in the input data.
\end{abstract}


\section{Introduction}\label{intro}

Same-day delivery (SDD) changes the way people shop as it combines immediate product availability and the convenience of ordering from electronic devices \citep{2014same}. Because of its attractive nature, SDD were already expected to reach 15\% of last-mile delivery volumes in 2020 \citep{joerss}, and retailers such as Amazon, Target, and Walmart were racing to expand their SDD options \citep{thomas2019}. With people seeking to limit social contact, the COVID-19 pandemic has accelerated demand \citep{peltz_morgan} causing even Amazon to struggle to fulfill orders \citep{bertoni2020}.

Yet, even before the surge of demand brought on by the COVID-19 pandemic, providing SDD was a challenge. The timing of requests and the delivery locations are not known until a customer places an order. Further, because of the need to meet tight delivery deadlines, consolidation opportunities are scarce, rendering the use of conventional delivery vehicles (hereafter referred to as ``vehicles'') inefficient, especially for delivery in less dense areas of a city. In addition, vehicles are slowed by congestion on urban streets. 

As an alternative, companies have begun to complement vehicles with Unmanned Aerial Vehicles (hereafter referred to as ``drones''). In 2016, Amazon Prime Air made its first drone delivery in Cambridgeshire, England \citep{kim2016}. In May 2019, Alphabet Inc., Google's parent company, announced its subsidiary Wing would launch drone deliveries in Finland starting in June 2019 \citep{pero}. Wing has also been approved for commercial drone deliveries in the United States and Australia. In September 2019, Wing announced it would begin to test drone deliveries in Christiansburg, Virginia, teaming up with FedEx and Walgreens to provide deliveries of customer packages and over-the-counter medicines as well as food and beverages \citep{bhagat_2019}. Because they provide an alternative way of delivery that supports social distancing for both customers and essential workers amid the COVID-19 pandemic, industry sees an emerging safety need for drones in last-mile delivery \citep{wolf}. In April 2020, UPS announced that it would dramatically expand their drone delivery program to deliver prescription medicines to the largest retirement community in the US \citep{shapiro}.

While drones offer some advantages over vehicles, they also have limitations. Existing drones can usually carry only one item at a time and require regular charging or battery swaps. As a result, drones may not entirely replace vehicles in last-mile delivery, especially when the volume of customer requests is high \citep{wang2016}. Further, recent studies show that there are benefits to combining fleets of vehicles and drones for SDD \citep{drones}. However, the challenge arises as to how to effectively exploit the strengths of the individual vehicle types (e.g. capacity, speed) to offer service to as many customers as possible per day.

In this paper, we address this challenge for the \textit{same-day delivery problem with vehicles and drones} (SDDPVD). We study a problem where, over the course of a day, vehicles and drones deliver goods from a depot to customers and then return to the depot for future dispatches. The vehicles and drones differ in their travel speeds, capacities, and the need for charging or battery swaps. Customers are unknown until they make requests, and each request is associated with a delivery deadline. For every request, the dispatcher must determine whether the request is accepted and, if so, whether a vehicle or drone will make the delivery. All accepted requests must be delivered within a certain period of time. The objective is to maximize the expected number of customers served.  

This problem is complex because each decision impacts the availability of drones and vehicles to serve future requests, and because customers are waiting for a response, decisions need to be made instantaneously. To overcome these challenges, we propose a deep Q-learning approach. Q-learning is a form of reinforcement learning that seeks to learn the value of state-action pairs. Deep Q-learning uses deep neural networks as approximation architecture. Because the neural networks can be trained offline, the method can be employed for real-time decision making. We compare the proposed approach to high-quality benchmark policies from the literature. We also develop new methods for this paper that seek to improve the quality of the benchmarks from the literature by adding new features to them. 

This research makes three important contributions to the literature:
\begin{enumerate}
\item We show that the proposed approach is an effective and fast approach to an important and emerging delivery problem. Our computational results demonstrate the proposed Q-learning approach provides higher-quality solutions than all of the benchmarks. 
\item Our computational study demonstrates that  the proposed solution method is robust to changes in the distributions on the timing and locations of customer requests as well as changes in the delivery fleet.
\item The paper is among the first papers to implement deep Q-learning techniques for SDD and for dynamic routing problems in general. Our work highlights the potential and challenges of reinforcement learning techniques in dynamic vehicle routing. Particularly, this paper shows the value of including features that reflect both resource utilization and the impact of action selection in the current state. The identification of these categories of features offers general guidance for feature selection in other dynamic routing problems. 
\end{enumerate}

The paper is organized as follows. In Section \ref{lit_review}, we present the literature related to SDD and reinforcement learning for routing-related problems. Section~\ref{description_model} describes the problem and presents a sequential decision process model of the problem. In Section~\ref{method}, we introduce the deep Q-learning solution approach for the SDDPVD and the selected features. Section~\ref{design} describes the details of instances, implementation, and the benchmark policies as well as presents the results of our computational study. Section~\ref{conclusions} closes the paper with conclusions and discussion on future work. 

\section{Literature Review}\label{lit_review}
In this section, we present the literature related to the SDDPVD. We first review the literature related to the SDD and then present the existing applications of reinforcement learning in vehicle routing problems (VRPs). For a general review of drone routing problems, we refer the reader to \cite{otto}. It is worth noting that, most of the papers cited by \cite{otto} do not consider the dynamism of the SDDPVD. In contrast to the work in this paper, most of the literature involving drone delivery assumes that the customers to be served are known a priori. \citet{boysen2020last} provide a review of the last-mile-delivery literature, including drone delivery, and the emerging methods for solving the challenge.  

\subsection{Same-day Delivery}\label{ssd_lit}
The existing related literature on the SSD is limited, but is increasing as the service becomes more popular. The most closely related work is found in \citet{drones}. Similar to this paper, \citet{drones} consider the SDD with a fleet of vehicles and drones. Their work is the first to consider a heterogeneous fleet in the context of SDD. It is also the first to investigate the impact of adding drones to a fleet of conventional vehicles for a dynamic routing problem. The authors introduce a parametric policy function approximation (PFA) approach to heuristically solve the assignment portion of the problem. They use a fixed travel time threshold to determine whether a customer should be served by a drone or vehicle. 

The PFA policy presented in \cite{drones} uses only the location of customers to determine whether to use a vehicle or drone to serve a given customer. Yet, there is considerably more available information that might help produce better decisions. The approach in this paper uses not only the location, but also other available information reflecting resources and demand. We demonstrate that this additional information greatly improves upon the performance of the PFA and other benchmarks. To facilitate the incorporation of this additional information, we turn to a different type of approximation, Q-learning---an approximation of the value of state-action pairs, which we implement using neural networks (NNs).

Also related is the work of \cite{liu} in which the author considers on-demand meal delivery using drones. In the paper, the dispatcher dynamically assigns drones of different capacities to pick up food from restaurants and deliver to customers. The author first introduces a mixed-integer-programming model to minimize the total lateness for the static version of the problem and then uses heuristics to solve the dynamic problem. The problem is similar to the SDDPVD in that they both consider the dynamism of customer orders and involve the routing with drones. However, in the dynamic case, \cite{liu} present a rolling-horizon approach, an approach that ignores the future when making current decisions. For the SDD, \citet{voccia} shows that accounting for the future leads to considerably better performance than a rolling-horizon approach. Thus, our approach uses deep Q-learning to incorporate each decision's current value as well as its value on the future. Further, in addition to an assignment (routing) decision, the dispatcher in the SDDPVD must also make an acceptance decision for each customer request, which makes the action space in the problem even larger than that in the problem studied by \cite{liu}. Finally, the SDDPVD requires the routing of vehicles, further complicating assignment decisions. 

\citet{grippa} also study a version of the SDD in which deliveries are made by a fleet of drones. The authors consider a system of drones that deliver goods from the depot to customers and model it as a queuing problem. Performance of the policies with different heuristics are evaluated computationally. Because of the drones' interaction with the vehicles and the possibility of consolidating multiple customers packages onto vehicles, the queueing approach proposed in \citet{grippa} does not apply to the problem in this work.

Other literature in SDD presents anticipatory methods for single-vehicle problems. \citet{klapp31} consider a dynamic routing problem of a vehicle traveling on a line, where the probabilistic information is used to anticipate future requests. \citet{klapp2018dynamic} use an a-priori-based rollout policy to determine the customers to serve and whether the vehicle should leave the depot for deliveries at the current time or wait for more customer orders coming in. \citet{klapp2020request} then consider making immediate acceptance decisions in SDD and use an a-priori-based rollout policy to minimize the sum of costs and penalties. \citet{ulmer45} consider a SDD problem in which the vehicle is allowed for preemptive returns to the depot, e.g., returning to the depot before completing the deliveries in the current route. The authors introduce an approximate dynamic programming (ADP) approach, where at each customer (or at the depot) a decision of whom the vehicle visits next is made using the information in the state. In contrast to \citet{klapp31,klapp2018dynamic,klapp2020request} and \citet{ulmer45}, this paper focuses on not only multiple vehicles but a fleet of both vehicles and drones. The methods proposed in \citet{klapp31,klapp2018dynamic,klapp2020request} and \citet{ulmer45} do not scale to the problem discussed in this paper.

\citet{azi} and \citet{voccia} consider using multiple, but homogeneous vehicles in dynamic routing problems. These papers solve the problems using multiple-scenario approaches (MSAs). While effective, MSA requires real-time computation that can be challenging in the context of the large fleet and the many incoming requests that we consider in this paper. Our proposed method uses offline computation to learn the approximation and can provide instantaneous solutions in real-time. \citet{dayarian2020crowdshipping} consider crowdshipping in SDD. In addition to the store's own fleet, in-store customers can also make deliveries for online orders on their way home. The problem is similar to ours in that it considers whether an order should be served by in-store customers or the store's own fleet. However, the authors propose two rolling horizon approaches while we introduce a deep Q-learning approach. 

\citet{dayarian} consider the SDD with drone resupply, where a vehicle performs the actual deliveries of goods to customers, and a drone resupplies goods from the depot to the vehicle en route. The use of the drone resupply enables the vehicle to serve a new set of customers without the need of returning to the depot to pick up the goods, and thus more customers can be served before their delivery deadlines. \citet{ulmer_station} consider the SDD with a fleet of autonomous vehicles, where the autonomous vehicles deliver the ordered goods from the depot to a set of pick-up stations. The authors introduce a PFA approach to minimize the expected sum of delivery times. The emerging business models studied in \citet{dayarian} and \citet{ulmer_station} complement the work in this paper.

\subsection{Reinforcement Learning for VRPs}\label{RL_VRP}
In this section, we focus only on the literature that uses reinforcement learning for VRPs. We do not consider supervised learning approaches such as are discussed in \citet{potvin}, \citet{vinyals2015}, and \citet{fagerlund2018}. Reinforcement learning (RL) is an area of machine learning that is often applied to sequential decision processes for problems in robotics, artificial intelligence, and signal control. \citet{sutton2018reinforcement} provide a general overview. There are two common types of RL algorithms: policy-based methods and value-based methods. Policy-based learning algorithms seek to optimize a policy directly. Value-based learning algorithms learn the value of being in particular states or of particular state-action pairs. We use a deep Q-learning approach, an approach that seeks to learn the value of state-action pairs. Deep Q-learning network was introduced by \citet{atari} who demonstrate its ability to play Atari games with super-human performance. 

There are a few papers presenting RL-work in dynamic vehicle routing, for example, \citet{toriello2014dynamic,perez2017,ulmer2017budgeting}, and \citet{joe2020}. For a recent overview, we refer the reader to \citet{meso}. The work in the literature usually draws on the concept of value-function approximation (VFA). VFAs approximate the value of post-decision states by means of simulations. The values are stored in approximation architectures, usually either functions or lookup tables. The main difference between VFAs and Q-learning is that the latter considers the value of state-decision pairs versus just the value of a post-decision state. Even if post-decision states are a summary of the result of taking a particular action in a particular state, we show that, for the SDDPVD at least, Q-learning's ability to take both state- and action-space information into account has an advantage over relying on just information available in the post-decision state. 

Work on using NNs in RL for VRPs is particularly scarce. \citet{madziuk} provides an overview of recent advances in the VRP literature over the last few years, where the author points out that ``it came as a surprise to the author that NNs have practically not been utilized in the VRP domain \ldots." 

Among the papers using RL with NNs applied to dynamic routing problems, the most closely related to this paper is \citet{chen2019}. \citet{chen2019} introduce an actor-critic framework, a policy-based RL method, for the problem of making pick-ups at customers who make dynamic requests for service. The problem is similar to the SDDPVD in that they both consider dynamic requests and customer locations are unknown in advance. The papers differ in that \citet{chen2019} learn a policy for a single vehicle and then apply this policy to all vehicles. As a result, the policy that is learned does not account for the interaction among the vehicles. Our results show that accounting for this interaction leads to superior solution quality. In addition, \citet{chen2019} do not make decisions on whether or not to offer service to customers, but rather they allow unserved customer requests to expire.



In their appendix, \citet{reza} discuss a problem in which a single vehicle serves dynamic requests. Requests not served in a specified time period are lost. The vehicle has a limited amount of inventory and must return to the depot to replenish once the inventory is depleted. Like \citet{chen2019}, \citet{reza} propose a policy-gradient approach. The problem studied by \citet{reza} is different and has a much smaller state and action space than the problem studied in this paper. Likewise, \citet{kool} use a policy-gradient approach to solve a stochastic prize collecting traveling salesman problem and also presents an application to a   variant in which customer locations and demands can change. Again, given the different problems, the commonality between \citet{kool} and this paper is the use of RL. Yet, the use of policy-gradient methods in \citet{chen2019}, \citet{reza}, and \citet{kool} suggests a future opportunity to explore the value of policy-gradient versus Q-learning approaches for dynamic vehicle routing. 

\section{Problem Definition}\label{description_model}

In this section, we present a formal definition of the SDDPVD, initially presented in \cite{drones}. We will first give a problem narrative and then model the problem as a sequential decision process.

\subsection{Problem Narrative}\label{prob_narrative}

In this problem, a fleet of vehicles and a fleet of drones deliver goods from a depot to customers requesting service over the course of the day. Vehicles and drones each have operating periods during the day, starting and ending their day at the depot. The operating periods are limited, for example, because of working-hour regulations or because the warehouse closes in the evening. Vehicles and drones differ in their characteristics, namely their capacity and their speed. Drones are faster but they can only carry one order at a time. Furthermore, drones require recharging after each trip. Vehicles are slower, but spatial capacity can be neglected \citep{amazon}. Both fleets require time for loading at the depot (independent of the number of parcels to load) and for drop off of parcels at each customer. For the vehicles, we assume preemptive depot returns are prohibited. Thus, the vehicles must finish one delivery tour and then start the next one.

Over the course of the day, customers request SDD service. The requests are unknown before they are revealed. 
Whenever a new customer requests delivery, the provider decides if the customer can be offered service, and if so, whether the delivery will be conducted by a vehicle or a drone. If no service is offered, the customer request is declined and leaves the system. If service is offered, it must be conducted before a deadline, a predefined time span after the request is placed. If the order is assigned to a vehicle, decisions are also made about how the planned routes of the vehicle fleet is adapted to integrate the new order. If the order is assigned to the drone fleet, the decision also indicates which drone will conduct the delivery. Due to different handling of parcels for drone and vehicle deliveries, we assume that the assignment to a fleet type is permanent. A decision is feasible if the planned routes ensure deliveries are made before their corresponding deadline, and vehicles and drones return to the depot before the end of their operating periods.

The objective of the SDDPVD is to maximize the expected number of deliveries per day.

\subsection{Preparation of the Model}\label{prep_model}

The problem presented in this paper is a sequential decision problem, and in modelling the problem, we follow the framework presented in \citet{POWELL2019795}. We also use \textit{route plans} to model the updates and evolution of routes. Route plans are described in detail in \cite{ULMER2020100008}. Before we define the components of the sequential decision problem, we will provide some general notation as well as the concepts related to route plans. We will then give an example for a state and a decision.

\subsubsection*{General Notation}

We denote the fleet of $m$ vehicles $\mathcal{V}=\{v_1,v_2,\dots,v_m\}$ and the fleet of $n$ drones $\mathcal{D}=\{d_1,d_2,\dots,d_n\}$. Delivery takes place in a service area $\mathcal{A}$, with the depot $\mathcal{N}$ being part of the area. The set of customers in the system are denoted by $\mathcal{C}$ and individual customers by $C_1,C_2,\dots$. The time of request for each customer $C_i$ is denoted by $t(C_i)$.

The travel times between two points in the area are determined by function $\tau_V(\cdot , \cdot)$ for vehicles and for a drone by $\tau_D(\cdot , \cdot)$. It takes $t_N^D$ units of time to load a package onto a drone and $t_C^D$ units of time for a drone to drop off a package at a customer. The charging time after each delivery trip for a drone is $t^D_B$. For vehicles, the loading time is denoted $t_N^V$ and the drop off time at a customer is $t_C^V$. Vehicles must return to the depot before $t^V_{\max}$ and drones before $t^D_{\max}$. Thus, the overall operation period is $[0,\max\{t^D_{\max},t^V_{\max}\}]$. We denote the delivery deadline of customer $C_i$ as $\delta(C_i)$. The delivery deadline per customer $C_i$ is $\bar{\delta}$ units of time after the request is placed, and thus, $\delta(C_i)=t(C_i)+\bar{\delta}$.

\subsubsection*{Route Plans}

Our model operates on route plans. A route plan represents the tentative routes vehicles and drones follow until the next decision update takes place (at the time of a new request). A planned route of a vehicle or drone contains information about the set of planned stops and the corresponding arrival times. For depot stops, it also contains the planned departure times of subsequent routes. Over the operation period, each decision leads to an update of the planned routes for vehicles and drones. We denote the set of planned routes for the vehicles by $\Theta_\mathcal{V}$ and the set of planned routes for the drones by $\Theta_\mathcal{D}$. Then, the set of all planned routes $\Theta$ is denoted as $\Theta=(\Theta_\mathcal{V},\Theta_\mathcal{D})$. 

The set of vehicle routes $\Theta_\mathcal{V}=(\theta(v_1),\dots,\theta(v_m))$ contains a route for each vehicle. A route is a sequence of stops with locations and information about the arrival (and departure) times. Because preemptive returns for vehicles are not allowed, the planned route for each vehicle contains information about only the end of its current delivery tour and the planned subsequent tour. We also note that the information per stop differs depending on whether it is a customer or a depot stop. Because waiting at the depot is possible, a planned route not only contains the planned time of arrival but also the time of departure from the depot. Mathematically, a route for a vehicle $v$ is represented by
$$\theta(v)=((N_1^\theta,a(N_1^\theta),  s(N_1^\theta)),(C_1^\theta,a(C_1^\theta)),\dots,(C_h^\theta,a(C_h^\theta)),(N_2^\theta,a(N_2^\theta),  t^V_{\max})).$$

The first entry of a planned route $\theta(v)$ represents vehicle $v$'s next depot visit $N_1^\theta$, the arrival time at the depot $a(N_1^\theta)$, and the time $s(N_1^\theta)$ at which the vehicle starts to load packages for the assigned customers in the next tour. Therefore, the difference between $a(N_1^\theta)$ and $s(N_1^\theta)$ is the time the vehicle spends waiting at the depot before it starts its next delivery tour. Following the first depot visit is a sequence of customers $C_i^\theta$, $i=1,2,\dots,h$, that are assigned to vehicle $v$ but not yet loaded. The customer stops are accompanied with arrival time information $a(C_i^\theta)$ for each customer $C_i^\theta$. The last entry in a planned route represents the vehicle's return to the depot at time $a(N_2^\theta)$, implying the vehicle will wait until $t^V_{\max}$.

A planned route for a drone is slightly different from that for a vehicle in that a customer entry must be between two depot entries due to the capacity of drones. For a drone $d$, its planned route $\theta(d)$ is represented by 
\begin{eqnarray}
\theta(d) = ((N_1^\theta,a(N_1^\theta),  s(N_1^\theta)),(C_1^\theta,a(C_1^\theta)),(N_2^\theta,a(N_2^\theta),  s(N_2^\theta)),\nonumber\dots,(C_{-2}^\theta,a(C_{-2}^\theta)),(N_{-1}^\theta,a(N_{-1}^\theta),  t^D_{\max}). \nonumber
\end{eqnarray}
Because there can be more than two depot entries in a drone's planned route, we use index $-1$ to represent the last entry, $-2$ the second to last and so on. The difference between $a(N_1^\theta)$ and $s(N_1^\theta)$ represents drone $d$'s charging time after the previous delivery and the time the drone spends on waiting at the depot before it starts to load the items for its next delivery tour. Finally, the waiting time after the last planned delivery is set to $t^D_{\max}$.

\subsection*{Illustrative Example}\label{illustrative_example}

\begin{figure}[h]
	\centering
	\includegraphics[width=0.7\textwidth]{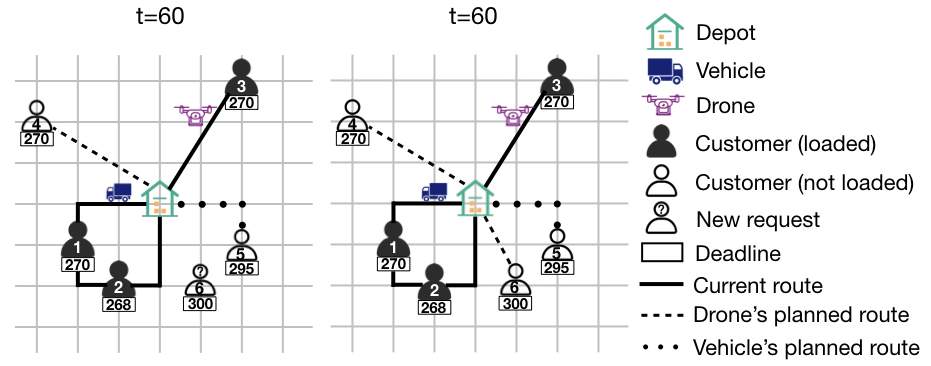}
	\caption{Decision and decision state.}
	\label{illustrative}
\end{figure}

In Figure \ref{illustrative}, we illustrate the SDDPVD for time $t=60$ (in minutes), an hour after the shift begins. At this time, we assume that the dispatcher receives a new customer request $C_6$. The two panels in the figure describe the corresponding $pre-$ and $post-decision$ states, which are formally described in Section~\ref{markov}. 
In this example, the depot is located in the center of the area. The fleet consists of a vehicle and a drone. We assume the vehicle travels on a Manhattan-style grid, and the drone travels in the Euclidean plane. The vehicle needs 20 minutes to travel a segment. The drone travels in the Euclidean plane with twice the speed of the vehicle. All travel times are rounded up to the nearest integral minute. Customer orders must be completed within $240$ minutes after acceptance. For both the vehicle and the drone, a loading time at the depot and a delivery time at a customer are 10 minutes. The charging time for the drone is 20 minutes. 

The panel on the left shows the status of the vehicle and drone before the dispatcher makes the acceptance and assignment decisions regarding $C_6$. The vehicle is currently en route serving $C_1$ and then $C_2$. Its planned route is $\theta(v)=((N,220,220),(C_5,290),(N,360,t^V_{\max}))$. It will arrive at $C_1$ at $t=60+40=100$, and then arrive at $C_2$ at $t=100+10+40=150$. The vehicle then will leave $C_2$ at $t=150+10=160$ and return to the depot at $t=160+60=220$. Thus, the arrival time of the first depot entry in the planned route is 220. The vehicle is planned to serve $C_5$ in its next route. Customer $C_5$ is accepted but not yet loaded because the vehicle has not returned to the depot to load the package for it. The vehicle plans to arrive at $C_5$ at $t=220+10+60=290$ and then return to the depot at $t=290+10+60=360$.  The drone is currently en route serving $C_3$, with the planned route $\theta(d)=((N,124,144),(C_4,191),(N,238,t^D_{\max}))$. The drone will arrive at $C_3$ at $t=40+10\sqrt{13}\approx77$, and return to the depot at $t=77+10+10\sqrt{13}\approx124$. Due to the charging time, the drone will not load the new package until $t=124+20=144$. It is then planned to leave the depot for $C_4$ and arrive at $C_4$ at $t=144+10+10\sqrt{13}\approx191$. The drone will return to the depot at $t=191+10+10\sqrt{13}\approx238$.

As the new customer $C_6$ makes a request, the dispatcher determines the feasibility of assigning the new customer to the vehicle and the drone. The delivery deadline of $C_6$ is $60+240=300$. If the vehicle serves $C_6$ right after $C_5$, then it will arrive at $C_6$ at $t=290+10+40=340$, which is not feasible because it passes the deadline of $C_6$. Similarly, it is not feasible either to serve $C_6$ first and then $C_5$. Overall, it is not feasible to serve $C_6$ by the vehicle. Applying the same analysis yields that it is feasible for the drone to serve $C_4$, return to the depot, and then serve $C_6$. Thus, it is feasible to serve $C_6$ with the drone because there exists a feasible route.


The dispatcher next makes the decision. Let us assume the dispatcher accepts the request $C_6$ and assigns it to the drone. Then, the update is shown in the panel on the right in Figure \ref{illustrative}. The vehicle's planned route remains the same, and the drone's planned route becomes
$$\theta(d)=((N,124, 144),(C_4,191),(N,238,258),(C_6,291),(N,324, t^D_{\max})).$$

\subsection{Sequential Decision Process Model}\label{markov}

In this section, we model the SDDPVD as a sequential decision process, a sequence of states connected by actions and transitions. Due to similarities in the problems, this model is similar to that found in \cite{drones}.

\subsubsection*{Decision Point} A decision point is a time at which a decision is made. In the SDDPVD, a decision point occurs when a customer requests service. We denote the $k^{\text{th}}$ requesting customer as $C_k$ and the time of the $k^{\text{th}}$ decision point as $t_k=t(C_k)$. 

\subsubsection*{State}
The state at a decision point summarizes the information needed to make the decision. In the SDDPVD, the state $S_k$ at decision point $t_k$ includes the information about the customers, the vehicles and drones as well as the planned routes. Because the planned routes include the relevant information for the fleets, the fleet information can be omitted from the state information.  Mathematically, a state has four components:

\begin{itemize}
    \item $t_k$: Time of the decision point, $t_k=t(C_k)$.
    \item $C_k$: New request.
    \item $\mathcal{C}_k=(\mathcal{C}_{\mathcal{V},k}^{\theta},\mathcal{C}_{\mathcal{D},k}^{\theta})$: Set of customers still to be loaded and served as well as their deadlines. The set is split between customers currently planned for service with vehicles $\mathcal{C}_{\mathcal{V},k}^{\theta}$ and with drones $\mathcal{C}_{\mathcal{D},k}^{\theta}$. 
    \item $\Theta_k=(\Theta_{\mathcal{V},k},\Theta_{\mathcal{D},k})$: Set of planned routes for vehicles $\Theta_{\mathcal{V},k}$ and drones $\Theta_{\mathcal{D},k}$ when the request $C_k$ is received.
    \end{itemize}

In summary, we represent the state $S_k$ as a tuple $S_k=(t_k,C_k, \mathcal{C}_k,\Theta_k)$. Note, in the initial state $t=0$, no customer requests are pre-assigned, so all planned routes contain only the depot stop. In this case, $N_1^\theta$ represents a vehicle's (or drone's) initial position at the depot, and $a(N_1^\theta)$ is $0$ because every vehicle (drone) is available once the shift begins.

\subsubsection*{Action and Reward}

At every decision point, we determine an action $x_k$. 
In the following, we present the full action space of the SDDPVD, noting that our solution methods will use heuristic measures to reduce the action space.

In the SDDPVD, an action incorporates whether the request is accepted and, if so, which vehicle or drone will provide the service. We represent the action at a decision point $t_k$ as a tuple $x_k=(\alpha_k,\Theta_{\mathcal{V},k}^x,\Theta_{\mathcal{D},k}^x)$, where $\alpha_k$ is the acceptance and assignment decision, and $\Theta_{\mathcal{V},k}^x$ ($\Theta_{\mathcal{D},k}^x$) is the updated set of vehicle (drone) planned routes. In addition, $\mathcal{C}^x_k=(\mathcal{C}_{\mathcal{V},k}^{\theta,x},\ \mathcal{C}_{\mathcal{D},k}^{\theta,x})$ represents the updated set of customers planned to be serviced by vehicles $\mathcal{C}_{\mathcal{V},k}^{\theta,x}$ and drones $\mathcal{C}_{\mathcal{D},k}^{\theta,x}$ but not yet loaded. 

The first part of the action is the acceptance and assignment decision $\alpha_k$, which is defined as: 
$$ \alpha_k=\left\{
\begin{array}{rcl}
0       &      & {\textrm{if order is not offered service,}}\\
1     &      & {\textrm{if order is assigned to a vehicle,}}\\
2     &      & {\textrm{if order is assigned to a drone.}}
\end{array} \right. $$

The second and third parts of an action $x_k$ are the corresponding updates of the route plans. An update $\Theta_{\mathcal{V},k}^x$ for the vehicle fleet is feasible if it satisfies the following six conditions:

\begin{enumerate}
    \item The route plan in $\Theta_{\mathcal{V},k}^x$ contains all the customers in $\mathcal{C}_{\mathcal{V},k}^\theta$. It also contains $C_k$, in the case $\alpha_k=1$.
   \item For every customer $C\in\mathcal{C}_{\mathcal{V},k}^{\theta,x}$, the planned arrival time is not later than the deadline, $a(C)\leq\delta(C)$.
   \item In each planned route $\theta(v)\in\Theta_{\mathcal{V},k}^x$, the start of loading at the depot $s(N_1^\theta)$ is not earlier than the arrival time $a(N_1^\theta)$.
   \item In each planned route, the difference between the beginning of loading for the next tour $s(N_1^\theta)$ at the depot and the arrival time at the next customer is the sum of travel time and loading time.
   \item In each planned route, the difference between the arrival times of two consecutive customers is equal to the sum of travel time and service time.
   \item The vehicles must return to the depot not later than the end of the shift, $a(N_2^\theta)\leq t^V_{\max}$.
\end{enumerate}

Similarly, a drone update $\Theta_{\mathcal{D},k}^x$ is feasible if it satisfies the following five conditions:

\begin{enumerate}
    \item The route plan in $\Theta_{\mathcal{D},k}^x$ contains all the customers in $\mathcal{C}_{\mathcal{D},k}^\theta$. It also contains $C_k$, in the case $\alpha_k=2$.
    \item For every customer $C\in\mathcal{C}_{\mathcal{D},k}^{\theta,x}$, the planned arrival time is not later than the deadline, $a(C)\leq\delta(C)$.
    \item In each planned route $\theta(d)\in\Theta_{\mathcal{D},k}^x$, the start of loading at the depot $s(N_1^\theta)$ is not earlier than the arrival time $a(N_1^\theta)$ plus the charging time $t^D_B$.
    \item In each planned route, every customer entry must be in between of two depot visits.
    \item In each planned route, the arrival time of the last depot visit is not later than the end of the shift, $a(N_{\textrm{-1}}^\theta)\leq t^D_{\max}$
\end{enumerate}

We note that the action $x_k=(0,\Theta_{\mathcal{V},k},\Theta_{\mathcal{D},k})$ is always feasible. The reward of an action $x_k$ is 
$$ R(S_k,x_k)=\left\{
\begin{array}{rcl}
0      &      & {\textrm{if } \alpha_k=0,}\\
1    &      & {\textrm{otherwise.}}
\end{array} \right. $$

\subsubsection*{Post-Decision State}

The combination of a state $S_k$ and an action $x_k$ leads to a post-decision state $S^x_k$.
The post-decision state contains the following information:
\begin{itemize}
    \item $t_k$: Point of time.
     \item $\mathcal{C}^x_k=(\mathcal{C}_{\mathcal{V},k}^{\theta,x}, \mathcal{C}_{\mathcal{D},k}^{\theta,x})$: Set of customers still to be loaded and served as well as their deadlines. The sets $\mathcal{C}_{\mathcal{V},k}^{\theta,x}$ and  $\mathcal{C}_{\mathcal{D},k}^{\theta,x}$ are updated dependent on $\alpha_k$.
    \item $\Theta^x_k=(\Theta_{\mathcal{V},k}^x,\Theta_{\mathcal{D},k}^x)$: Updated set of planned routes for vehicles and drones.
\end{itemize}

\subsubsection*{Stochastic Information and Transition}

Transitions occur as a result of the realization of random customer requests. These requests are exogenous to the decision-making process. The realized information $\omega_{k+1}$ either reveals a new request and the time of the request $\omega_{k+1}=\{C_{k+1}, t(C_{k+1})\}$ or leads to termination of the process because the operation periods ended, $\omega_{k+1}=\{\}$. In the latter case, the process terminates in the final state, $S_{k+1}=S_K$.

Otherwise, the new state $S_{k+1}$ is a combination of post-decision state $S^x_k$ and stochastic information $\omega_{k+1}$ with the following updates:

\begin{itemize}
    \item The time of the decision point $t_{k+1}$ is $t(C_{k+1})$, the time of the new request  $C_{k+1}$.
\item The vehicles and drones may have started new delivery tours in the time span between $t_k$ and $t_{k+1}$. That is, there exists some $s(N_1^\theta)<t_{k+1}$, indicating a set of customers were loaded before $t_{k+1}$. We denote the set of such customers by $\mathcal{C}^x_{k,k+1}$. These customers are no longer part of the state, and therefore, $\mathcal{C}_{k+1}=\mathcal{C}^x_k \backslash \mathcal{C}^x_{k,k+1}$. The $\mathcal{C}_{\mathcal{V},k}^{\theta,x}$ and  $\mathcal{C}_{\mathcal{D},k}^{\theta,x}$ are also updated in the same way.
\item Similarly, the routes $\Theta_{k+1}$ are updated by removing all delivery trips that already started before $t_{k+1}$.
\end{itemize}


\subsubsection*{Objective Function}

A solution to the SDDPVD is a policy $\pi\in\Pi$ that assigns an action to each state. The optimal solution is a policy $\pi^*$ that maximizes the total expected reward and can be expressed by
\begin{equation}\label{obj}
\pi^*= \argmax_{\pi \in \Pi} \ \mathbb{E} \bigg[\sum_{k=0}^{K} R(S_k,X_k^\pi(S_k))|S_0\bigg], \nonumber
\end{equation}
where $X_k^\pi(S_k)$ is the action selected by policy $\pi$ at state $S_k$.

\section{Solution Approach}\label{method}


In this section, we present our solution approach. It is well known that the solution to an instance of a sequential decision process model can theoretically be found using backward induction applied to the Bellman equation:
\begin{equation}\label{eq:bellman}
V(S_k)=\max_{x\in X(S_k)}\{R(S_k,x)+\mathbb{E}[V(S_{k+1})|S_k]\},
\end{equation}
where $X(S_k)$ is the set of actions available at state $S_k$.

As with many other dynamic routing problems, however, the problem studied in this paper incurs the ``curses of dimensionality." For one, the state space of the SDDPVD is too large to even enumerate for realistically-sized problem instances. In addition, we also encounter a very large action space. Notably, the dispatcher must not only determine whether to offer service, but also to which fleet to assign accepted customers and how to route any customers assigned to vehicles. Thus, we propose a heuristic. In the following, we give a conceptual overview and an outline of our heuristic. We then describe the components in detail.

\subsection{Motivation}

For the SDDPVD, decisions not only need to be effective but fast. They should be effective by accounting for immediate and expected future revenues. Meanwhile, they should be made fast as well because customers expect immediate feedback about their service requests. To satisfy both requirements, we draw on methods of reinforcement learning \citep{sutton2018reinforcement}. The idea is to learn the value of a state and action by means of simulation. This simulation is done \enquote{offline} in a learning phase. The learned values can then be accessed within the \enquote{online} execution without any additional calculation time required. Because the size of the action space would prohibit even offline learning, we also draw on a runtime-efficient routing heuristic, reducing the action space to $\hat{X}$. In $\hat{X}$, actions are concerned with whether offering service to a customer and, if so, what fleet providing the service. The routing and assignment heuristic determines which individual vehicle or drone to make the delivery.

Learning values is challenging because of the enormous sizes of state and action spaces. For the SDDPVD, multiple vehicles and drones are routed to serve many customers. Storing the value for every potential state and action combination not only leads to substantial memory consumption, it also makes frequent value observations and therefore learning impossible. Thus, in addition to the action space, we also reduce the state space to a set of selected features and approximate the value for each feature vector by means of deep Q-learning. 

\begin{figure}[h]
	\centering
	\includegraphics[width=0.7\textwidth]{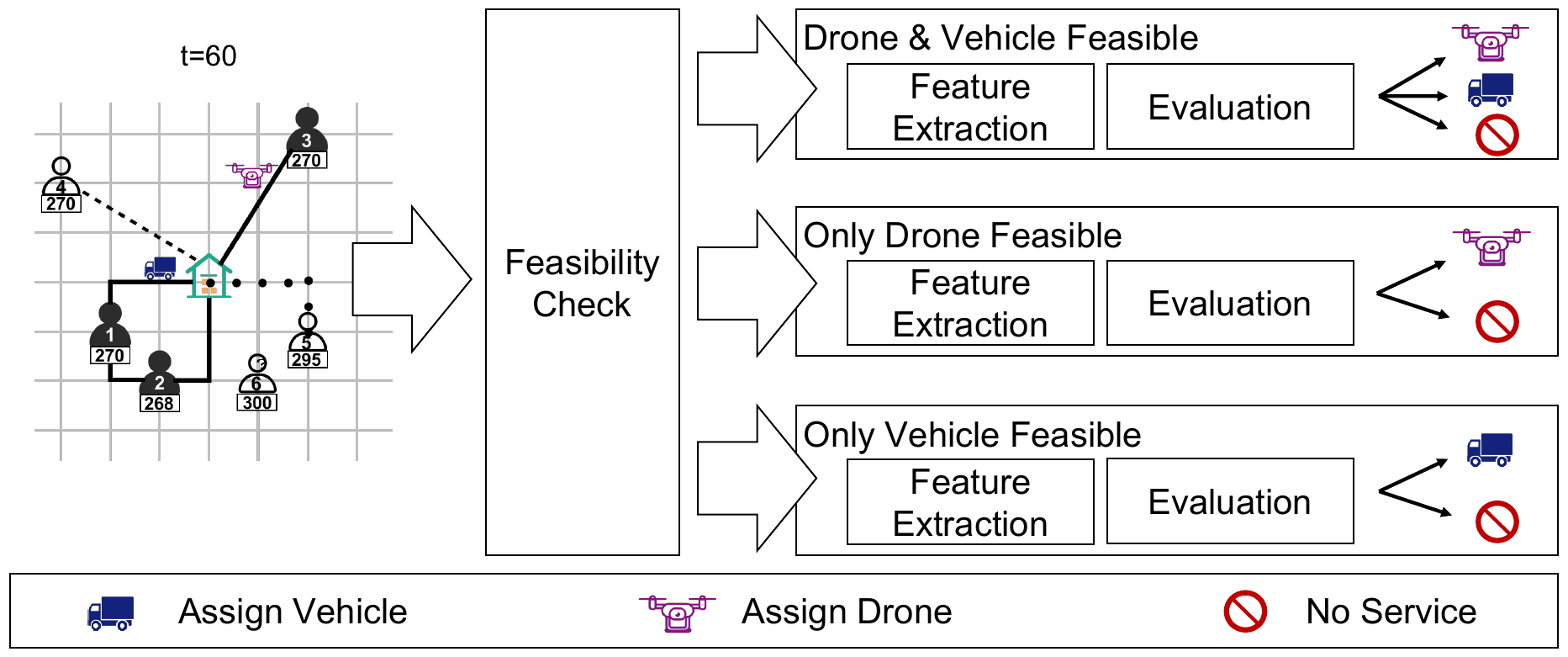}
	\caption{Conceptual Process of Decision Making.}
	\label{concept}
\end{figure}

Figure~\ref{concept} presents a flow chart of the process. In a pre-decision state, we first check feasibility for vehicles and drones by means of the routing heuristic. If serving the customer is infeasible, no service is offered. Otherwise, dependent on the drone and vehicle feasibility, we extract state features and evaluate the state and corresponding routing decision provided by the heuristic. Based on the evaluation, we assign the customer to a drone or vehicle, or we do not offer service to the customer at all.

Two main challenges arise from the process: what features to extract and how to evaluate the value of a specific state-decision pair. For the first, we will present a selection of features in Section~\ref{feature}. For the latter, we draw on Q-learning as discussed in the following. We denote our policy $\pi\textsuperscript{Q}$.

\subsection{Deep Q-learning}\label{deepql}


In this section, we introduce the deep Q-learning (DQL) solution approach and structure of the NNs used for the SDDPVD. Q-learning is a reinforcement learning approach that learns the value of taking an action in a given state. Thus, Q-learning learns a value $Q(S_k, x)$ for each state $S_k$ and action $x \in X(S_k)$, and this value is an approximation of the expected future reward when making action $x$ in state $S_k$. Given that we operate on the restricted action space $\hat{X}$, we learn a value for each state $S_k$ and action $x \in \hat{X}(S_k)$. With these Q-values and the reduced action space, we can solve an approximation of Equation~\ref{eq:bellman} written as
\begin{equation}\label{eq:qBellman}
\hat{V}(S_k)=\max_{x\in \hat{X}(S_k)}\{Q(S_k, x)\}.
\end{equation}

Depending on the existence of feasible route plans for vehicles and drones, we have a potential action set for each of the three cases when the decision is not trivially \textit{No Service} (see Figure \ref{concept}). When both vehicles and drones can feasibly provide the service, the set contains three actions \textit{Drone}, \textit{Vehicle} and \textit{No Service}. When only drones or only vehicles can feasibly do so, the set contains two actions \textit{Drone/Vehicle} and \textit{No Service}. For each of the potential sets, we create a NN to approximate the value of the corresponding decisions. This set of NNs are denoted as $\Phi=\{\phi^j, j=1,2,3\}$, where each $\phi^j$ represents the set of weights or parameters in the corresponding NN. They are indexed in the same order as the cases are presented in Figure \ref{concept} from top to bottom. In the following, we present the structure of the NNs and the procedure to train them.

\subsubsection*{Setup}
A NN is characterized by its input layer, hidden layers, and output layer. In our approach, each of the three NNs has the same structure:
\begin{itemize}
\item Input Layer. The input layer receives the extracted features and passes them to the hidden layers. Each NN uses the same features, and we present the features in Section \ref{feature}.

\item Hidden Layers. Each NN has $n$\textsubscript{hidden} hidden layers and $n$\textsubscript{nodes} nodes in each hidden layer. We use the rectified linear unit function (ReLU) as the activation function for each hidden layer.

\item Output Layer. The output layer of the NNs in the SDDPVD outputs the Q-values for each possible action $\hat{X}(S_k)$ for a state $S_k$. Specifically, the possible actions are (\textit{Drone}, \textit{Vehicle}, \textit{No Service}) for $\phi^1$, (\textit{Drone}, \textit{No Service}) for $\phi^2$, and (\textit{Vehicle}, \textit{No Service}) for $\phi^3$. Because the NNs approximate real-valued future rewards, there is no activation function on the output layer. 
\end{itemize}

To determine the best model for the NNs, we test different numbers of hidden layers for each NN, $n\textsubscript{hidden}\in\{2,5,10\}$, and different numbers of hidden nodes, $n\textsubscript{nodes}\in\{20,50,10\cdot|\mathcal{V}|\}$ with $|\mathcal{V}|$ the number of vehicles for each layer. The computational results show the combination of $n\textsubscript{hidden}=2$ and $n\textsubscript{nodes}=10\cdot|\mathcal{V}|$ outperforms the others.

\subsubsection*{Training}
We learn the parameters of the NNs in a training phase. The training set consists of $500$ instances. An instance is a set of customer requests made during a day, including their request times and locations. Training is performed by sampling an instance with replacement from the training set and then simulating on the sampled instance. We refer to each simulated day as a sample path.
During training, we usually select the action that maximizes the total expected reward associated with a state. This is known as exploitation. However, it is well known that occasionally randomly selecting an action improves the quality of the approximation \citep[Ch. 12]{powell2011approximate}. These random selections are known as exploration. We set the probability of exploring $\epsilon$ and exploiting $1-\epsilon$, where $\epsilon$ decays from $1$ to $0.01$ over the training steps. Each sample path is a training step of the NNs. Thus, we update the NNs after each sample path with a batch of the new observations and previous observations, a practice known \textit{experience replay} \citep{lin}. The goal of experience replay is to overcome the correlations between successive states in a sample path and the similarities between different paths. We use the mean-squared error as the loss function and minimize it using the well known $\textit{Adam}$ optimizer \citep{adam}, a stochastic gradient-based algorithm. As a result of experimentation, for weight updates, we use a learning rate that exponentially decays from $0.01$ with the base $0.96$ and the decay rate $1/6000$. 

\subsection{Routing and Assignments}\label{heuristics}

The action space is very large because an action determines an update of the routing plan. To overcome the large action space, we apply the assignment and routing heuristic from \cite{drones} (where the algorithmic details can be found). 

The heuristic maintains the plans for drones and vehicles in the case the customer is not assigned to the corresponding fleet type. Otherwise, it changes the plans for drones and vehicles as follows.

For drones, the heuristic assigns customers in a first-in-first-out (FIFO) manner. This procedure prioritizes drones idling at the depot and assigns a new customer to an idling drone prior to one en route. If all the drones are en route, we assign a new customer to the drone with the earliest availability. We arbitrarily choose a drone if there is a tie. It is not feasible to service a customer by drone if no drones can deliver the goods by the customer's delivery deadline.

For vehicles, the heuristic first checks if there is a vehicle currently idling at the depot. In that case, it assigns the new customer to this vehicle. If no vehicles are idling at the depot, the heuristic iterates through all vehicles to find the route with the cheapest feasible insertion position. An insertion (update) is feasible if it satisfies the conditions described in Section~\ref{markov}. For every feasible insertion, we then calculate the increase in the tour time of the vehicle to which the new customer is inserted, called insertion cost. We assign the new customer to the vehicle with the insertion that minimizes the cost over all feasible insertions. We denote this insertion cost by $\Delta$\textsubscript{Vehicle}. If there is no feasible insertions for any vehicle, then it is not feasible to serve the new customer by the fleet of vehicles. Based on the observations in \cite{drones}, we omit waiting at the depot and start the next tour right after arriving back from the previous one.

We note that the heuristic provides only one possible vehicle and one possible drone assignment to the DQL-approach. This procedure has the advantage that the action space is limited and the learning process is likely to be fast. However, such a limited action space may ignore more effective assignments. We explore this question in Appendix~\ref{sec:assignment}.

\subsection{Features}\label{feature}

In this section, we present the features in the DQL for the SDDPVD. In Appendix \ref{analytical}, we present the analytical results that motivate the choice of features by examining the functional form of the Bellman equation as well as several situations in which we can characterize optimal action selection. Using the analysis in Appendix \ref{analytical}, we can identify information that needs to be extracted from the state that is to be input into the NNs. The features comprise information about the time, the customers, and the fleet. Our DQL approach uses all the features presented below. Extracted features are normalized using \textit{min-max normalization} before being input to the NNs. 

\textit{\textbf{Time.}} Given a state $S_k$, the first feature is the time of the decision point $t_k$.  Proposition~\ref{monotone_time} shows that the value function is monotonically decreasing in time, and the point of time of a decision point should help determine the Q-value. Computational results presented in papers such as \citet{marlin} have also demonstrated the value of the point in time for SDD problems. 


\textit{\textbf{Fleet.}} To reflect the availability of delivery resources in the state $S_k$ at a decision point, we include as features the time at which the vehicles and drones return to the depot from their ongoing routes. As shown in Propositions~\ref{monotone_veh} and \ref{monotone_drone}, we know that the value function is monotonically decreasing in these values. 
The available time for vehicle $v$ is $a(N_2^\theta)$ indicated in its planned route as described in Section \ref{prep_model}. Similarly, for drone $d$, its available time is $a(N_{-1}^\theta)$.


\textit{\textbf{Actions.}} We also include two features that capture the resource consumption that would occur if the customer request is accepted. These features also incorporate information about the requesting customer. The first feature is the point-to-point distance $d(C_k)$ between $C_k$ and the depot $\mathcal{N}$. Proposition~\ref{rej_prop} shows that this distance plays an important role in determining whether or not to assign a customer to a drone. However, the feature is not suitable for making decisions concerning the assignments to vehicles as they do not travel in a point-to-point fashion. To this end, we also consider a second feature $\Delta$\textsubscript{Vehicle}, the insertion time required when assigning the customer to a vehicle. 

\section{Computational Study}\label{design}

In this section, we present the computational study for the SDDPVD. 
We first present the settings for our computations and describe the benchmark policies. We then analyze solution quality, the impact of feature selection, and performance with distributional data change. We refer the reader to Appendix \ref{appendix} for detailed illustrations of decision-making process, and comparisons of our proposed approach to its extension or some other method in the literature. The results presented in these sections are computed on a test set of $1000$ instances that are different from the training set, but follow the same distributions. 

\subsection{Computational Settings}\label{instance}

The settings assume delivery requests are made from $8$ am to $3$ pm and that vehicle drivers work eight hours from $8$ am to $4$ pm thus $t^V_{\max}=8$ hours. The drones are available from $8$ am to $8$ pm thus $t^D_{\max}=12$ hours. We assume that accepted requests must be serviced within $\bar{\delta}=4$ hours. The loading and service times for both vehicle and drones are each $3$ minutes ($t_N^V=t_N^D=t_C^V=t_C^D=3$ minutes). The battery charging time required for drones is set to $t^D_B=20$ minutes, which is a conservative estimate based on the discussion in \cite{grippa}.

The vehicles travel at a speed of $30$km/h. We assume vehicles travel on a road network. To reflect the effect of road distances and traffic, we transform Euclidean distances using the method introduced by \cite{boscoe}.  The method transforms Euclidean into an approximate street-network distance by multiplying the Euclidean distance between two points by $1.5$. We compute the travel time based on these transformed distances. 

We assume drones travel in a point-to-point fashion at a speed of $40$km/h. This speed is a conservative estimate of drone capability accounting for security measures within the city \citep{pero}. As described in Section~\ref{prob_narrative}, we assume drones are capable of carrying any package in the SDDPVD. 

In our experiments, we consider three different sets of customer requests that differ in spatial and temporal distributions. \begin{itemize}
    \item Homogeneous Customer Demand. Both the temporal and spatial distributions of customer requests are homogeneous for this set. For time of requests, we assume customers make delivery requests according to a homogeneous Poisson process with $500$ requests expected in each day. For customer locations, the $x$ and $y$ coordinates of customer locations are generated from independent and identical normal distributions with the depot at the center. This set of instances is the one used in \citet{drones}. This geography reflects the structure of many cities in Europe where most customers are in the central area and the rest are sparsely located in the suburb. We set the standard deviation to $3.0$km for each coordinate resulting in $50$\% of the customers being in a core that is within 10-minute vehicle travel time from the depot (approximately $3.3$km) and about $99.9$\% of the customers being within $30$ minutes (approximately $10$km). This distance is within the travel capability of existing drones, which can carry payloads weighing up to $5$ pounds and travel up to $12.5$ miles ($20$km) \citep{ups_drone}. 
    \item Spatially Heterogeneous Customer Demand. In this set, the temporal distribution of requests is the same as that for homogeneous customer demand. The spatial distribution is heterogeneous over time. This geography is motivated by the idea that, in the beginning of the day, customers often order to their homes, in the middle of the day, more customers order to work, and later, more customers order to their homes again. Thus, we vary the standard deviation of the customer coordinates. In the first and last two hours, it is $3.0$km as before. For the three hours in between, it is reduced to $1.0$km.  
    \item Temporally Heterogeneous Customer Demand. The spatial distribution of this set is the same as that for the homogeneous customer demand. For the temporal distribution, we consider a heterogeneous rate that mimics peak and off-peak hours of requests in real world. The arrivals still follow a Poisson process. However, we vary the arrival rate of requests over time in a way that they are normally distributed in two peaks around times $t=90$ and $t=300$ with standard deviations of 30 minutes. 
\end{itemize}

For each of the three sets of requests, we test nine different combinations of fleet sizes. We consider combinations of $2$, $3$, and $4$ vehicles with $5$, $10$, and $15$ drones.

\subsection{Benchmarks}\label{benchmark}

In this section, we introduce the policies that we use to benchmark the performance of our DQL approach. In the main body of the paper, we present comparisons to five benchmarks, three existing and two new policies. Additional benchmarks are introduced and analyzed in Appendix. We first briefly review the PFA approach introduced in \citet{drones} and its two special cases. We then introduce two additional PFA variants. 



As discussed in Section \ref{ssd_lit}, \cite{drones} introduce a PFA approach to solve the SDDPVD. The authors consider a policy that incorporates the intuition that drones are suitable to serve the customers that are farther from the depot and vehicles serve those that are closer. The PFA policy $\pi\textss{PFA}$ is parameterized by a vehicle-travel-time threshold $\tau$. That is, the only feature that $\pi\textss{PFA}$ uses is $d(C_k)$. The policy works as follows. When customer $C_k$ makes a delivery request at $t_k$, the dispatcher assigns a request either to a drone if $d(C_k)$ greater than the threshold $\tau$ or to a vehicle, if $d(C_k) \leq \tau$. When the corresponding fleet type is not available, the order is assigned to the alternative fleet, or it is rejected in the case such an assignment is also not feasible.

We then consider two special cases of the PFA policy. The first one is $\pi\textss{PFA\_dro}$, where we preferably serve customers by drones ($\tau=0$). We address it as \textit{drones first} as it always assigns requests to drones whenever it is feasible, and considers assignments to vehicles only when no drones are available. The second one is $\pi\textss{PFA\_veh}$ in which customers are served preferably by vehicles ($\tau=\infty$). It is addressed as \textit{vehicles first} as it oppositely considers assignments to vehicles first and then drones.

We also consider a similar policy to $\pi\textss{PFA}$ that allows the rejection of feasible customers. We call this policy $\pi\textss{PFA\_rej}$. In this policy, when a customer is infeasible with regard to the fleet designated by the threshold, the customer is not offered service. For example, if a customer falls within the threshold but cannot be feasibly served by any vehicle, then it is offered no service rather than being considered for service by a drone. For both policies $\pi\textss{PFA}$ and $\pi\textss{PFA\_rej}$, we learn the threshold $\tau$ in the manner described in \citet{drones}.

To take advantage of alternative state information, we also consider a PFA-based policy $\pi$\textsuperscript{Delta} that is parametrized by an insertion-cost threshold $\kappa$ for vehicles. As described in Section \ref{heuristics}, every new customer is associated with a potential insertion cost $\Delta$\textsubscript{Vehicle} if it is feasible to serve the customer by vehicles. One of the goals of  using drones in SDD is to discourage vehicles from traveling to remote areas because such dispatches can be costly. The $\pi$\textsuperscript{Delta} is designed to avoid these less preferred decisions by controlling the insertion cost to be within a threshold $\kappa$. Given a new customer request, the policy checks feasibility for vehicles and the corresponding insertion cost $\Delta$\textsubscript{Vehicle}. If service by a vehicle is feasible and $\Delta\textsubscript{Vehicle}<\kappa$, the customer is assigned to the vehicle. Otherwise, the policy selects delivery by drones if feasible, or no service is offered. Parameter $\kappa$ is determined by enumeration.

\subsection{Solution Quality}\label{s_q}

We use the solution quality $\mathcal{Q}$, the average number of served customers, as the measure of performance of different policies. For each policy $\pi$, we define its solution quality as the average percentage of served orders:
\begin{equation}
\mathcal{Q}(\pi)=\frac{\textrm{Served}}{\textrm{Requests}}.
\end{equation}
To compare the performance of different policies to the main benchmark $\pi\textss{PFA}$, we define the improvement of $\pi^a$ over $\pi\textss{PFA}$ as:
\begin{equation}
\mathcal{P}(\pi^a)=\frac{\mathcal{Q}(\pi^a)-\mathcal{Q}(\pi\textss{PFA})}{\mathcal{Q}(\pi\textss{PFA})}.
\end{equation}

The results are summarized in Table~\ref{results}. We denote our DQL approach $\pi\textsuperscript{Q}$. For $\pi$\textsuperscript{PFA}, we present the average number of customers served. We then use $\pi$\textsuperscript{PFA} as a benchmark and report the other policies' performance relative to $\pi$\textsuperscript{PFA}. We also perform a paired-sample $t$-test with $\pi$\textsuperscript{PFA}. The mark * indicates that the p-value of a paired-sample $t$-test is greater than $0.01$. 

\begin{table}[h!]
    \normalsize
	\centering
	\begin{tabular}{ccccccc}
	\toprule
		\multicolumn{1}{c}{\begin{tabular}[c]{@{}c@{}}Fleet Size\\
		(Veh, Drone)\end{tabular}} &
		\multicolumn{1}{c}{$\pi\textss{PFA}$}  &
		\multicolumn{1}{c}{$\mathcal{P}(\pi\textss{PFA\_dro})$}  &
		\multicolumn{1}{c}{$\mathcal{P}(\pi\textss{PFA\_veh})$}  &
		\multicolumn{1}{c} {$\mathcal{P}(\pi\textss{Delta})$}&
		\multicolumn{1}{c}{$\mathcal{P}(\pi\textss{PFA\_rej})$} &
		\multicolumn{1}{c}{$\mathcal{P}(\pi\textss{Q})$} \\
		\cmidrule{1-7}
\multicolumn{7}{c}{homogeneously distributed customers}\\
		\cmidrule{1-7}
		2, 5   & 227.6  & -8.0  & -6.9   & 5.8 & 18.7 &  22.0\\ 
		2, 10  &  312.8 & -10.1 & -9.2  & 0.5 &8.8 & 10.7\\ 
		2, 15  & 391.1  & -11.6 &-9.6   & -1.4 &4.4 & 6.8\\ 
		3, 5  & 293.4   &-8.1   &-7.4   &  3.5 &13.5  & 16.7\\ 
		3, 10 & 376.2   & -11.1  &-9.7  & -0.3*   &6.5   &9.2\\ 
		3, 15 & 460.3   &-14.6   &-11.9  &  -3.7  &   0.3  &3.0\\ 
		4, 5 & 354.7    &-8.1    &-7.4  &   1.9  &9.0  & 11.0\\ 
		4, 10 & 439.9   &-13.0   &-10.9   &   -2.3 &2.7 &3.6\\ 
		4, 15 & 499.6   &-12.2   &-9.4  &  -2.2  &-0.9 &0.0*\\ 
		\cmidrule{1-7}
		\multicolumn{7}{c}{heterogeneously distributed customers in space}\\
		\cmidrule{1-7}
		2, 5  & 255.2  &-10.1  &-6.0   & 9.5  &14.4  &21.9\\ 
		2, 10 & 349.9 &-12.7  &-10.0  &  2.1  &4.9 &10.2\\ 
		2, 15 & 449.4 & -9.6 &-12.6  &  -3.8 &-2.5 & 3.2\\ 
		3, 5 & 336.2  &-11.3  & -8.8 &   9.0  &10.9  &16.6 \\ 
		3, 10 & 441.6 &-11.7  &-13.7  &  -0.4   &0.7  &5.2\\ 
		3, 15 & 498.0 &-7.3  & -9.1  &  -1.0 &-1.4   &0.0*\\ 
		4, 5 & 425.2 &-10.6  &-12.1  &   2.6 &4.5   & 6.5\\ 
		4, 10 & 497.7 &-7.8  &-11.0  &  -1.2 &-1.2    &-0.1  \\ 
		4, 15 & 499.2 &-0.9  &-1.3 &   0.0*  &0.0*  &0.0\\ 
		\cmidrule{1-7}
		\multicolumn{7}{c}{heterogeneously distributed customers in time}\\
		\cmidrule{1-7}
		2, 5  & 216.7  &-8.7  &-10.3   & 5.9  & 20.4  & 20.9\\ 
		2, 10 & 291.7 &-10.9  &-11.4  &  0.8  & 10.6 & 12.2\\ 
		2, 15 & 363.2 & -11.6 &-11.6  &  -1.7 & 5.3 & 4.7\\ 
		3, 5 & 280.8  &-8.5  & -9.7   &  4.4  & 16.6  & 17.1 \\ 
		3, 10 & 358.0 &-12.0  &-11.8  &  -0.6 & 8.0  & 6.7\\ 
		3, 15 & 433.8 &-13.6  & -13.4  & -4.0 & 1.8   & -1.4\\ 
		4, 5 & 342.2 &-8.6  & -9.0   & 2.7   & 11.3   & 10.7\\ 
		4, 10 & 422.0 &-12.2 &-12.2  & -2.3  & 3.3   & 1.6 \\ 
		4, 15 & 485.9 &-12.8  &-12.3  & -3.8  & 0.0   & -3.4 \\ 
		\bottomrule
	\end{tabular}
	\caption{Solution Quality: Improvements (\%) over $\pi\textsuperscript{PFA}$ for different fleet setups and customer distributions.}
	\label{results}
\end{table}

Overall, our DQL approach $\pi\textss{Q}$ outperforms the benchmarks and improves upon $\pi\textss{PFA}$ by up to $22\%$. The value of $\pi\textsuperscript{Q}$ is greatest in the cases in which resources are more constrained. For example, with $2$ vehicles and $5$ drones, $\pi\textss{PFA}$ can only serve about $45.5\%$ ($51.0\%$/$43.3\%$) of homogeneously (spatial-/temporal- heterogeneously) distributed customers. The policy $\pi\textss{Q}$ improves the solution quality by more than $20\%$ in all three cases. As resources become less constrained, the relative performance of $\pi$\textsuperscript{Q} diminishes. Simply, with abundant resources, there is sufficient slack to overcome the poorer decisions of the benchmark policies.

In only a very few cases, $\pi\textss{PFA}$ achieves better results than $\pi\textsuperscript{Q}$. Yet, in those settings, the difference is small, and both $\pi\textss{Q}$ and $\pi\textss{PFA}$ serve a large percentage of customers. Given sufficient resources, the PFA is an effective policy. In these cases, the option of not offering service is therefore not beneficial. Because decision making of $\pi\textss{Q}$ is based on approximated values, in a few cases no service is offered nonetheless. This is likely the reason for the slight inferiority of $\pi\textss{Q}$ for these instances. 





With the same size of the fleet, all policies serve more customers in the spatially heterogeneous instances than those in the homogeneously distributed instances. This difference results from the fact that, on average, customers for the spatially heterogeneous distribution are closer to the depot and therefore easier to serve. For the temporally heterogeneous instances, fewer customers are served, likely because, during off-peak times, vehicles are not fully utilized while during peak-times, the number of customers is too large for the given resources.

Another interesting observation is the relative performance of policies $\pi\textss{PFA}$ and $\pi\textss{Delta}$. For instances with only a few drones, $\pi\textss{Delta}$ outperforms $\pi\textss{PFA}$. This changes when the number of drones increases. Policy $\pi\textss{PFA}$ utilizes a drone-centric feature of direct travel time to the customer while $\pi\textss{Delta}$ draws on the corresponding, but vehicle-centric feature of the increase in route duration. Thus, by shifting the fleet composition from vehicles to drones, the performance advantage shifts from $\pi\textss{Delta}$ to $\pi\textss{PFA}$ as well. We further observe that $\pi\textss{PFA\_rej}$ performs better than $\pi\textss{PFA}$. Thus, just having the option of not offering service already improves solution quality. Finally, both policies $\pi\textss{PFA\_dro}$ and $\pi\textss{PFA\_veh}$ perform very poorly regardless of the instance setting.

In the next sections, we analyze the impact of feature selection and robustness of our policies. For an illustration of decision-making process, we refer the reader to Appendix \ref{appendix_other_benchmark}.


\subsection{Impact of Feature Selection}\label{feature_selection}

To demonstrate the value of our feature selection, we present how our policy performs for different subsets of features. For the computations in this section, we use the customer demand that follows a homogeneous distribution in both time and location with $3$ vehicles and $10$ drones. 

\begin{figure}[h!]
	\begin{minipage}[b]{0.5\linewidth}
		\centering
		\includegraphics[width=0.9\linewidth]{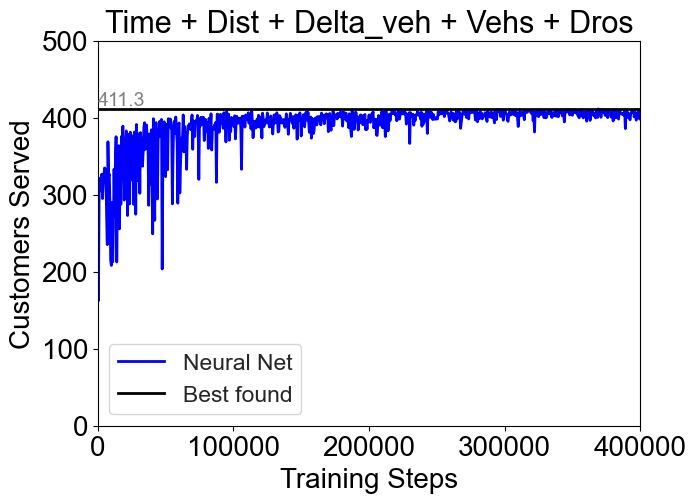}
	\end{minipage} 
	\begin{minipage}[b]{0.5\linewidth}
		\centering
		\includegraphics[width=.9\linewidth]{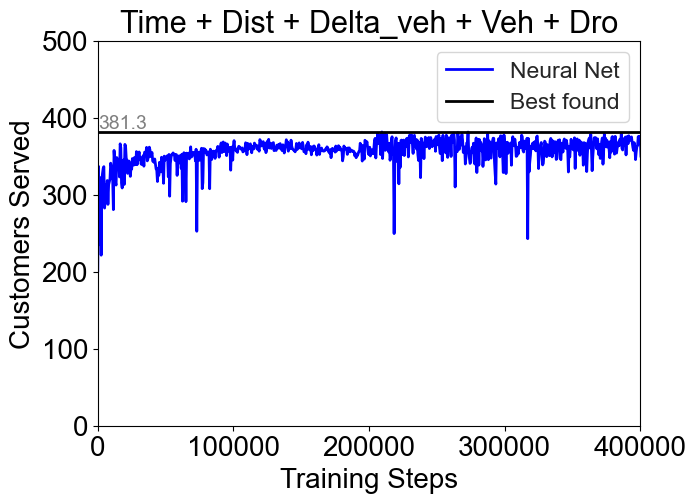}
	\end{minipage} 
	\begin{minipage}[b]{0.5\linewidth}
		\centering
		\includegraphics[width=.9\linewidth]{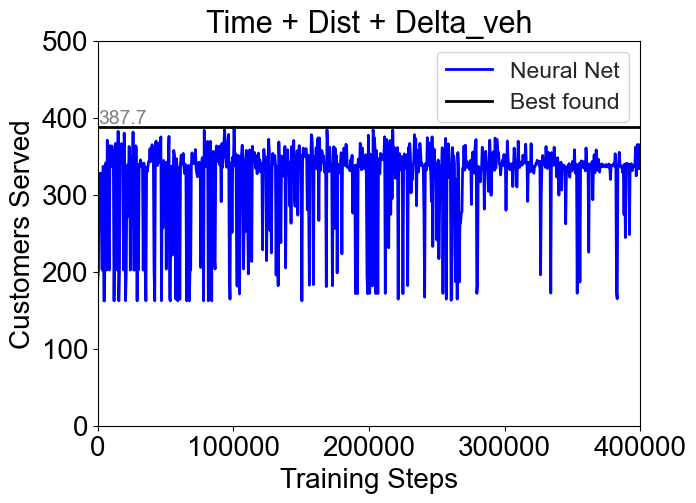}
	\end{minipage} 
	\begin{minipage}[b]{0.5\linewidth}
		\centering
		\includegraphics[width=.9\linewidth]{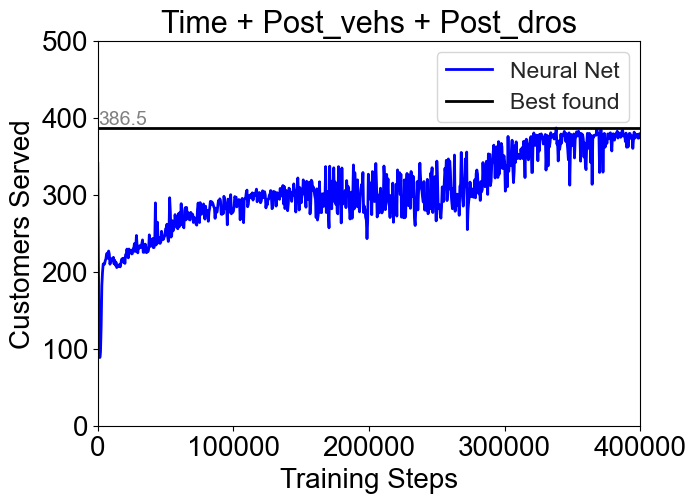}
	\end{minipage} 
	\caption{Learning process for different sets of features (The proposed DQL-approach is shown on the top left), for homogeneous customer demand with 3 vehicles and 10 drones.}
	\label{3v10d_with_rej}
\end{figure}

Figure~\ref{3v10d_with_rej} presents the results of the training for four different feature sets. Additional sets are presented in Appendix~\ref{appendix_feature_selection}. All sets contain the point of time of the current decision point (\enquote{Time}). The first set, whose results are shown in the top left, contains the features proposed in this paper, the DQL-features reflecting the state and action spaces. The action space is represented by \enquote{Dist}, which is a proxy for the travel time when sending a drone, and \enquote{Delta\_veh}, which is the additional travel time when assigning a vehicle. The state space is reflected by the availability times for all drones (\enquote{Dros}) and vehicles (\enquote{Vehs}). The second subset, whose results are shown are shown on the top right, contains similar features, but focuses on the directly affected single vehicle (\enquote{Veh}) and single drone (\enquote{Dro}), ignoring the state of the rest of the fleets. This set provides a comparison to the approach of \citet{chen2019} who do not consider fleet interactions in their approach. Q-learning considers the value of a state-action pair. In contrast, the third and fourth sets consider only one part, either the action or state information. The third set, whose results are shown on the bottom left, contains features that reflect only the action space of the problem (\enquote{Dist} and \enquote{Delta\_veh}). This set therefore ignores the state information about the utilization of the fleets. 

As a fourth set of features, we consider a different approach. We use features representing the post-decision state. The results of using post-decision features are shown on the bottom right of Figure~\ref{3v10d_with_rej}.

The post-decision state represents the state immediately following an action selection, but before the realization of new exogenous information. In this case, the post-decision state includes the point of time as well as the availability times of vehicles (\enquote{Post\_vehs}) and drones (\enquote{Post\_dros}) that result from an assignment or a rejection of a request. These availability times are extracted after the heuristics are applied, in the case the request is offered service. However, the post-decision features do not include features of the action space such as the distance to the request or the deviation in the vehicle routes needed to include the request, the value we refer to as $\Delta_{\text{Vehicle}}.$ 
 
The use of the post-decision state is similar to what is proposed in \citet{joe2020}. We also evaluate the ability to learn the value of post-decisions states because post-decision state VFA is common in the literature (see \citet{marlin}) and particularly because post-decision state VFA is used in \citet{drones}, a paper that also looks at SDD but does not consider a heterogeneous fleet.

Each sub-figure of Figure~\ref{3v10d_with_rej} plots the solution quality curve for the first 400,000 training steps. The horizontal axis represents the number of training steps, and the vertical axis represents the solution quality. The horizontal line represents the best found solution value. 

We observe that the feature selection proposed in this paper outperforms all other selections. The second and third sets show substantially worse solution quality and limited learning. The fourth set shows a constant, but slow learning process.  Even after 400,000 training runs, the solution quality still increases. We note that the VFA can implicitly access the action-space features (\enquote{Dist} and \enquote{Delta\_veh}) when comparing the different assignment decisions. However, using the features explicitly as input in the DQL such as we do in the first set leads to more effective approximation and better solution quality, at least for this number of training steps. The action-space features are likely to guide decision making, especially in early training runs when the approximation is still weak. This indicates that enriching a value function approximation with action-space features may be beneficial for large-scale problems such as those often observed in dynamic vehicle routing.

\subsection{Robustness of DQL Policies}\label{robustness_main}

In this section, we explore how DQL policies perform when the instance data is varied. We first present the performance of the polices when the fleet size changes. We then show more results when the spatial/temporal distribution of demand is varied.

\subsubsection{Varying Fleet Size}\label{sec:varying_fleet_size}

Our presented approach learns the values for predefined numbers of vehicles and drones. We now analyze how our policy performs when the size of fleet changes. To this end, we take the policy trained for $3$ vehicles and $10$ drones and test its performance when the number of vehicles changes. Specifically, we change the number of vehicles to $2$ and $4$. To apply our method, we then adjust the features for each case because the trained policy expects $3$ vehicle availability times. For experiments with $2$ vehicles in the fleet, we set the availability time of the third vehicle to $t^V_{\max}$ throughout the day. For $4$ vehicles, we use the earliest and latest vehicle availability times along with the average value of the middle two as the three adjusted features. 

\begin{table}[h]
	\centering
	\normalsize
	\begin{tabular}{cccc}
		\toprule
		\multicolumn{1}{c}{\ } & \multicolumn{1}{c}{2 vehs. \& 10 drones} &  \multicolumn{1}{c} {3 vehs. \& 10 drones}&  
		\multicolumn{1}{c}{4 vehs. \& 10 drones}  \\
		\cmidrule{1-4}
		upper bound & 346.2 &   410.7  & 455.9 \\ 
		$\pi\textsuperscript{Q}$ (3 vehs. \& 10 drones) & 333.6 &   410.7  & 454.5\\ 
		$\pi\textsuperscript{PFA}$ (3 vehs. \& 10 drones) & 308.8 &376.2& 430.8\\ \bottomrule
	\end{tabular}
	\caption{Solution quality vs. varied fleet sizes over homogeneous customer demand.}
	\label{varying_fleet_size}
\end{table}

Table~\ref{varying_fleet_size} presents the solution quality for both $\pi\textsuperscript{Q}$ and $\pi\textsuperscript{PFA}$ evaluated with different numbers of vehicles. The first row \textit{upper bound} presents the policies trained for each fleet size using the original approach where the NNs are trained on the actual fleet size. As shown in the table, $\pi\textsuperscript{Q}$ still outperforms $\pi\textsuperscript{PFA}$ when the number of vehicles changes. When the delivery resources become constrained, both policies show a significant gap compared to the upper bound, with -$3.6\%$ for $\pi\textsuperscript{Q}$ and -$10.8\%$ for $\pi\textsuperscript{PFA}$. When there are additional vehicle resources, $\pi\textsuperscript{Q}$ can serve almost as many customers as the upper bound does, with the gap of only -$0.3\%$, while $\pi\textsuperscript{PFA}$ suffers a loss of $5.5\%$ customers compared to the upper bound. This analysis suggests that, when the fleet size changes moderately, our proposed approach still enables effective decision making. 

\subsubsection{Varying Demand Distribution}\label{sec:varying_demand}

In the following, we analyze how our policies handle changes in the demand distribution. To this end, we consider two pairs of customer distributions, the homogeneous and temporally heterogeneous distributions, and the homogeneous and spatially heterogeneous distributions. For each pair, we then stepwise vary the evaluation instances between the two distributions. In each step, we take a percentage of requests from the heterogeneous distribution while the rest of requests is taken from the homogeneous one. The percentage ranges from $0\%$ (fully homogeneous) to $100\%$ (fully heterogeneous), with a step size of $10\%$. Note, $0\%$ and $100\%$ indicate the data sets on which our presented DQL policies are trained. For all the following tests, we consider a fleet of 3 vehicles and 10 drones. As a benchmark, we apply the same procedure for $\pi\textsuperscript{PFA}$.

\begin{figure}[h]
	\centering
	\includegraphics[width=0.7\textwidth]{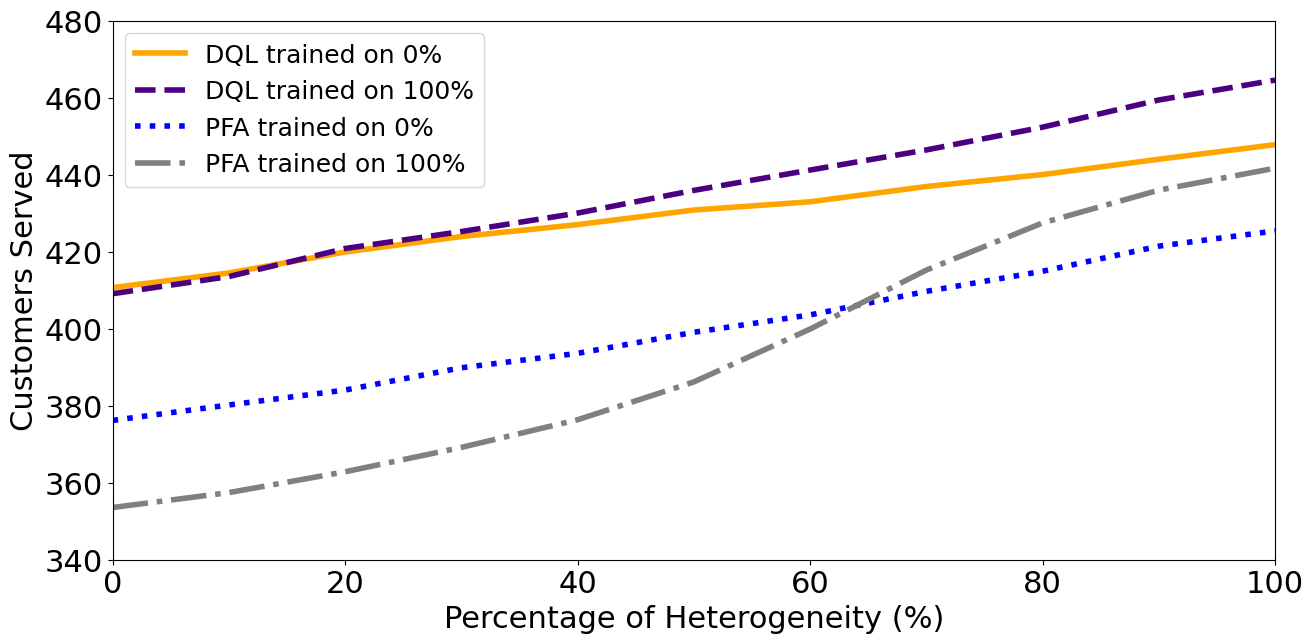}
	\caption{Solution quality for DQL policies $\pi\textsuperscript{Q}(0\%)$ ($\pi\textsuperscript{PFA}(0\%)$) trained on homogeneous data and policy $\pi\textsuperscript{Q}(100\%)$ ($\pi\textsuperscript{PFA}(100\%)$) trained on spatially heterogeneous data for varied spatial distribution of requests.}
	\label{fig:spatial_main}
\end{figure}

We first analyze changes in the spatial distribution. Figure~\ref{fig:spatial_main} presents the solution quality for the policies as we vary the spatial distribution of requests. Policies $\pi\textsuperscript{Q}(0\%)$ and $\pi\textsuperscript{PFA}(0\%)$) are trained on the fully homogeneous data, while $\pi\textsuperscript{Q}(100\%)$ and $\pi\textsuperscript{PFA}(100\%)$ are trained on the fully heterogeneous data. All policies show a monotonic increase as percentage of heterogeneity increases because customers become more clustered and geographically closer to the depot. 

We observe that $\pi\textsuperscript{Q}$ outperforms $\pi\textsuperscript{PFA}$ for every instance setting. For both policy types $\pi\textsuperscript{Q}$ and $\pi\textsuperscript{PFA}$, we see that the policies trained on the original data outperform the respective counterpart. However, $\pi\textsuperscript{Q}(100\%)$ can serve almost as many customers as $\pi\textsuperscript{Q}(0\%)$ for the fully homogeneous case, while $\pi\textsuperscript{Q}(0\%)$ serves about $17$ customers fewer on average than $\pi\textsuperscript{Q}(100\%)$ in the fully heterogeneous case. Thus, the DQL-policy trained on heterogeneous data seems more \enquote{robust} to changes, likely because the range of observations during training might be larger.

\begin{figure}[h]
	\centering
	\includegraphics[width=0.7\textwidth]{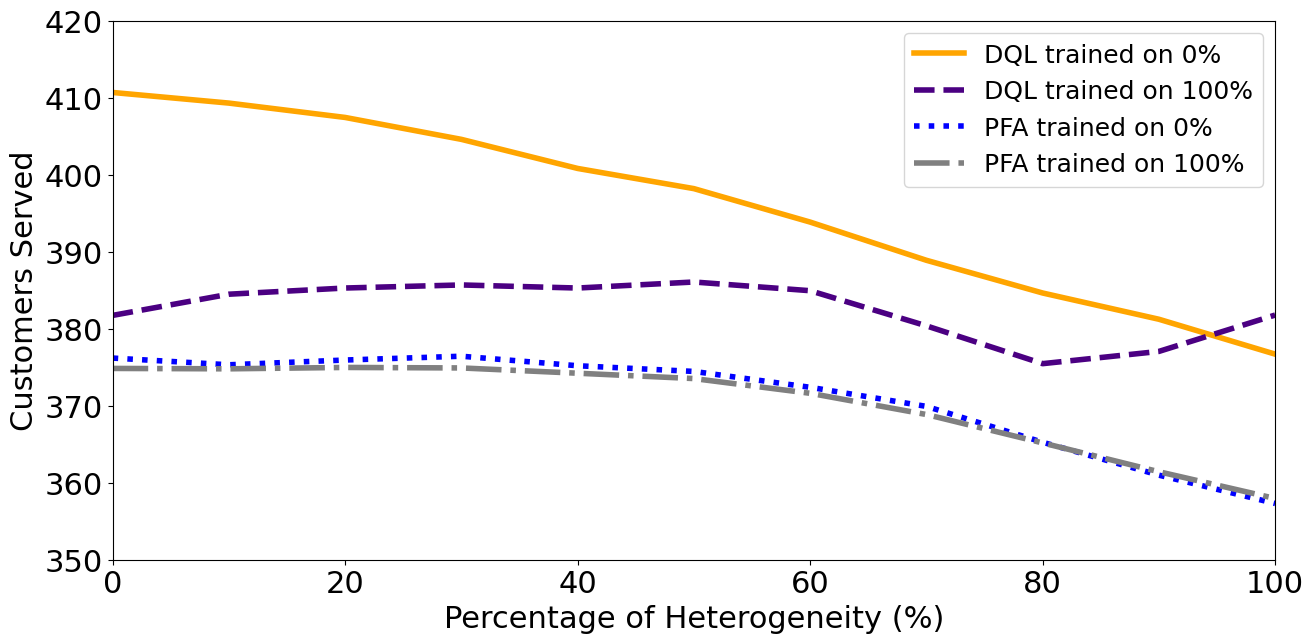}
	\caption{Solution quality for DQL policies $\pi\textsuperscript{Q}(0\%)$ ($\pi\textsuperscript{PFA}(0\%)$) trained on homogeneous data and policy $\pi\textsuperscript{Q}(100\%)$ ($\pi\textsuperscript{PFA}(100\%)$) trained on temporally heterogeneous data for varied temporal distribution of requests.}
	\label{fig:temporal_main}
\end{figure}

We next analyze how the policies perform when the arrival rates change. Figure~\ref{fig:temporal_main} presents the solution quality for the four stated policies evaluated on the data with different heterogeneity in the temporal distribution. 

Again, the DQL policies outperform the PFA in all the temporal settings. Further, we observe a decrease in solution quality with increasing heterogeneity in request times. This indicates that demand peaks may lead to a congestion of the fleet resources and therefore fewer services.

As with the spatial experiments, the respective policies perform best when applied to instances similar to their training data. However, we observe that the policy trained on instances with high temporal heterogeneity $\pi\textsuperscript{Q}(100\%)$ performs significantly worse compared to $\pi\textsuperscript{Q}(0\%)$. Policy $\pi\textsuperscript{Q}(100\%)$ reserves resources in off-peak times in expectation of a large amount of requests (see Figure~\ref{T_dql_quantify}). For instances without such peaks, such reservations are unnecessary and may lead to an ineffective use of the fleet. 

\section{Conclusions and Future Work}\label{conclusions}



In this paper, we present a SDD problem with vehicles and drones and approximately solve it using reinforcement learning. For this purpose, we identify particular features of the state to include as inputs to the NNs and estimate values of state-decision pairs. Computational results demonstrate that the method is capable of making service decisions and assignments that appropriately balance the use of vehicles and drones throughout the day resulting in increased expected number of customers served relative to benchmarks. We also show that our policies can still maintain the effectiveness when the fleet size or input data changes moderately.

There are various directions for future research. The analytical results show that the value function of the SDDPVD is monotonic in a number of the state elements. One direction for future research is to explore the enforcement of monotonicity to the learning process for the problem. To the best of the authors' knowledge, the enforcement of monotonicity in NN approximations is relatively unexplored. 
Our results show that the proposed solution method can cope with changing data. However, this ability depends on the training data. Future work could further explore this issue to more systematically identify how to structure training data for this and other dynamic vehicle routing problems.

We can consider additional instance parameters, for example, different vehicle travel speeds during peak and off-peak hours, and assignment restrictions due to package weight, aerial restrictions and etc.  In addition, the problem in this paper considers accepting or not customer requests for service. An alternative would be to consider the pricing of the deadline as a way of serving more customers. 

Finally, our policy learns assignments of customers to fleet types based on fleet and customer features. This strategy may be valuable for a variety of dynamic routing problems with heterogeneous fleets and/or heterogeneous customers.


\bibliography{refe2021.bib}
\bibliographystyle{abbrvnat}

\newpage

\appendix

\section{Appendix}\label{appendix}

In Appendix, we present the analytical results along with proofs, illustrations of the decision making, additional feature combinations, an analysis of the value of not offering service, and the results for expanding the assignment decisions.

\subsection{Analytical Results}\label{analytical}

In this section, we present a series of analytical results with proofs for the SDDPVD. We begin by characterizing Equation~\eqref{eq:bellman}. Our goal is to identify how the value function reacts to changes in the state. First, we show that Equation~\eqref{eq:bellman} is monotonically decreasing in time, the return time of the vehicles, and the return time of the drones. The results are formalized in Propositions~\ref{monotone_time}, \ref{monotone_veh}, and \ref{monotone_drone}.
\begin{proposition}\label{monotone_time}
	In the SDDPVD, let $t$ represent time of the decision point in the state. Then, the expected reward is monotonically decreasing in $t$.
\end{proposition}
\begin{proof}
	Note, in the SDDPVD and the simplified version, we always assume $t_v$ and $t_d$ are no earlier than $t$. For example, if $t=10$ in the current state, and the vehicle becomes available at $t=9$, then we set $t_v=10$.
	Consider two states that only differ in time of the decision point. Say $s=(t,\dots\ \dots)$ and $s'=(t',\dots\ \dots)$ such that $0\leq t<t'$. If the dispatcher is in the state $s$, the worst case is that, the dispatcher does not accept any requests that arrive during $[t,t')$, and then, starting $t'$, follows the same path as it will do from the state $s'$. In this worst case, the expected rewards are equal, $V(s)=V(s')$. On the other hand, the dispatcher expects to receive $n=\mu(t'-t)$ requests during $[t,t')$. If the dispatcher can accept feasible requests, then, compared to being in the state $s'$, the dispatcher has more requests from which to choose. Therefore, considering both cases, we have $V(s)\geq V(s')$.
\end{proof}
\begin{proposition}\label{monotone_veh}
	Suppose, in the SDDPVD, the fleet has one vehicle. Let $t_v$ represent the time at which the vehicle returns to the depot from the current route. Then, the expected reward is monotonically decreasing in $t_v$.
\end{proposition}
\begin{proposition}\label{monotone_drone}
	Suppose, in the SDDPVD, the fleet has one drone. Let $t_d$ represent the time at which the drone returns to the depot from the current route. Then, the expected reward is monotonically decreasing in $t_d$.
\end{proposition}
\begin{proof}
The proofs for $t_v$ and $t_d$ are similar, so we only present that for $t_v$. Consider two states that differ only in the time at which the vehicle becomes available. Say $s=(t_v,\dots\ \dots)$ and $s'=(t_v',\dots\ \dots)$ such that $t\leq t_v< t_v'$. The worst case is that the dispatcher can simply postpone $t_v$ to $t_v'$ and then follows the same path as it will do starting from the state $s'$. Thus, we have $V(s)=V(s')$ for the worst case. Similarly, on the other hand, the earlier available time of the vehicle results in $t_v'-t_v$ extra units of time for the vehicle to make deliveries. Hence, we have $V(s)\geq V(s')$.
\end{proof}
We next seek to characterize circumstances in which we can identify optimal actions. Our goal is to identify information that is needed to determine these optimal actions and to make sure that this information is available as a feature. Importantly, we analytically demonstrate the value that not offering service to a particular customer can have on the objective. Appendix~\ref{appendix_benchmark} offers a computational demonstration.

Because the structure of the SDDPVD is complex, we simplify the problem for this analysis. We assume that customers dynamically make delivery requests over $[0,\max\{t^D_{\max},t^V_{\max}\}]$. Customer requests are revealed following a Poisson process with rate $\mu$. Each requesting customer's distance from the depot $D$ follows a uniform distribution with the support $[0,D_{\max}]$, where $D_{\max}$ is the maximum possible distance. The distance $D$ between a customer and the depot is converted to the corresponding vehicle's travel time. In the remainder of this section, we use the term \enquote{distance} synonymously with travel time of the corresponding fleet type. The arrival times of requests are independent, and the random variables, time $t$, and distance $D$ are also independent. The fleet consists of a vehicle and a drone. We omit the loading and charging times for the fleet. The capacity of the vehicle is unlimited, and the capacity of the drone is 1. We assume that the drone travels $c$ ($>1$) times as fast as the vehicle.

We first consider the situation at the end of the day. Proposition~\ref{lastdispatch} identifies the circumstances in which a drone idling at the depot can serve at least one more customer before the end of the day. 
\begin{proposition}\label{lastdispatch}
	Suppose a drone is available when a new customer that is $b'$ (vehicle travel time) units from the depot requests service at time $t'<t^D_{\max}$. Then, we can serve this customer if $t'\leq t^D_{\max}-\frac{b'}{c}$.
\end{proposition}
\begin{proof}
	If the drone is dispatched to serve the request $b'$, it will return to the depot at $t'+\frac{b'}{c}$. For the drone returning to the depot no later than the end of the horizon $T$, simply solving the inequality $t'+\frac{b'}{c}\leq t^D_{\max}$ gives $t'\leq t^D_{\max}-\frac{b'}{c}$. 
	
	This proposition can also be seen as the condition for the idling drone to serve at least $1$ customer during $[t',t^D_{\max}]$. Simply dispatching the drone to serve $b'$ will increase the number of customers served by 1. It is also possible when the drone returns to the depot from $b'$, it is dispatched for other customers. This results in that, at least $1$ customer can be served by the drone during $[t',t^D_{\max}]$.
\end{proof}
We now determine whether we should serve an end-of-the-day customer request with an idling drone. Assume the vehicle is in its last route and will return to the depot at $t^V_{\max}$. The drone can feasibly serve the customer $b'$ at time $t'$ as given in Proposition~\ref{lastdispatch}. To determine whether to serve this request at $t'$, we want to know how many requests the drone is expected to serve during $[t',t^D_{\max}]$. Importantly, it is possible that, instead of serving the current request, the drone could serve several customers whose request arrive after $t'$ but that are close to the depot. 

To address this question, we calculate the probability that the drone can serve at least one more customer after serving $b'$. We present the probability in Lemma \ref{prob}. 
\begin{lemma}\label{prob}
	Suppose that the drone is available when a customer request that is located $b'$ (vehicle travel time) units from the depot that is revealed at time $t'$. If the drone is dispatched to serve the customer, then the probability that the drone can serve at least one more customer after returning to the depot is 
	\begin{equation}
	P_{[t',t^D_{\max}],n\geq 1\big|accept}=1-e^{-\frac{c\mu(t^D_{\max}-t'-\lceil\frac{b'}{c}\rceil)(t^D_{\max}-t'+\lceil\frac{b'}{c}\rceil-1)}{2D_{\max}}}.
	\end{equation}
\end{lemma}
\begin{proof}
At $t'$, the drone is dispatched distance\footnote{Throughout the proofs, we use \enquote{distance} synonymously with the travel time of the corresponding fleet type.} $b'$ so it will return to the depot at $t'+\frac{b'}{c}$, which divides $[t',  t^D_{\max}]$ into two disjoint sub-intervals $[t', t'+\lceil\frac{b'}{c}\rceil]$ and $(t'+\lceil\frac{b'}{c}\rceil, t^D_{\max}]$. We use the ceiling because the drone will not be dispatched until the end of a unit period.

We consider $[t', t'+\lceil\frac{b'}{c}\rceil]$ as a single period because the drone cannot start a second delivery tour until it is back to depot. Note, $\mu$ is the average number of arriving customer requests per unit period, and $[t', t'+\lceil\frac{b'}{c}\rceil]$ consists of $\lceil\frac{b'}{c}\rceil$ unit periods. The new rate $\mu_{[t', t'+\lceil\frac{b'}{c}\rceil]}$ for the single interval $[t', t'+\lceil\frac{b'}{c}\rceil]$ is 
	\begin{equation}
	\mu_{[t', t'+\lceil\frac{b'}{c}\rceil]}=\mu*\lceil\frac{b'}{c}\rceil.
	\end{equation} When the drone returns to the depot, the time left till the end of service period is $t^D_{\max}-t'-\lceil\frac{b'}{c}\rceil$, so for a customer revealed during $[t', t'+\frac{b'}{c}]$ to be able to be serviced by the drone, according to Corollary \ref{lastdispatch}, its distance $d$ must satisfy $$\frac{D}{c} \leq t^D_{\max}-t'-\lceil\frac{b'}{c}\rceil$$ 
	\begin{equation}
	D \leq c(t^D_{\max}-t'-\lceil\frac{b'}{c}\rceil).
	\end{equation}
Since the distance of customers is a uniform random variable with the support $[0, D_{\max}]$, the probability of each arriving request during $[t', t'+\lceil\frac{b'}{c}\rceil]$ that can be served by the drone is
	\begin{equation}
	P_{[t', t'+\lceil\frac{b'}{c}\rceil],feasible}=P(D \leq c(t^D_{\max}-t'-\lceil\frac{b'}{c}\rceil))=\frac{c(t^D_{\max}-t'-\lceil\frac{b'}{c}\rceil)}{D_{\max}}.
	\end{equation}
	Because customer request arrivals and service times are independent, it follows that such feasible customer requests arrive following a Poisson process with rate 
	\begin{equation}\mu_{[t', t'+\lceil\frac{b'}{c}\rceil],feasible} =\mu_{[t', t'+\lceil\frac{b'}{c}\rceil]}*P_{[t', t'+\lceil\frac{b'}{c}\rceil],feasible}.
	\end{equation}
	Hence, the probability that no such feasible requests arrive during $[t', t'+\frac{b'}{c}]$ is 
$$
	P_{[t', t'+\lceil\frac{b'}{c}\rceil],n=0}=P(n=0)
$$
$$
	=e^{\mu_{[t', t'+\lceil\frac{b'}{c}\rceil],feasible}}\frac{\mu_{[t', t'+\lceil\frac{b'}{c}\rceil],feasible}^0}{0!}
$$
$$
	=e^{-\mu_{[t', t'+\lceil\frac{b'}{c}\rceil],feasible}}
$$
\begin{equation}
=e^{-\frac{c\mu\lceil\frac{b'}{c}\rceil(t^D_{\max}-t'-\lceil\frac{b'}{c}\rceil)}{D_{\max}}}.
\end{equation}

Next, we consider $(t'+\lceil\frac{b'}{c}\rceil, t^D_{\max}]$ as $t^D_{\max}-t'-\lceil\frac{b'}{c}\rceil$ individual periods. For the $k$th minute after $t'+\lceil\frac{b'}{c}\rceil$, we can repeat the similar process performed for the interval $[t', t'+\lceil\frac{b'}{c}\rceil]$ and get the probability of no feasible requests arriving during the $k$th minute is
	\begin{equation}
	e^{-\frac{c\mu(t^D_{\max}-t'-\lceil\frac{b'}{c}\rceil-k)}{D_{\max}}}.
	\end{equation}
Because we assume arrival times of requests are independent, the probability that no feasible requests are revealed during the interval $(t'+\lceil\frac{b'}{c}\rceil, t^D_{\max}]$ is
$$
	P_{(t'+\lceil\frac{b'}{c}\rceil, t^D_{\max}],n=0}=P(n=0)
$$

	\begin{equation}
	=\prod_{k=1}^{t^D_{\max}-t'-\lceil\frac{b'}{c}\rceil } e^{-\frac{c\mu(t^D_{\max}-t'-\lceil\frac{b'}{c}\rceil-k)}{D_{\max}}}.
	\end{equation}

Therefore, the probability that, no feasible requests for the drone will arrive if we dispatch the drone to serve $b'$ is
$$P_{[t',t^D_{\max}],n=0\big|accpt}=P_{[t', t'+\lceil\frac{b'}{c}\rceil],n=0}*P_{(t'+\lceil\frac{b'}{c}\rceil, t^D_{\max}],n=0}$$
$$
=e^{-\frac{c\mu\lceil\frac{b'}{c}\rceil(t^D_{\max}-t'-\lceil\frac{b'}{c}\rceil)}{D_{\max}}}*\prod_{k=1}^{t^D_{\max}-t'-\lceil\frac{b'}{c}\rceil } e^{-\frac{c\mu(t^D_{\max}-t'-\lceil\frac{b'}{c}\rceil-k)}{D_{\max}}}
$$
\begin{equation}
=e^{-\frac{c\mu(t^D_{\max}-t'-\lceil\frac{b'}{c}\rceil)(t^D_{\max}-t'+\lceil\frac{b'}{c}\rceil-1)}{2D_{\max}}}.
\end{equation}
Then, the desired probability is
\begin{equation}
    P_{[t',t^D_{\max}],n\geq1\big|accpt}=1-P_{[t',t^D_{\max}],n=0\big|accpt}=1-e^{-\frac{c\mu(t^D_{\max}-t'-\lceil\frac{b'}{c}\rceil)(t^D_{\max}-t'+\lceil\frac{b'}{c}\rceil-1)}{2D_{\max}}}.
\end{equation}
\end{proof}

Instead of serving the current customer request arriving at time $t'$, the dispatcher can choose not to offer the service to customer. We would make such a decision if by doing so we would expect to serve more customers thereafter. The probability that the drone can serve at least one customer is given in Lemma \ref{rej_prob}. 
\begin{lemma}\label{rej_prob}
	The drone is available when a customer request $b'$ units from the depot is revealed at $t'$. If the dispatcher does not offer service to the new customer, then the probability that the drone can instead serve more than 1 customer thereafter is 
	\scriptsize
	\begin{multline}
	P_{[t', t^D_{\max}],n\geq 2\big|rej}=1-e^{-\frac{c\mu (t^D_{\max}-t')(t^D_{\max}-t'-1)}{2D_{\max}}}-\sum_{k=1}^{t^D_{\max}-t'}\bigg({e^{-\frac{kc\mu (2t^D_{\max}-2t'+k-1)}{2D_{\max}}}}\\
	\sum_{m=1}^{\lfloor c(t^D_{\max}-t'-k)\rfloor}\frac{\mu}{D_{\max}}e^{-\frac{\mu}{2D_{\max}}(2D_{\max}+ck+ck^2-ct^D_{\max}-2ckt^D_{\max}+cT^2+ct'+2ckt'-2ct^D_{\max}t'+ct'^2+c\lceil\frac{m}{c}\rceil-c\lceil\frac{m}{c}\rceil^2)} \bigg).
	\end{multline}
	\normalsize
\end{lemma}
\begin{proof}
We tackle the desired probability $P_{[t', t^D_{\max}],n\geq 2\big|rej}$ by computing\\ $1-P_{[t', t^D_{\max}],n=0\big|rej}-P_{[t', t^D_{\max}],n=1\big|rej}$.
\begin{itemize}
    \item $P_{[t', t^D_{\max}],n=0\big|rej}$
    
    The proof is similar to that for Lemma \ref{prob}. Simply replace the interval $(t'+\lceil\frac{b'}{c}\rceil, t^D_{\max}]$ by $[t',t^D_{\max}]$, and we get the probability
\begin{equation}
P_{[t', t^D_{\max}],n= 0\big|rej}=\prod_{k=1}^{t^D_{\max}-t'} e^{-\frac{c\mu(t^D_{\max}-t'-k)}{D_{\max}}}=e^{-\frac{c\mu(t^D_{\max}-t')(t^D_{\max}-t'-1)}{2D_{\max}}}.
\end{equation}
    \item $P_{[t', t^D_{\max}],n=1\big|rej}$
    
    Assume the drone is dispatched to the only customer it serves during $[t',t^D_{\max}]$ at the $k$th minute after $t'$. It divides $[t',t^D_{\max}]$ into three intervals $[t',t'+k-1]$, $(t'+k-1,t'+k]$ and $(t'+k,t^D_{\max}]$.
    
    During $[t',t'+k-1]$, the probability that there are no requests the drone can feasibly serve arriving is 
    \begin{equation}
        \prod_{l=0}^{k-1}e^{-\frac{c\mu (t^D_{\max}-t'+l)}{D_{\max}}}.
    \end{equation}
    
    We assume the only request the drone serves is made during $(t'+k-1,t'+k]$, and its distance is between integers $m-1$ and $m$. By Proposition \ref{lastdispatch}, the maximum distance of this feasible request is $c(t^D_{\max}-t'-k)$, which results in $m\in\{1,2,3,\dots,\lfloor c(t^D_{\max}-t'-k)\rfloor\}$. Given the distance of the request, by Proposition \ref{prob}, the probability the drone serves this customer and cannot serve any more customer thereafter is
    \begin{multline}
        \sum_{m=1}^{\lfloor c(t^D_{\max}-t'-k)\rfloor}e^{-\frac{\mu (m-1)}{D_{\max}}}e^{-\frac{\mu}{D_{\max}}}(\frac{\mu}{D_{\max}})\\
        e^{-\frac{\mu (D_{\max}-m)}{D_{\max}}}e^{-\frac{c\mu}{2D_{\max}}(t^D_{\max}-t'-k-\lceil\frac{m}{c}\rceil)(t^D_{\max}-t'-k+\lceil\frac{m}{c}\rceil-1)}.
    \end{multline}
    
    Iterating $k$ from 1 to $t^D_{\max}-t'$, we get the probability 
    \begin{multline}
    P_{[t', t^D_{\max}],n=1\big|rej}=\sum_{k=1}^{t^D_{\max}-t'}\Bigg( \bigg(\prod_{l=0}^{k-1}e^{-\frac{c\mu (t^D_{\max}-t'+l)}{D_{\max}}}\bigg)*\bigg(\sum_{m=1}^{\lfloor c(t^D_{\max}-t'-k)\rfloor}\\
    e^{-\frac{\mu (m-1)}{D_{\max}}}e^{-\frac{\mu}{D_{\max}}}(\frac{\mu}{D_{\max}})e^{-\frac{\mu (D_{\max}-m)}{D_{\max}}}e^{-\frac{c\mu}{2D_{\max}}(t^D_{\max}-t'-k-\lceil\frac{m}{c}\rceil)(t^D_{\max}-t'-k+\lceil\frac{m}{c}\rceil-1)} \bigg) \Bigg).
    \end{multline}
\end{itemize}
Hence, we get the simplified desired probability
\scriptsize
\begin{multline}
    P_{[t', t^D_{\max}],n\geq 2\big|rej}=1-P_{[t', t^D_{\max}],n=0\big|rej}-P_{[t', t^D_{\max}],n=1\big|rej}\\
    =1-e^{-\frac{c\mu (t^D_{\max}-t')(t^D_{\max}-t'-1)}{2D_{\max}}}-\sum_{k=1}^{t^D_{\max}-t'}\bigg({e^{-\frac{kc\mu (2t^D_{\max}-2t'+k-1)}{2D_{\max}}}}\\
	\sum_{m=1}^{\lfloor c(t^D_{\max}-t'-k)\rfloor}\frac{\mu}{D_{\max}}e^{-\frac{\mu}{2D_{\max}}(2D_{\max}+ck+ck^2-ct^D_{\max}-2ckt^D_{\max}+c(t^D_{\max})^2+ct'+2ckt'-2ct^D_{\max}t'+ct'^2+c\lceil\frac{m}{c}\rceil-c\lceil\frac{m}{c}\rceil^2)} \bigg).
\end{multline}
\normalsize
\end{proof}

Using Lemmas~\ref{prob} and \ref{rej_prob}, Proposition~\ref{rej_prop} identifies when the dispatcher should not accept a feasible request. 
\begin{proposition}\label{rej_prop}
	Assume that the drone is available when a customer request that is $b'$ units from the depot is revealed at $t'$. Let $b^*$ be the travel time that equates the probabilities in Lemmas \ref{prob} and \ref{rej_prob}. The dispatcher should always accept and assign the feasible request to the drone if $b'\leq b^*$, and not accept it if $b'>b^*$.
\end{proposition}	
\begin{proof}
Consider the probabilities in Lemmas \ref{prob} and \ref{rej_prob} as functions in $b'$, denoted $y_1(b')$ and $y_2(b')$. Because both functions involve ceiling and/or floor functions, it is impossible to solve for $b^*$ explicitly. Instead, we prove the proposition using the first derivative of $y_1(b')$. Note that $y_2(b')$ is constant in $b'$ because its expression does not contain $b'$. When $b'=0$,
\scriptsize
\begin{multline}
    y_1(0)-y_2(0)=\sum_{k=1}^{t^D_{\max}-t'}\bigg({e^{-\frac{kc\mu (2t^D_{\max}-2t'+k-1)}{2D_{\max}}}}\\
    \sum_{m=1}^{\lfloor c(t^D_{\max}-t'-k)\rfloor}\frac{\mu}{D_{\max}}e^{-\frac{\mu}{2D_{\max}}(2D_{\max}+ck+ck^2-ct^D_{\max}-2ckt^D_{\max}+c(t^D_{\max})^2+ct'+2ckt'-2ct^D_{\max}t'+ct'^2+c\lceil\frac{m}{c}\rceil-c\lceil\frac{m}{c}\rceil^2)} \bigg)
    >0.
\end{multline}
\normalsize
If we drop the ceiling function, the derivative of $y_1(b')$ is 
\begin{equation}
    y_1'(b')=\frac{\mu}{2cD_{\max}}(c-2b')e^{-\frac{\mu(ct^D_{\max}-ct'xb')(-c+ct^D_{\max}-ct'+b')}{2cD_{\max}}}.
\end{equation}
The critical number is $b'=\frac{c}{2}>0$. It means $y_1(b')$ is increasing on $(0,\frac{c}{2})$ and decreasing on $(\frac{c}{2},\infty)$. Given $y_1(0)>y_2(0)$, we can always find the $b^*$ at which $y_1(b')$ and $y_2(b')$ intersect. 

Furthermore, because of the increasing and decreasing intervals, we have $y_1(b')\geq y_2(b')$ for $0<b'\leq b^*$, and $y_1(b')< y_2(b')$ for $b^*<b'< \infty$. If $y_1(b')$ is greater, it means the probability of achieving 1 immediate reward and at least 1 future reward is higher than that of achieving at least 2 future rewards. In such case, the dispatcher should always accept the request and assign it to the drone. Otherwise, the dispatcher should not provide service for the customer because the probability of having at least 2 future rewards is higher. 
\end{proof}

\begin{figure}[h!]
	\begin{minipage}[b]{0.32\linewidth} 
		\centering
		\includegraphics[width=.8\linewidth]{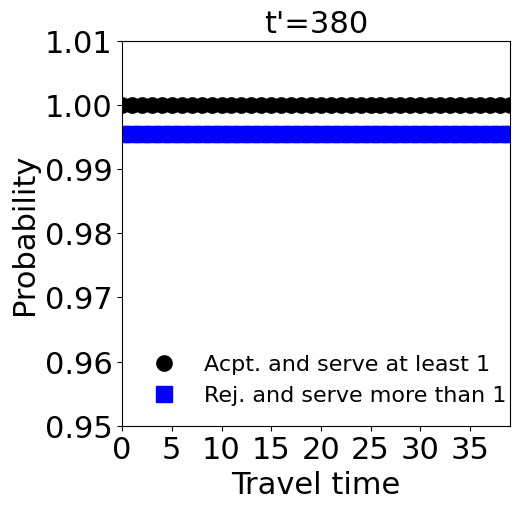}
	\end{minipage} 
	\begin{minipage}[b]{0.32\linewidth}
		\centering
		\includegraphics[width=.8\linewidth]{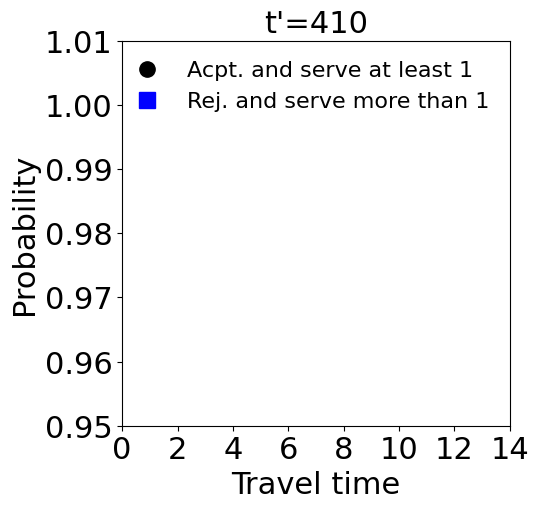}
	\end{minipage} 	
	\begin{minipage}[b]{0.32\linewidth}
		\centering
		\includegraphics[width=.8\linewidth]{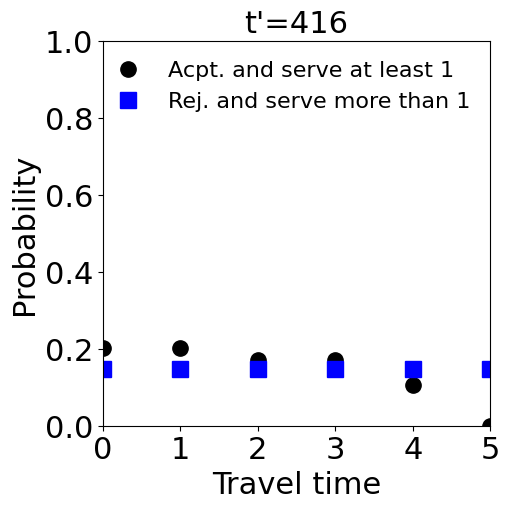}
	\end{minipage}
	\caption{The probabilities vs. the distance of the customer request}
	\label{lemmas}
\end{figure}

For three values of $t'$, Figure~\ref{lemmas} illustrates Proposition \ref{rej_prop} for parameter settings $c=1.5$, $\mu=1$, $D_{\max}=40$, and $t^D_{\max}=420$. The horizontal axis represents the possible distance of the new feasible customer, and the vertical axis is the probability. The probability in Lemma \ref{prob} (black dots in Figure \ref{lemmas}) can be seen as that of achieving $1$ immediate reward and at least $1$ future reward. The probability in Lemma \ref{rej_prob} (blue dots) can be seen as that of achieving at least $2$ future rewards. 

As shown in the left-hand figure, given these parameter settings, there is a time before the horizon at which we always accept the customer. When it is long enough before the end of the horizon, $b^*$ is much greater than $D_{\max}$ so the two probabilities do not intersect in the plot. The probability that the drone can serve at least one more customer after serving the current one is always higher than the other one. In this case, the dispatcher should always take the immediate reward. As it is closer to the end of horizon, $t'=410$, the probability of achieving a future reward of $2$ goes down. The two probabilities intersect at about $b^*=7.5$. In this situation, the dispatcher should accept the feasible request if $b'$ is between $0$ and approximately $7.5$. Otherwise, the dispatcher should consider not accepting the request, because as larger $b'$ becomes larger, the chance that we cannot serve any more customer thereafter grows. At $t'=416$ when it is even closer to the end of the horizon, both probabilities are relatively low. In this situation, we observe the value of $b^*$ is smaller than that for $t'=410$. As shown in this illustrative example, the threshold concerning denial decisions should incorporate the time in a shift. 

In the case the drone is unavailable for the rest of the day, and the vehicle is idling at the depot, we can derive results analogous to Proposition~\ref{rej_prop}. In the case of the vehicle though, we need to consider the cost of inserting a customer in the vehicle's route rather than just the travel time from the depot. We denote this cost as $\Delta\textsubscript{Vehicle}$, the increase in the vehicle's tour time that results from inserting the customer to the vehicle's planned route. Due to their different capacities, the times at which the drone and the vehicle are planned to return to the depot are determined by different quantities. The drone has a capacity one so it must return to the depot after every delivery. The vehicle can serve multiple customers in a route. If the dispatcher assigns a new request to the vehicle, the time at which it is planned to return is postponed by $\Delta\textsubscript{Vehicle}$, the insertion cost of this new request. We wanted to develop the proposition for the vehicle similar to Proposition \ref{rej_prop}. However, due to the unlimited capacity of the vehicle, we expect it is far more complicated and thus not practical to do so. Because of the inter-dependencies of decisions, we were not able to provide a straightforward proof.


Although the SDDPVD is way more complicated than the simplified version, we can still use the similar logic for the analysis of the SDDPVD. For example, when the delivery resources become limited, the dispatcher should consider offering no service to some feasible requests in exchange for a higher expected reward in the future. We take the logic in this analysis as motivation to select the features for the SDDPVD.

\subsection{Illustration of the Decision Making}\label{appendix_other_benchmark}

In this section, to gain insight into why $\pi\textsuperscript{Q}$ outperforms $\pi\textsuperscript{PFA}$, we graphically illustrate the differences in the policies $\pi\textss{PFA}$ and $\pi\textss{Q}$. We first present the results for the homogeneous customer demand and then for the temporally heterogeneous demand because it allows a very clear distinction between the two policies, followed by that for the spatially heterogeneous demand.

\subsubsection*{On Homogeneous Customer Demand}
To demonstrate these differences, we first select a single instance realization (a day) on which different policies are evaluated. Then, for each policy, we plot the acceptance and assignment decisions of each customer throughout the day. We consider instances with a fleet of $3$ vehicles and $10$ drones. We first assume homogeneous demand. For this selected instance, policies serve slightly more customers than on average with policy $\pi\textss{PFA}$ serving 394 and $\pi\textss{Q}$ serving 444 customers.

\begin{figure}[h]
	\centering
	\includegraphics[width=0.75\textwidth]{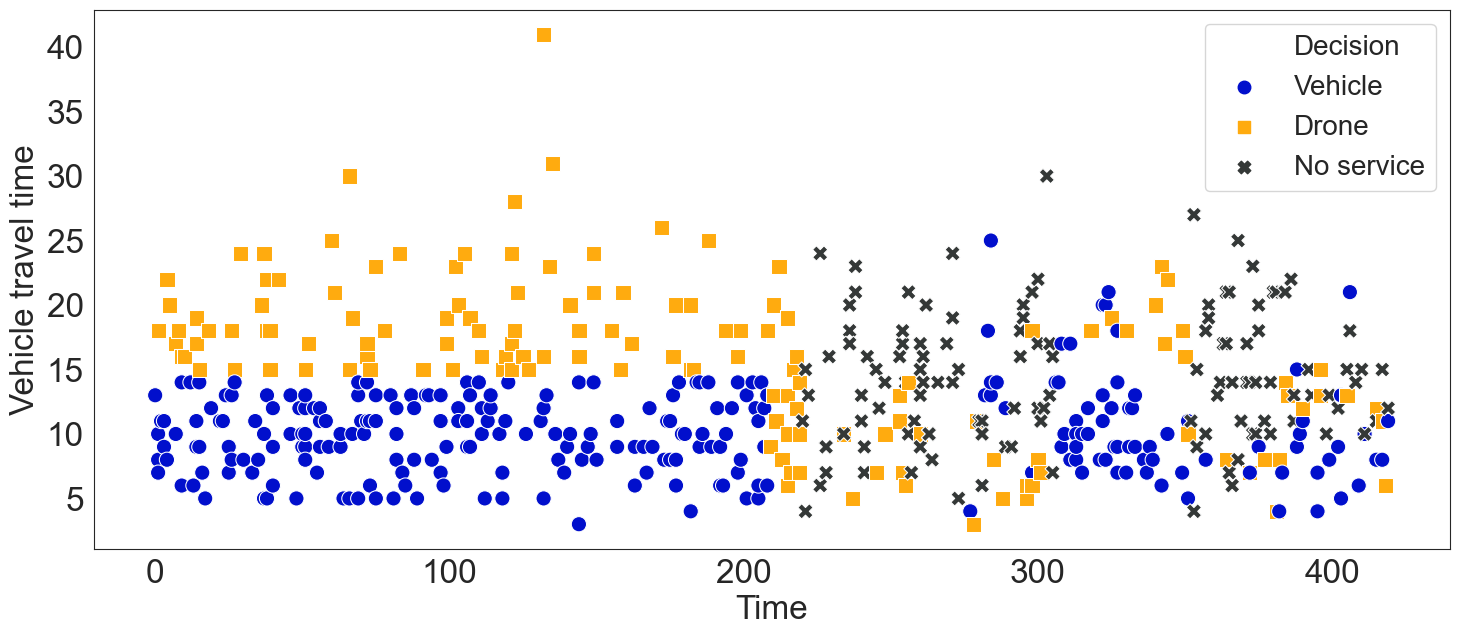}
	\caption{Illustration of decision making under $\pi^{\textrm{PFA}}$. Assignment decisions over time and customer distance for the selected homogeneous instance realization.}
	\label{pfa}
\end{figure}

\begin{figure}[h]
	\centering
	\includegraphics[width=0.75\textwidth]{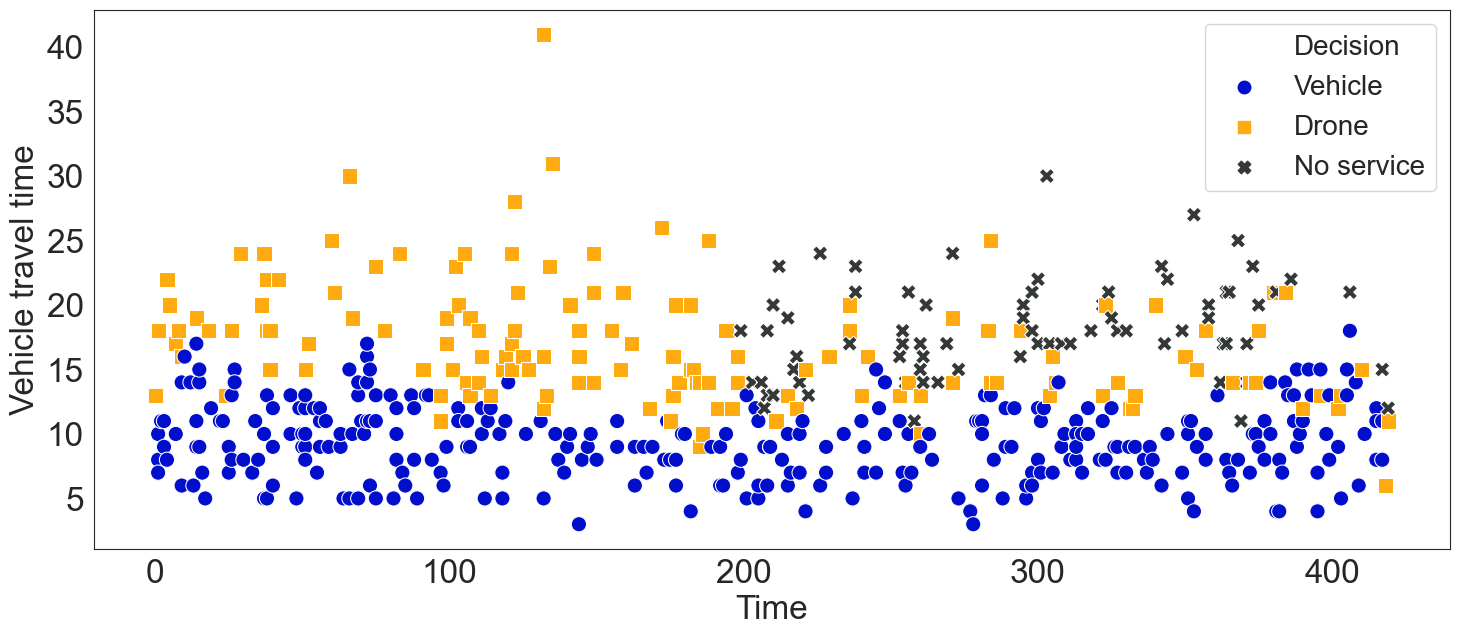}
	\caption{Illustration of decision making under $\pi^{\textrm{Q}}$. Assignment decisions over time and customer distance for the selected homogeneous instance realization.}
	\label{dql}
\end{figure}

Figures~\ref{pfa} and \ref{dql} illustrate the served customers and how they are served. The horizontal axis is the time in minutes ranging from $0$ to the latest possible order time $420$, and the vertical axis represents the vehicle travel time needed by a vehicle to service a given request where the travel time is based on the customer's distance from the depot. Each dot in the figures represents a customer order whose coordinates on the plot are determined by when and where they make the request and whose color is depends on the decision made by the corresponding policy, service by vehicle, drone, or no service.

Figure~\ref{pfa} presents the decisions for the policy $\pi\textss{PFA}$. This policy uses the vehicle travel time threshold $\tau=14$ to choose between vehicles and drones. The figure shows that, in the interval $[0,200]$, both vehicles and drones can feasibly serve customers. However, starting around $t=200$, vehicle capacity becomes limited because of existing assignments, and vehicles are unable to meet the delivery deadlines of new requests. Thus, the policy $\pi\textss{PFA}$ starts to assign customers close to the depot to drones. This result follows from the fact that $\pi\textss{PFA}$ will assign customers that cannot be served by vehicles to the drone fleet if such an assignment is feasible. As a result, the vehicle travel time threshold vanishes over the second half of the day, and both drone and vehicle capacity becomes limited. Eventually, a relatively large number of customers are left unserved. 

Intuitively, a good policy serves as many closer customers as possible because the cost of traveling to them is relatively low. Yet,  consider customer $(221,4)$ that is close to the depot and orders in the middle of the day.  This customer does not get served because both drone and vehicle capacity is consumed. This occurs because customers, like that customer $(278,3)$ who is close to the depot, are and assigned to a drone. Such an assignment does not make an efficient use of the drone because the relatively long setup and charging times outweighs the travel speed advantage for customers close to the depot. This example illustrates the shortcomings of the policy $\pi\textss{PFA}$ serving customers whenever it is feasible for any fleet type. 

Figure~\ref{dql} illustrates the decision making of policy $\pi\textss{Q}$. In the selected instance, $\pi\textss{Q}$ demonstrates significant improvements over all the benchmarks, ranging from $12.1\%$ (over $\pi\textss{PFA\_rej}$) to $12.7\%$ (over $\pi\textss{PFA}$). Although $\pi\textss{Q}$ does not operate on any kind of threshold, Figure~\ref{dql} indicates an emergent time-dependent threshold that results from the learned Q-values. In the beginning of the day when the delivery resources are sufficient, $\pi\textss{Q}$ assigns most customers within the threshold (about $14$ minutes) to vehicles and distant customers to drones. During about $[100,200]$, when vehicle capacity is mostly consumed, $\pi\textss{Q}$ shows a slightly diminishing threshold, maintaining the availability of vehicles. Unlike the benchmarks, even when drones can feasibly serve closer customers during $[100,200]$,
$\pi\textss{Q}$ does not assign them to drones. In fact, it is true for the whole day except customer $(418,6)$ at the very end of the day. This exception can be explained by Lemma~\ref{rej_prob}. Because the time of the decision point is one of the features, $\pi\textss{Q}$ recognizes that this request is made nearly at the end of the shift. Rather than offering no service to the customer, $\pi\textss{Q}$ takes the immediate reward and assigns it to a drone because the probability of serving at least $2$ customers in the future if not serving the current one is relatively low. The policy $\pi\textss{Q}$ self-learns the policy without any explicit knowledge of the lemmas and propositions.

\begin{figure}[h]
	\centering
	\includegraphics[width=0.7\textwidth]{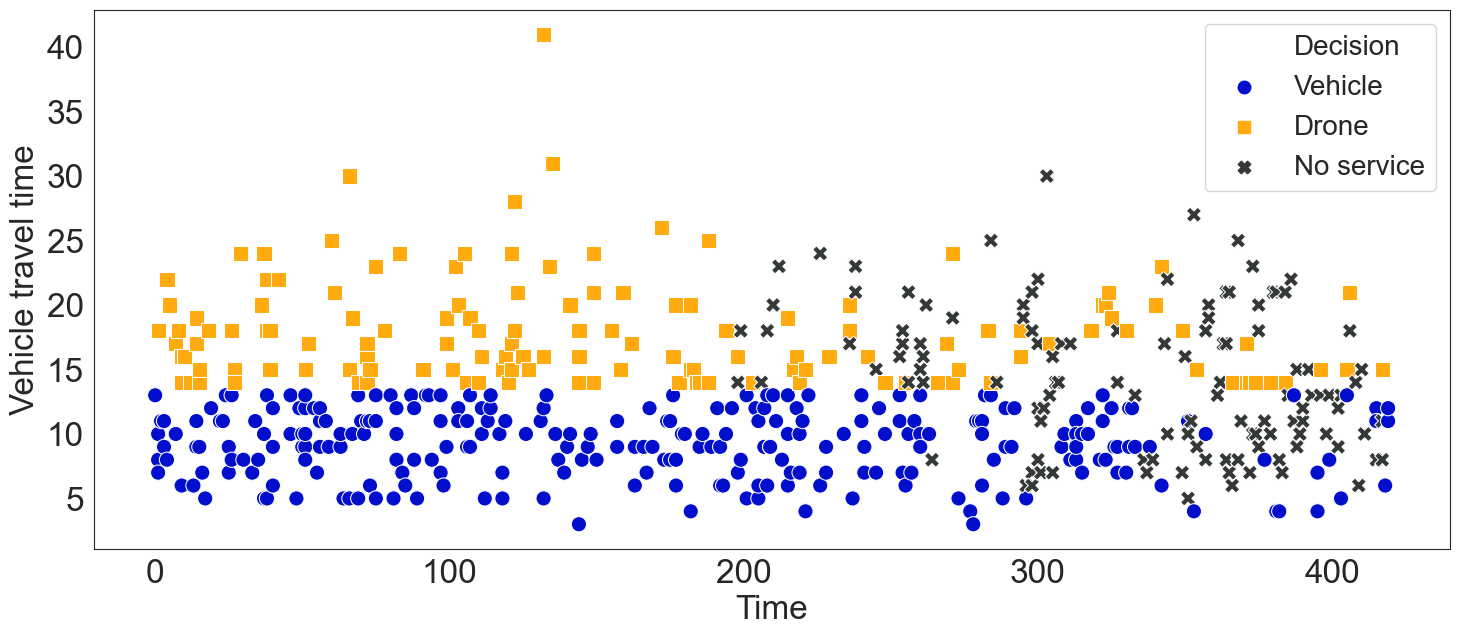}
	\caption{Time vs. travel time vs. decision under $\pi^{\textrm{PFA\_rej}}$ on the selected homogeneous instance.}
	\label{pfarej}
\end{figure}

Figure~\ref{pfarej} presents an illustration of the decision making of policy $\pi\textss{PFA\_rej}$. In this case, the policy has a hard threshold $\tau=13$. Because the policy strictly obeys the threshold, we observe an explicit horizontal line throughout the day. In rigidly maintaining the threshold, we also see that the policy does not provide service to a number of customers over the second half of the day. Yet, in determining which customers do not receive service in a more controlled way than the policy $\pi\textss{PFA}$, the policy $\pi\textss{PFA\_rej}$ serves $2$ more customers than $\pi\textss{PFA}$ in this selected instance and even about $6.5\%$ on average for the instance setting.

\begin{figure}[h]
	\centering
	\includegraphics[width=0.7\textwidth]{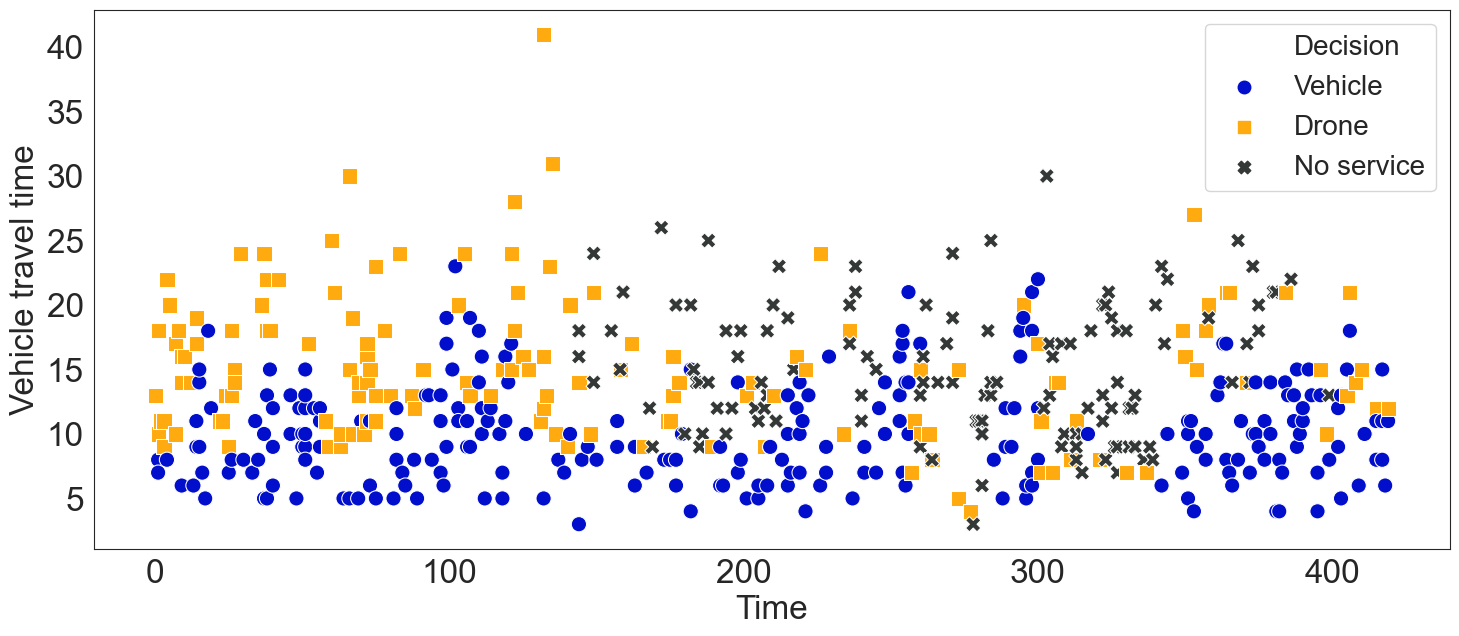}
	\caption{Time vs. travel time vs. decision under $\pi^{\textrm{Delta}}$ on the selected homogeneous instance.}
	\label{delta}
\end{figure}

Figure~\ref{delta} presents an illustration of the decision making of $\pi\textss{Delta}$. Because $\pi\textss{Delta}$ decides whether to offer service to a customer on the insertion cost, no threshold is visible. While the policy performs well relative to the other benchmarks, we can see times when the rule-based decision making leads to less desirable decisions. Consider the assignments near time 100. At this point, the vehicles end up serving customers that are relatively far from the depot. The threshold-based policies would have controlled the farther away customers from being added to the routes and thus less efficient use of the vehicles. The result is that a series of relatively closer customers are assigned to drones in the time interval $[100, 200]$. Then, just after time 200, a number of requests are denied service because both vehicle and drone capacity have been consumed.

\begin{figure}[h!]
	\centering
	\includegraphics[width=0.7\textwidth]{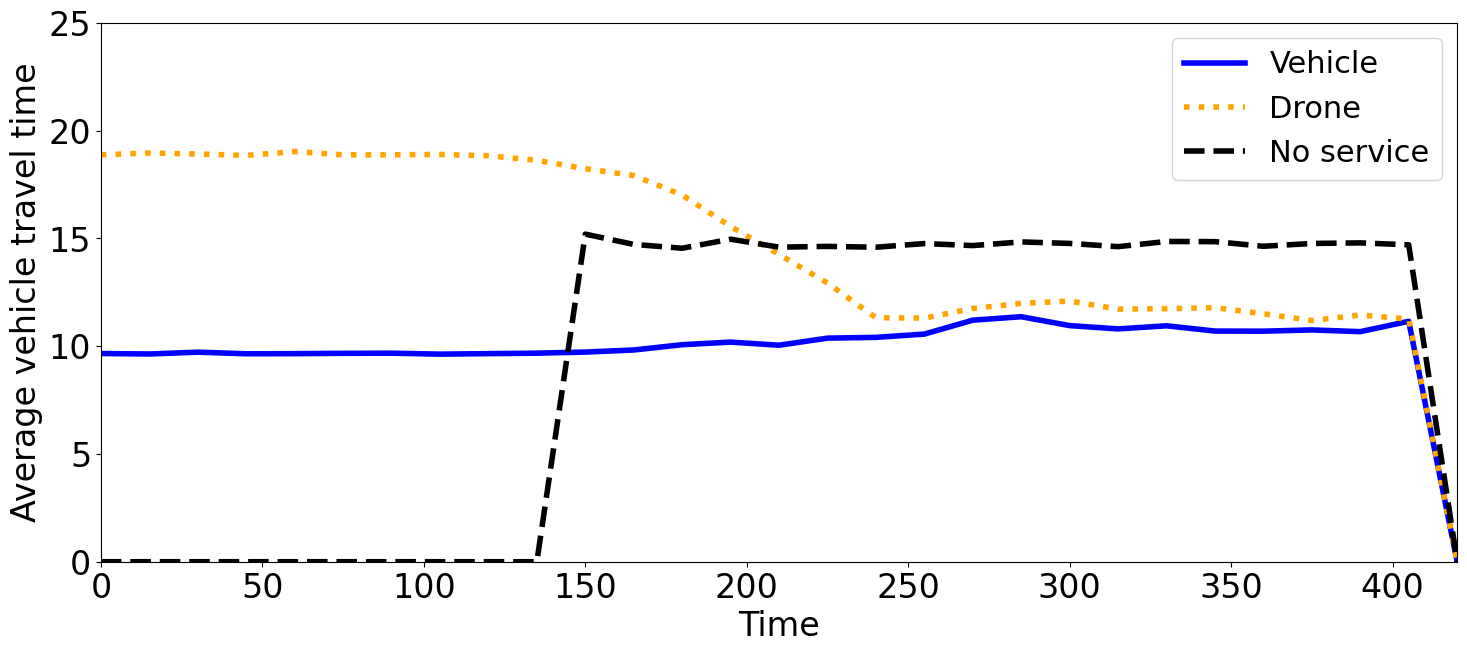}
	\caption{Illustration of decision making under $\pi^{\textrm{PFA}}$. Average assignment decisions over time and customer distance over all homogeneous instances.}
	\label{pfa_quantify}
\end{figure}

\begin{figure}[h!]
	\centering
	\includegraphics[width=0.7\textwidth]{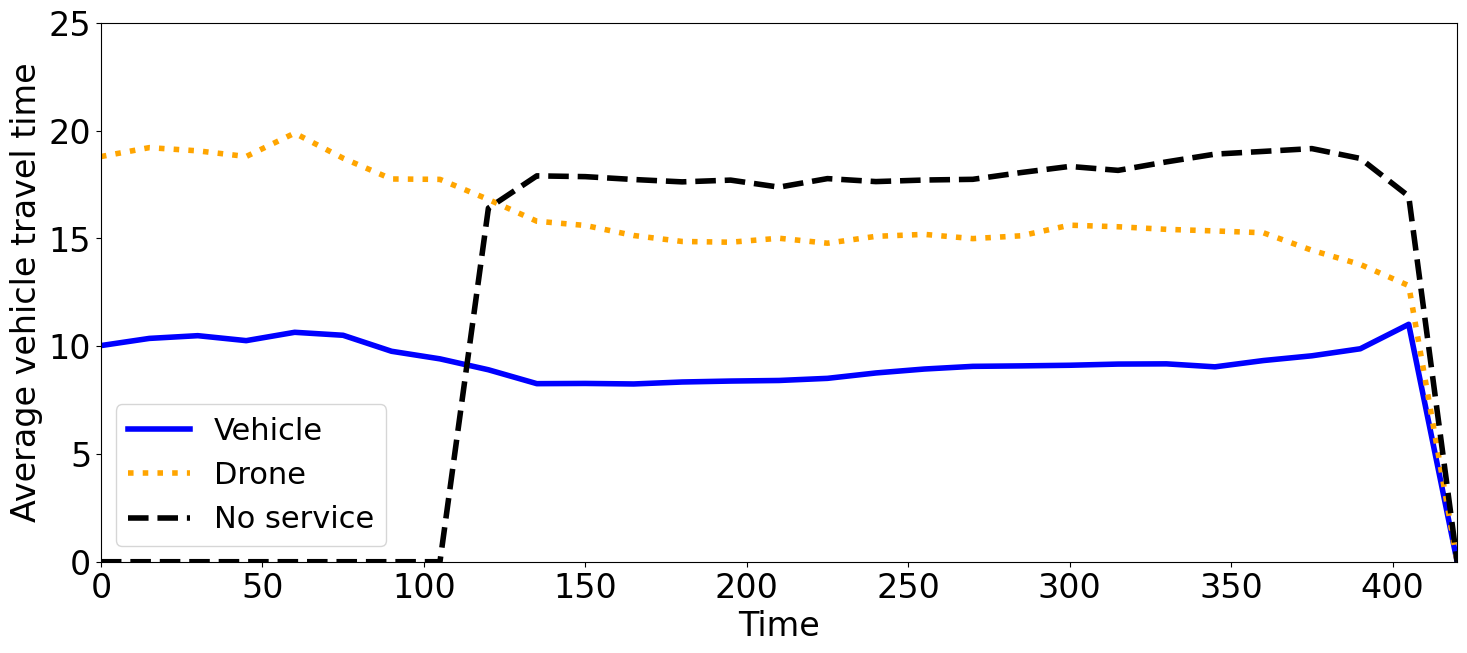}
	\caption{Illustration of decision making under $\pi^{\textrm{Q}}$. Average assignment decisions over time and customer distance over all homogeneous instances.}
	\label{dql_quantify}
\end{figure}

We then extend this anecdotal analysis to the full set of instance realizations. For each policy and assignment decision, we calculate the average distance of the corresponding customer for time intervals of 15 minutes. An average value of zero indicates that no customers received the corresponding decision. The results for $\pi^{\textrm{PFA}}$ and $\pi^{\textrm{Q}}$ are depicted in Figures~\ref{pfa_quantify} and \ref{dql_quantify}. We note that the figures do not depict customer quantities. 
For both policies, we observe similar patterns in the beginning. Customers assigned to drones are further away from the depot than customers assigned to vehicles. Further, all customers are offered service. Later, the two policies differ. For the PFA, the first customers that cannot be served occur around time 150. Furthermore, after that time, the distances for customers assigned to drones and to vehicles even out. This confirms the observation in Figure~\ref{pfa} that for the second half of the day, no clear assignment pattern can be observed.

For policy $\pi\textsuperscript{Q}$, the first customers where no service is offered occur earlier compared to $\pi\textss{PFA}$. Further, they are more distant. This indicates that while the customers might be feasible to serve, policy $\pi\textsuperscript{Q}$ starts budgeting resources to keep the fleet's flexibility. As a result, the clear and effective structure of drone assignment to distant customers and vehicle assignment to closer customers can be maintained throughout the day.

\begin{figure}[h]
	\centering
	\includegraphics[width=0.7\textwidth]{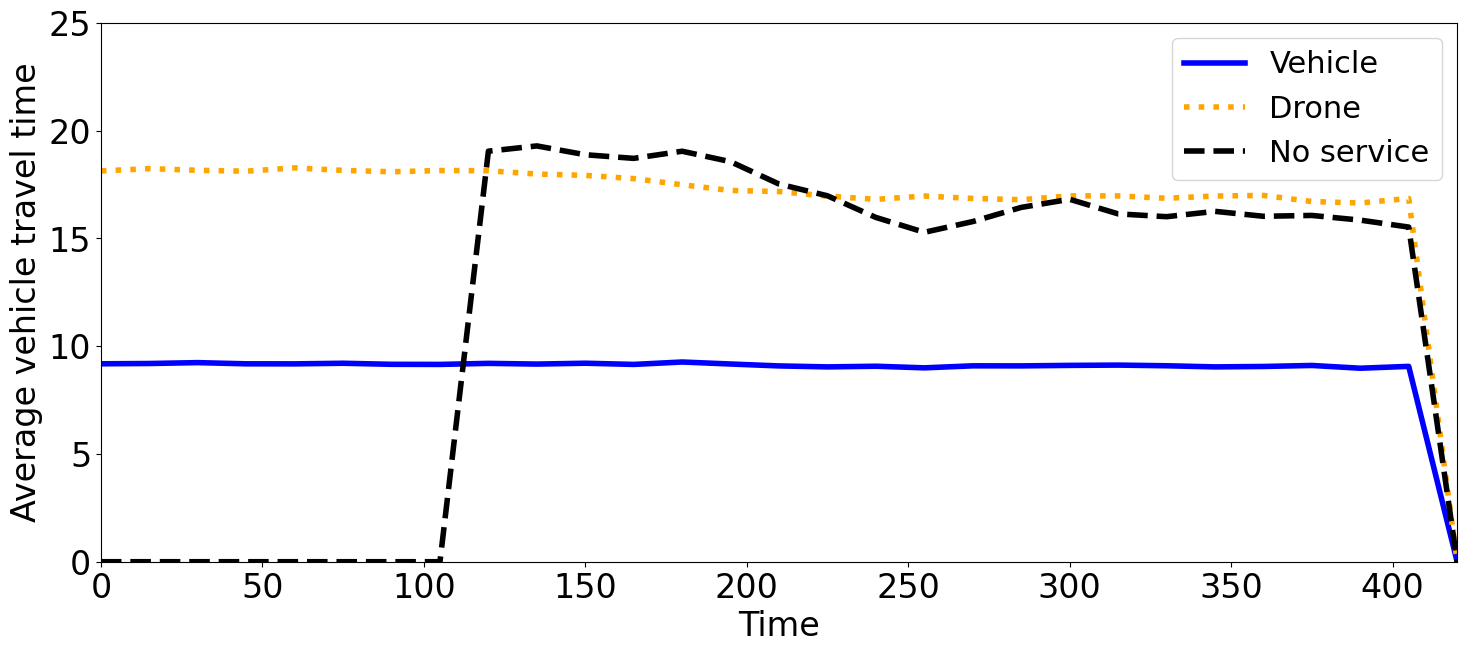}
	\caption{Average distance of requests vs. assignment decisions under $\pi\textsuperscript{PFA\_rej}$ for homogeneous demand.}
	\label{pfa_with_rej_quantify}
\end{figure}

\begin{figure}[h]
	\centering
	\includegraphics[width=0.7\textwidth]{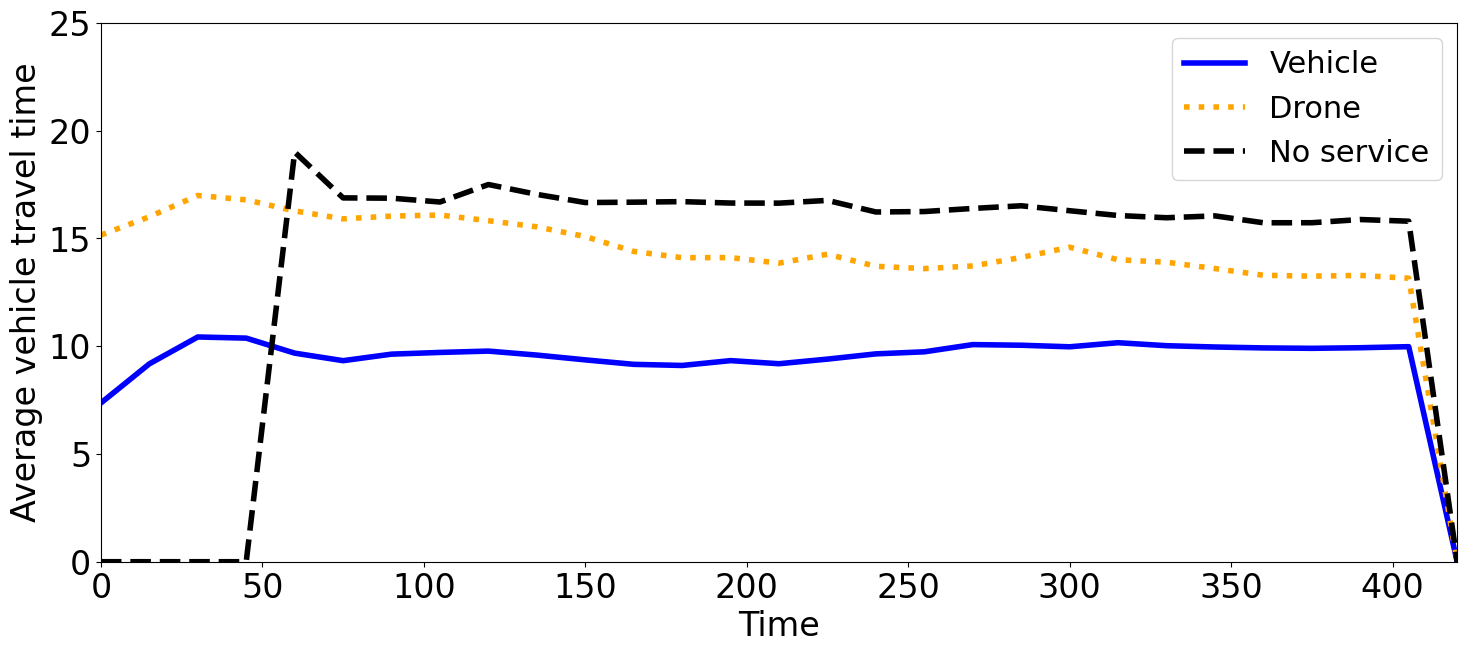}
	\caption{Average distance of requests vs. assignment decisions under $\pi\textsuperscript{Delta}$ for homogeneous demand.}
	\label{delta_quantify}
\end{figure}

Figures \ref{pfa_with_rej_quantify} and \ref{delta_quantify} present the average distance of customer requests with respect to different assignment decisions under $\pi\textsuperscript{PFA\_rej}$ and $\pi\textsuperscript{Delta}$. 
Because it does not allow vehicles and drones to cross the threshold, $\pi\textsuperscript{PFA\_rej}$ shows a relatively consistent average distance for decisions \textit{vehicle} and \textit{drone} throughout the day. It starts to offer \textit{no service} to some customers shortly after $t=100$. The average distance for \textit{no service} decreases as the resources become more constrained thereafter. For $\pi\textsuperscript{Delta}$, the average distance for \textit{drone} decreases throughout the day, while that for \textit{vehicle} is relatively consistent. Among all the policies considered, $\pi\textsuperscript{Delta}$ is the earliest to offer \textit{no service} to some customers. One explanation is that the policy carefully assigns requests to vehicles but does not consider the efficiency of drones, and thus uses up the overall resources relatively faster.

\subsubsection*{On Temporally Heterogeneous Customer Demand}

For the selected single instance with temporally heterogeneous demand, $\pi\textss{PFA}$ serves $324$ customers, while $\pi\textss{PFA\_rej}$ serves $371$, $\pi\textss{Delta}$ serves $358$, 
and $\pi\textss{Q}$ serves $361$ customers. 

\begin{figure}[h!]
	\centering
	\includegraphics[width=0.7\textwidth]{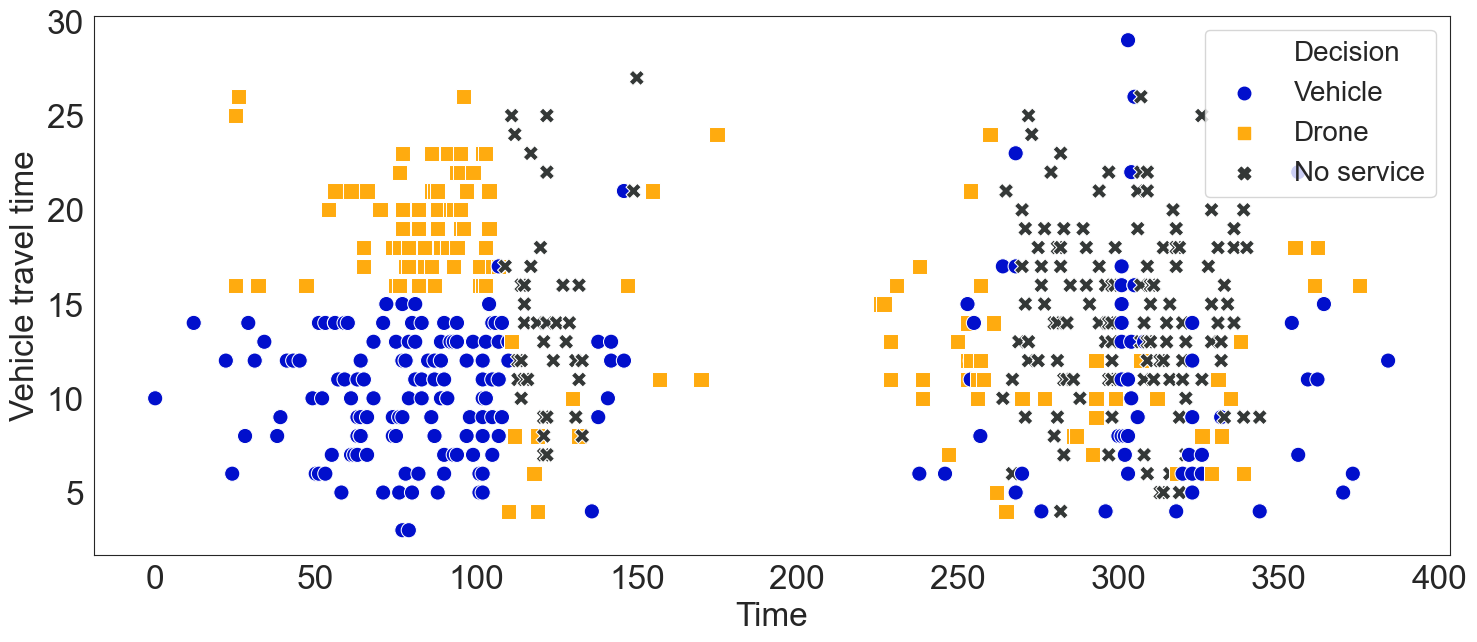}
	\caption{Time vs. travel time vs. decision under $\pi^{\textrm{PFA}}$ on the selected temporally heterogeneous instance.}
	\label{T_pfa}
\end{figure}

\begin{figure}[h!]
	\centering
	\includegraphics[width=0.7\textwidth]{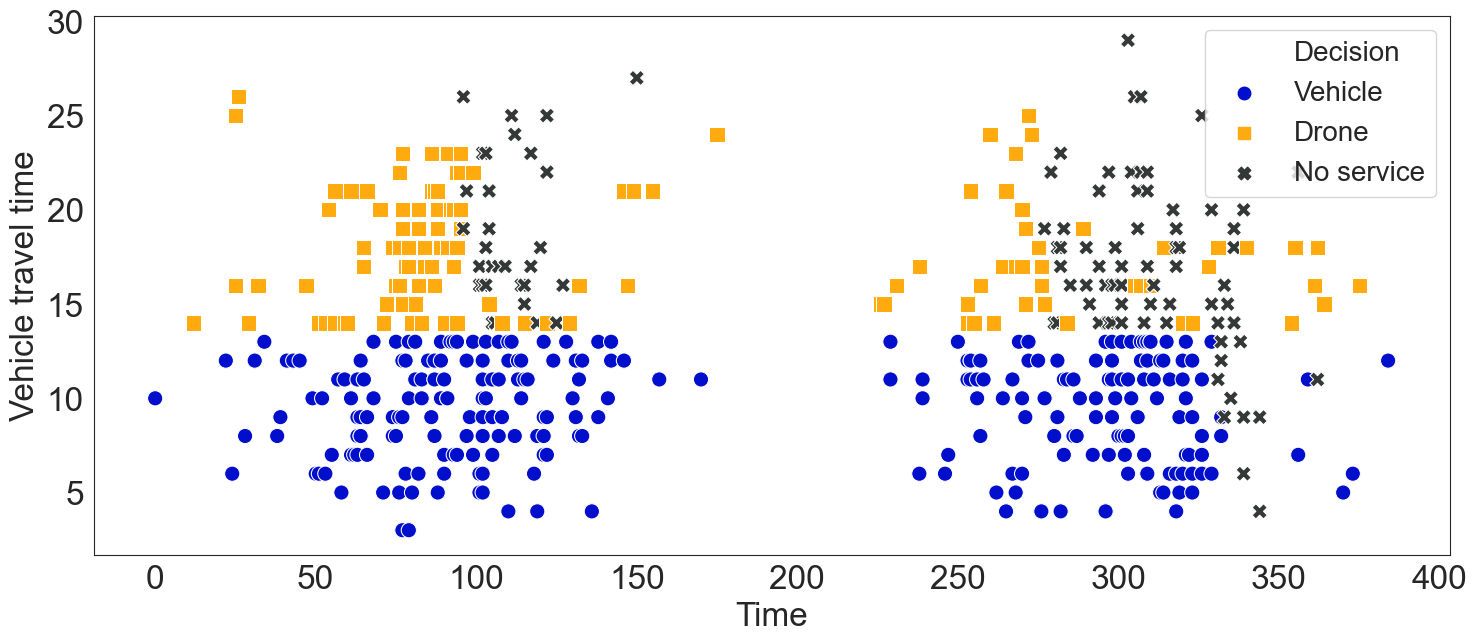}
	\caption{Time vs. travel time vs. decision under $\pi^{\textrm{PFA\_rej}}$ on the selected temporally heterogeneous instance.}
	\label{T_pfa_rej}
\end{figure}

\begin{figure}[h!]
	\centering
	\includegraphics[width=0.7\textwidth]{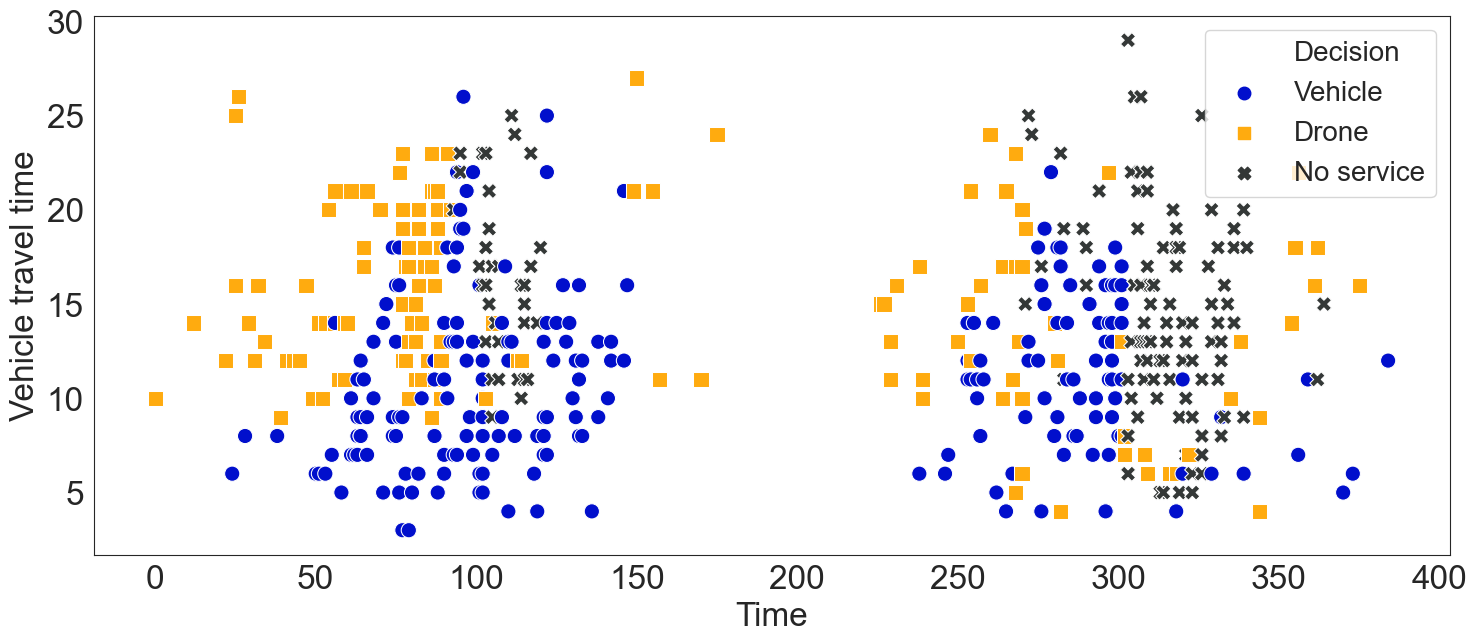}
	\caption{Time vs. travel time vs. decision under $\pi^{\textrm{Delta}}$ on the selected temporally heterogeneous instance.}
	\label{T_delta}
\end{figure}

\begin{figure}[h!]
	\centering
	\includegraphics[width=0.7\textwidth]{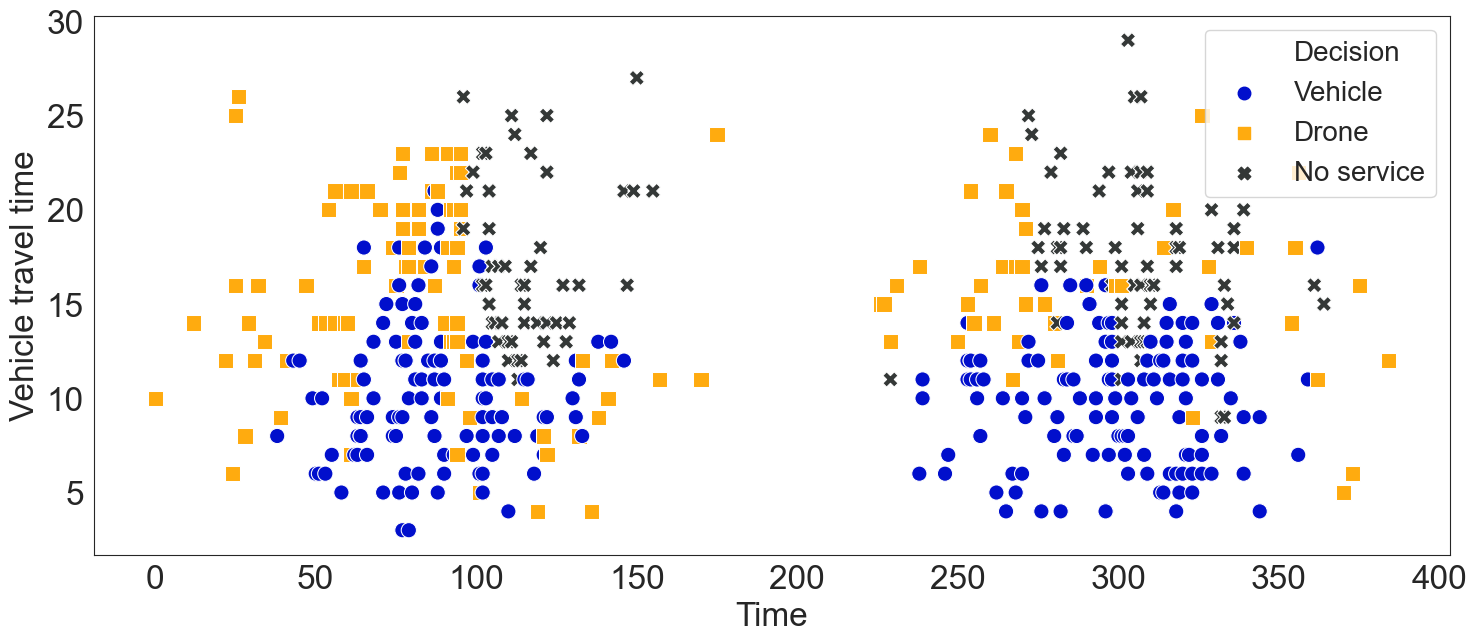}
	\caption{Time vs. travel time vs. decision under $\pi^{\textrm{Q}}$ on the selected temporally heterogeneous instance.}
	\label{T_dql}
\end{figure}

On the selected single instance, all the policies start to offer no service to some customers at the end of the first peak, and many are not offered service during the second peak. A notable difference between $\pi^{\textrm{Q}}$ and the others is that, in the early stage, $\pi^{\textrm{Q}}$ assigns most of the requests to drones, while the others uses both fleets. 

\begin{figure}[h!]
	\centering
	\includegraphics[width=0.7\textwidth]{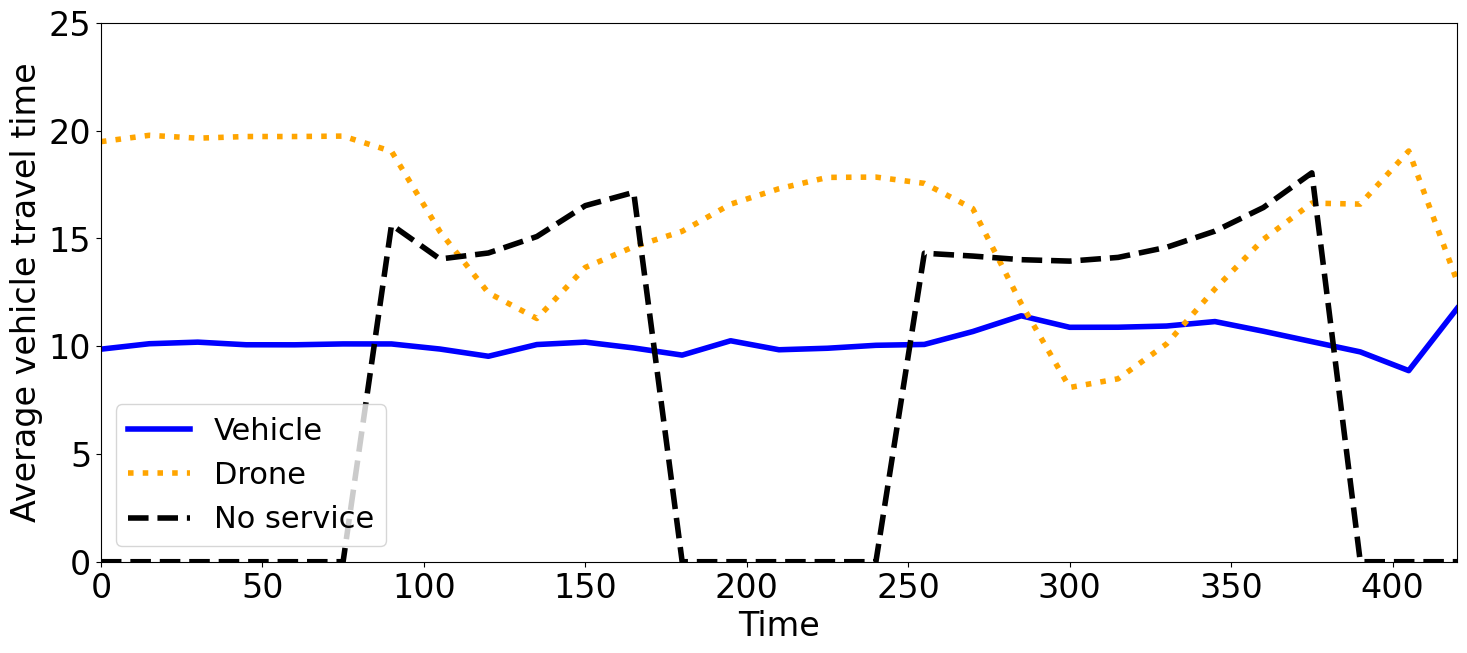}
	\caption{Illustration of decision making under $\pi^{\textrm{PFA}}$. Average assignment decisions over time and customer distance over all temporally heterogeneous instances.}
	\label{T_pfa_quantify}
\end{figure}

\begin{figure}[h!]
	\centering
	\includegraphics[width=0.7\textwidth]{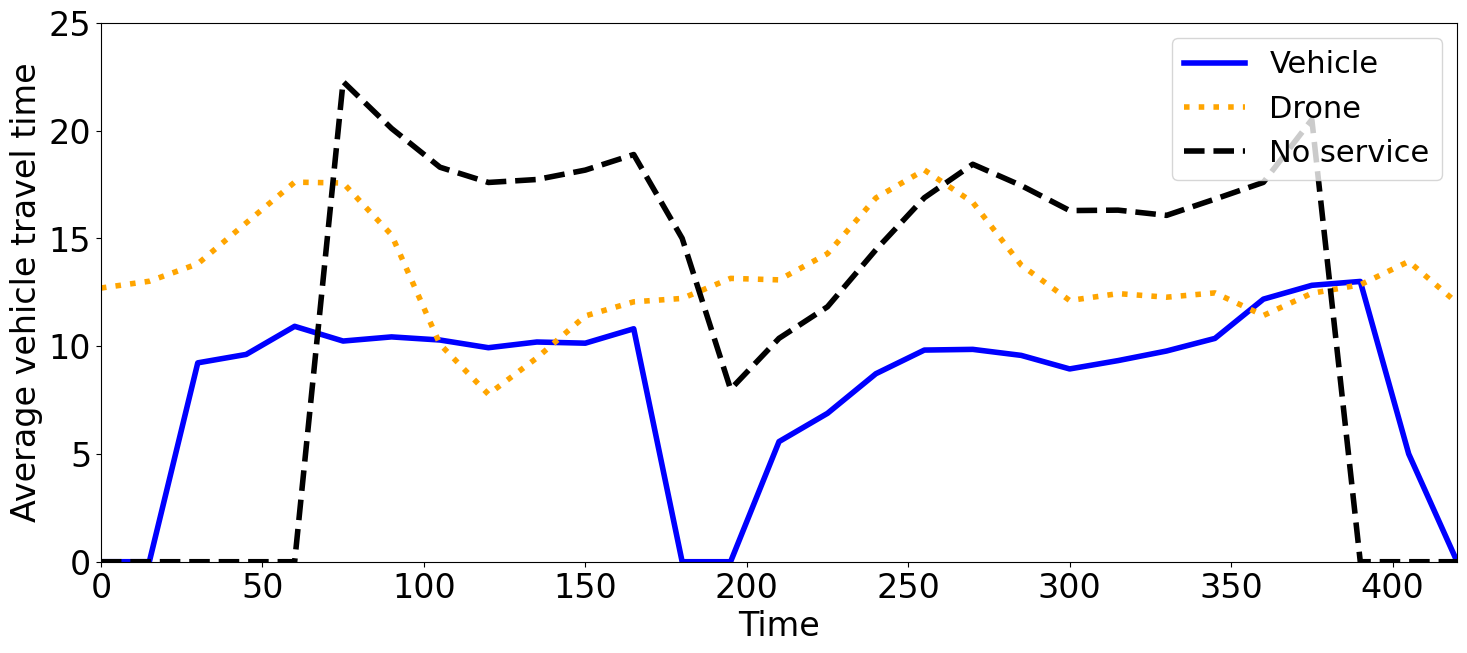}
	\caption{Illustration of decision making under $\pi^{\textrm{Q}}$. Average assignment decisions over time and customer distance over all temporally heterogeneous instances.}
	\label{T_dql_quantify}
\end{figure}

\begin{figure}[h!]
	\centering
	\includegraphics[width=0.7\textwidth]{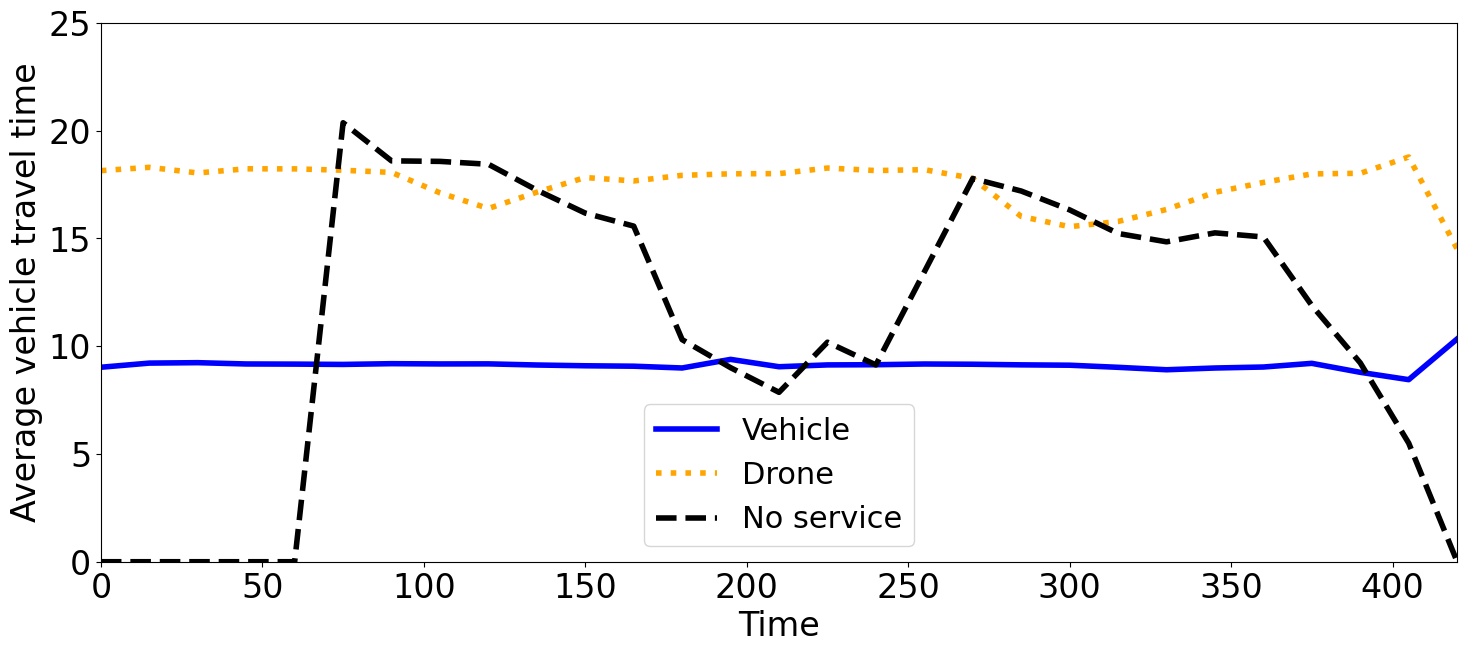}
	\caption{Average distance of requests vs. assignment decisions under $\pi\textsuperscript{PFA\_rej}$ for temporally heterogeneous demand.}
	\label{T_pfa_rej_quantify}
\end{figure}

\begin{figure}[h!]
	\centering
	\includegraphics[width=0.7\textwidth]{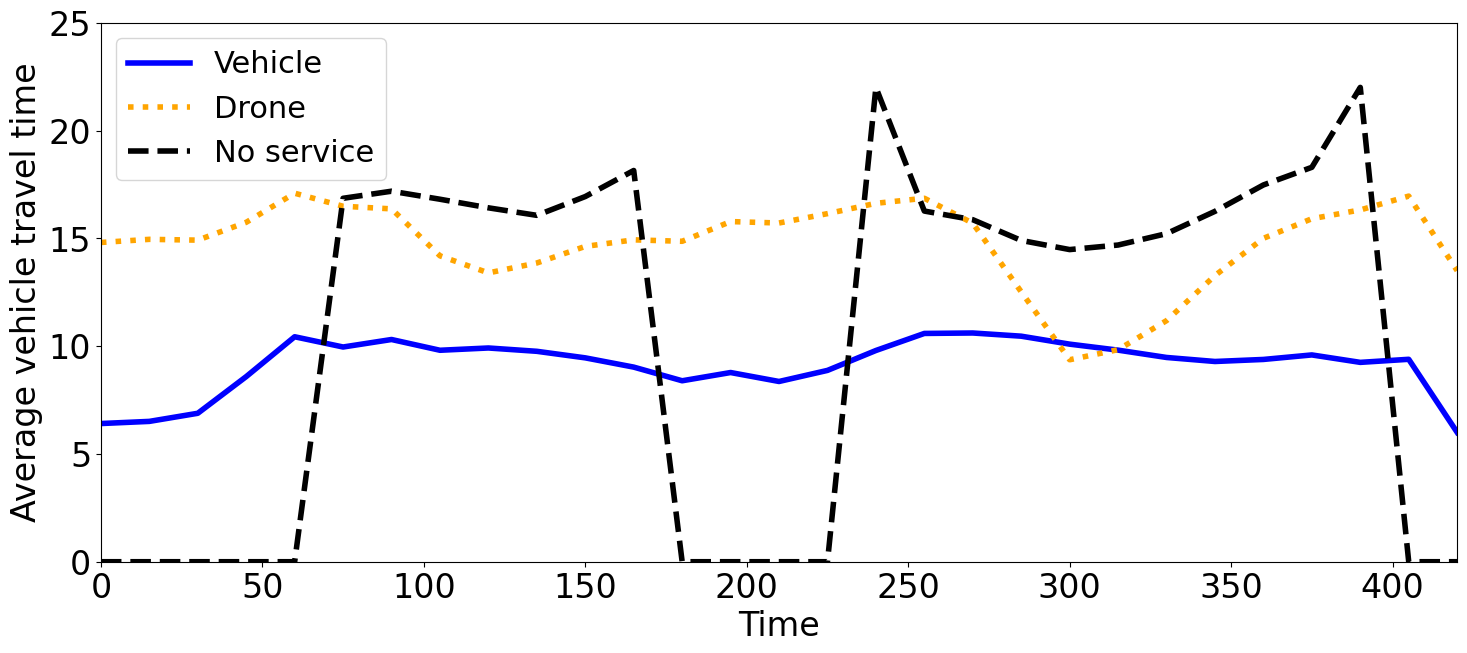}
	\caption{Average distance of requests vs. assignment decisions under $\pi\textsuperscript{Delta}$ for temporally heterogeneous demand.}
	\label{T_delta_quantify}
\end{figure}

We now analyze the results for the entire set of instances with temporally heterogeneous demand, again with 3 vehicles and 10 drones. Similar to Figures~\ref{pfa_quantify}-\ref{delta_quantify}, we depict the average distances over time in Figures~\ref{T_pfa_quantify}-\ref{T_delta_quantify}.
For the instances, two demand-peaks occur between times $50$ and $150$ as well as $250$ and $350$. We observe that the structure of $\pi\textss{PFA}$ only lasts up to the first peak, recovers between the peaks and is then disturbed again in the second peak. Shortly after beginning of the first peak and right when the second peak begins, customers are not offered service. The first peak consumes resources fast and at time the second peak starts, the resources have not fully recovered, especially, the vehicle fleet.

There are several differences for policy $\pi\textsuperscript{Q}$. As for the homogeneous instances, offering no service starts earlier than for $\pi\textss{PFA}$ and is also applied in the off-peak times, likely to free resources for the peak-times. However, we also observe that the vehicles are used differently. During off-peak times, at some times, no orders at all are assigned to vehicles, and if orders are assigned, they are close to the depot. Instead, off-peak orders are served by drones. This behavior indicates that $\pi\textsuperscript{Q}$ reserves vehicle resources for the peak-times where the consolidation potential of the vehicles become particularly valuable. That is, the advantage that the vehicles have for serving many customers on a single route becomes available as the customer density increases.

\subsubsection*{On Spatially Heterogeneous Customer Demand}

For the selected instance with spatially heterogeneous demand, policies serve slightly more customers than those with homogeneous demand on average with policy $\pi\textss{PFA}$ serving 447, $\pi\textss{PFA\_rej}$ serving 437, $\pi\textss{Delta}$ serving 447, 
and $\pi\textss{Q}$ serving 476 customers. 

\begin{figure}[h!]
	\centering
	\includegraphics[width=0.7\textwidth]{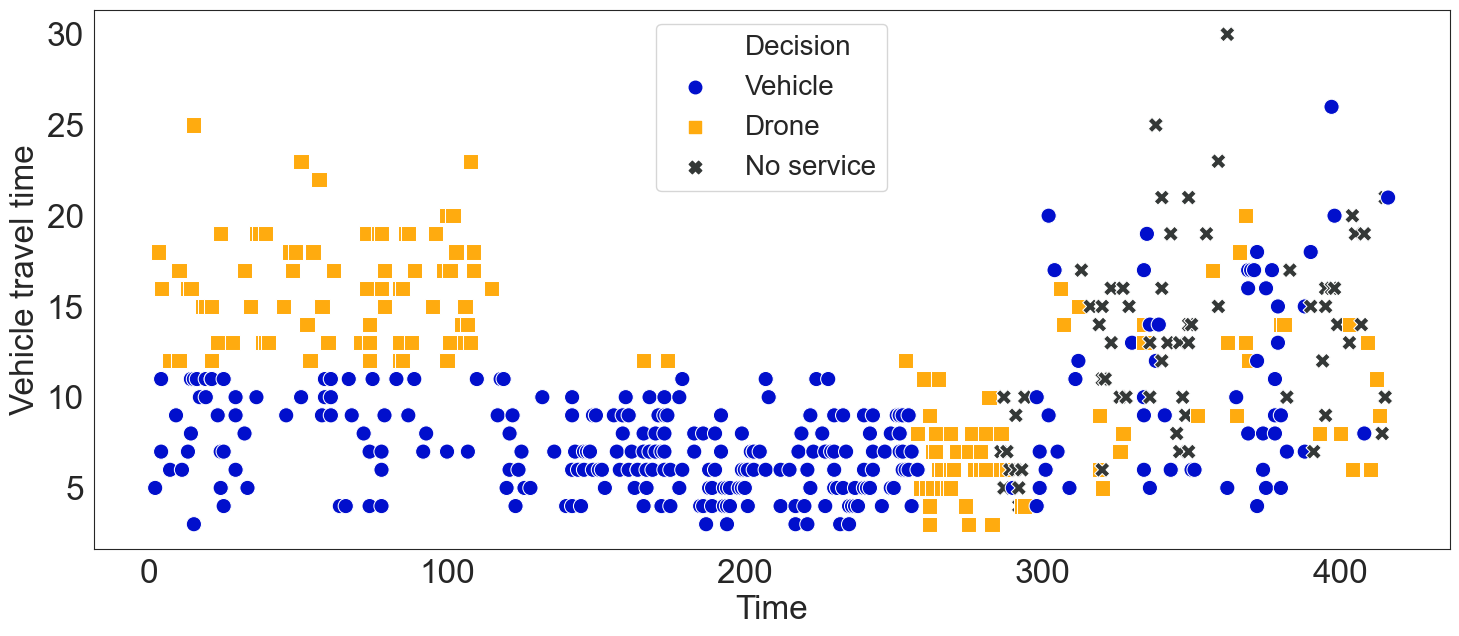}
	\caption{Time vs. distance vs. decision under $\pi^{\textrm{PFA}}$ on the selected spatially heterogeneous instance.}
	\label{H_pfa}
\end{figure}

\begin{figure}[h!]
	\centering
	\includegraphics[width=0.7\textwidth]{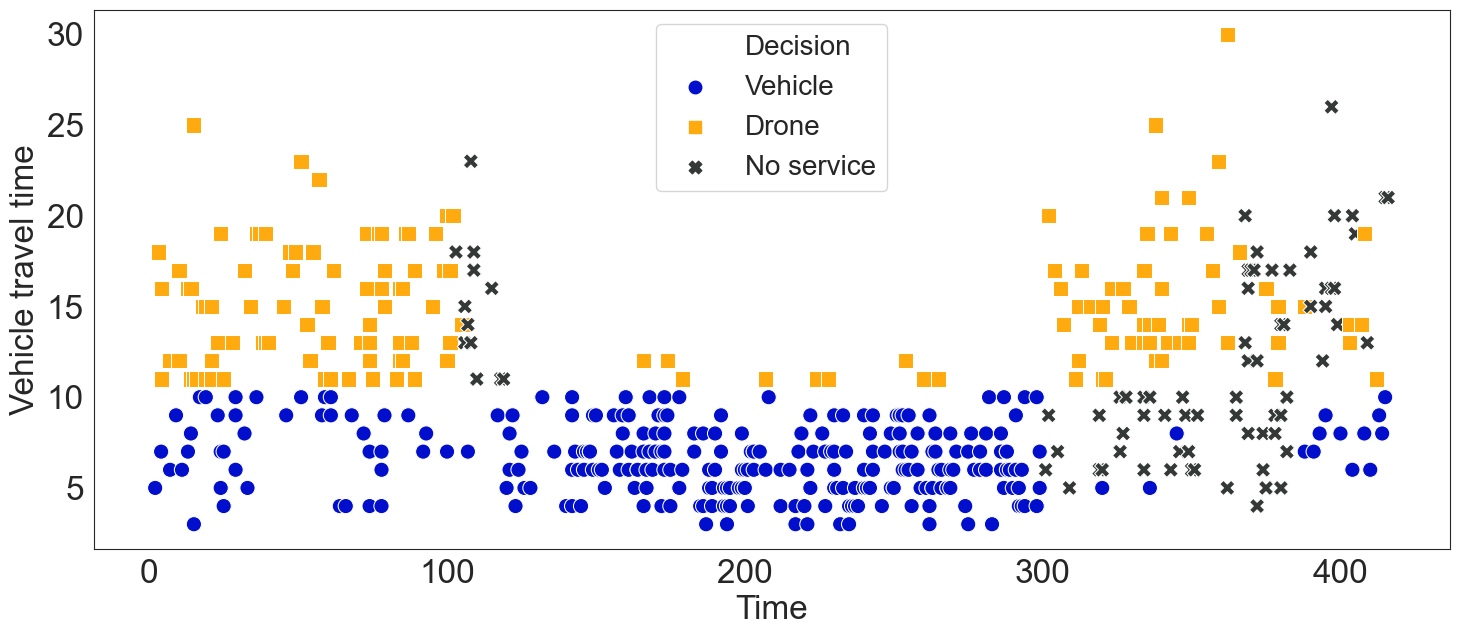}
	\caption{Time vs. distance vs. decision under $\pi^{\textrm{PFA\_rej}}$ on the selected spatially heterogeneous instance.}
	\label{H_pfa_with}
\end{figure}

\begin{figure}[h!]
	\centering
	\includegraphics[width=0.7\textwidth]{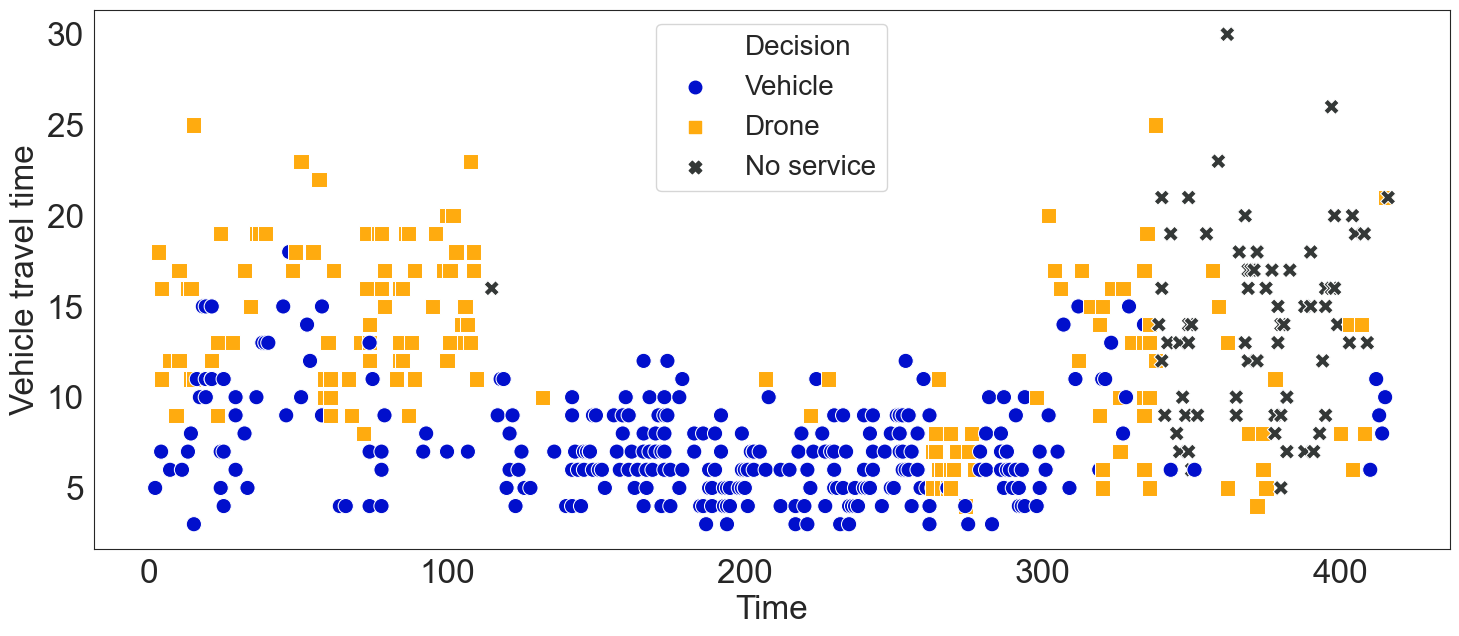}
	\caption{Time vs. distance vs. decision under $\pi^{\textrm{Delta}}$ on the selected spatially heterogeneous instance.}
	\label{H_delta}
\end{figure}

\begin{figure}[h!]
	\centering
	\includegraphics[width=0.7\textwidth]{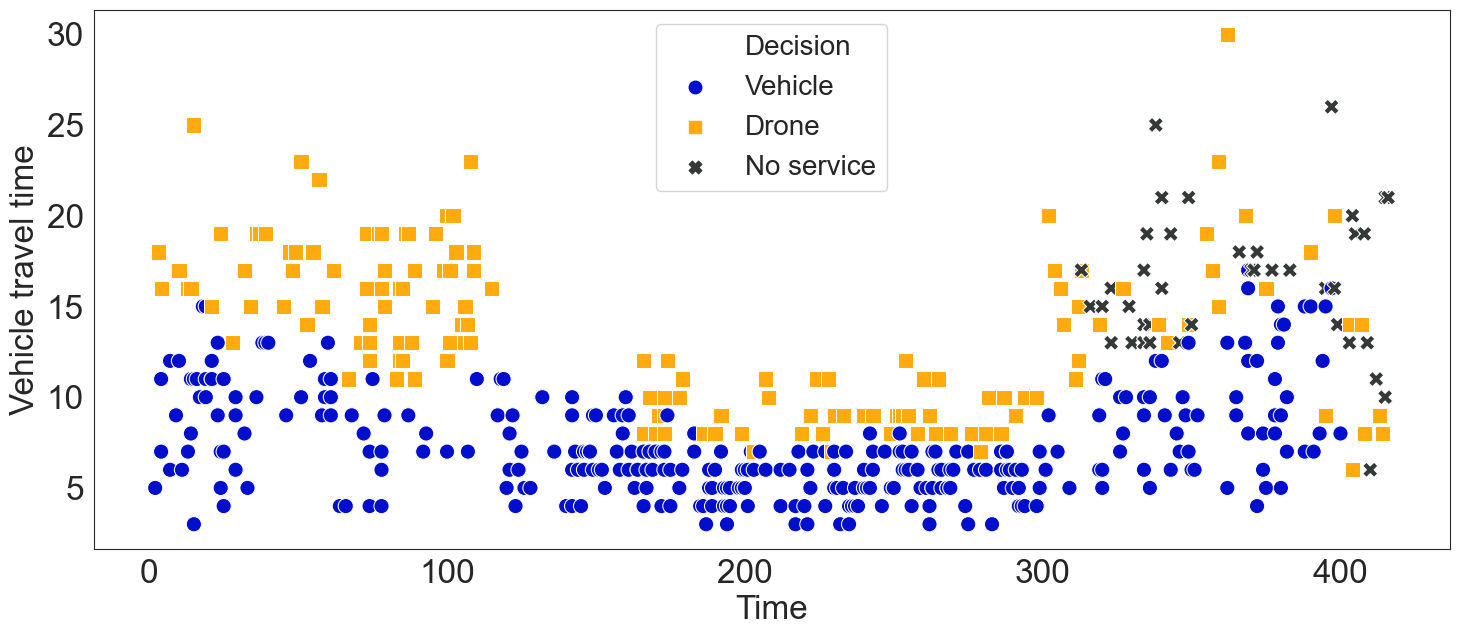}
	\caption{Time vs. travel time vs. decision under $\pi^{\textrm{Q}}$ on the selected spatially heterogeneous instance.}
	\label{H_dql}
\end{figure}

As for the decision making, we observe similar solution structures to those for homogeneous customer demand. Figures~\ref{H_pfa} through \ref{H_delta} present the decision making on the single instance for $\pi\textss{PFA}$, $\pi\textss{PFA\_rej}$ and $\pi\textss{Delta}$. Figure~\ref{H_dql} shows the results for the heterogeneous distribution and policy $\pi^{\textrm{Q}}$. 



\begin{figure}[h!]
	\centering
	\includegraphics[width=0.7\textwidth]{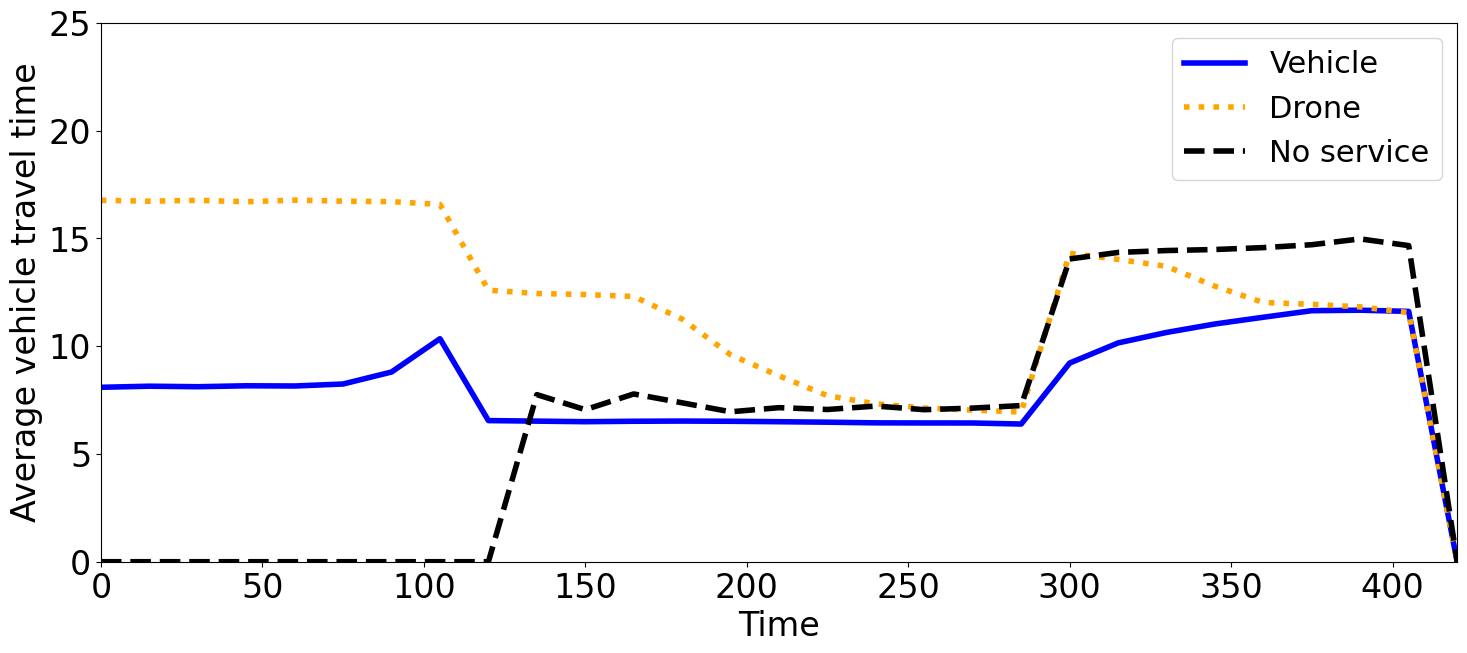}
	\caption{Average distance of requests vs. assignment decisions under $\pi\textsuperscript{PFA}$ for spatially heterogeneous demand.}
	\label{H_pfa_quantify}
\end{figure}

\begin{figure}[h!]
	\centering
	\includegraphics[width=0.7\textwidth]{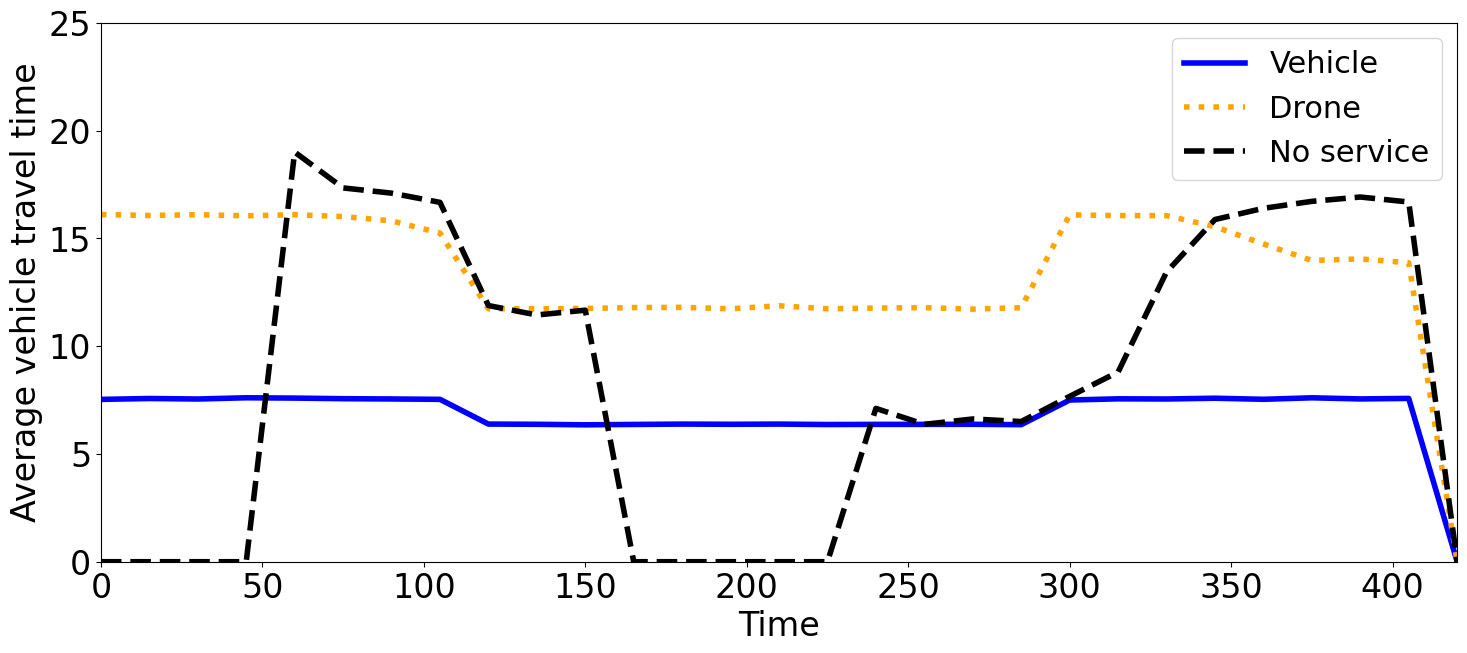}
	\caption{Average distance of requests vs. assignment decisions under $\pi\textsuperscript{PFA\_rej}$ for spatially heterogeneous demand.}
	\label{H_pfa_rej_quantify}
\end{figure}

\begin{figure}[h!]
	\centering
	\includegraphics[width=0.7\textwidth]{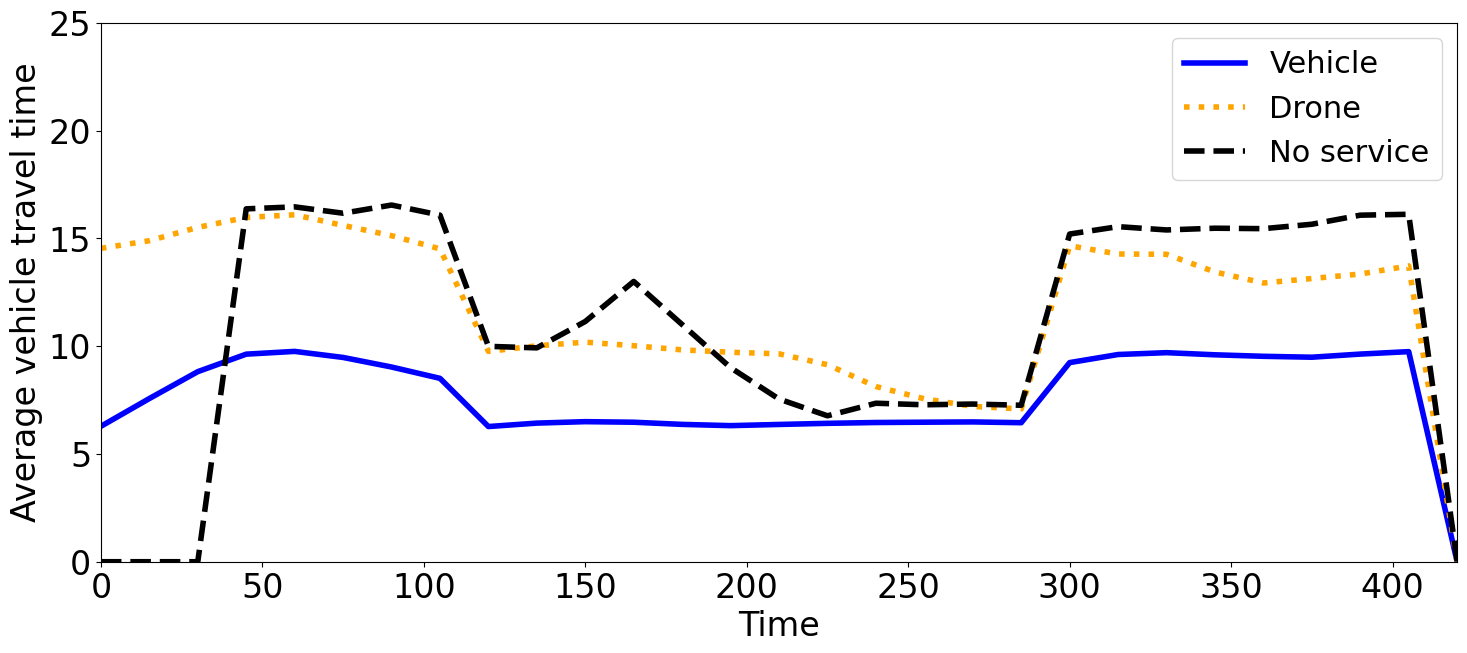}
	\caption{Average distance of requests vs. assignment decisions under $\pi\textsuperscript{Delta}$ for spatially heterogeneous demand.}
	\label{H_delta_quantify}
\end{figure}

\begin{figure}[h!]
	\centering
	\includegraphics[width=0.7\textwidth]{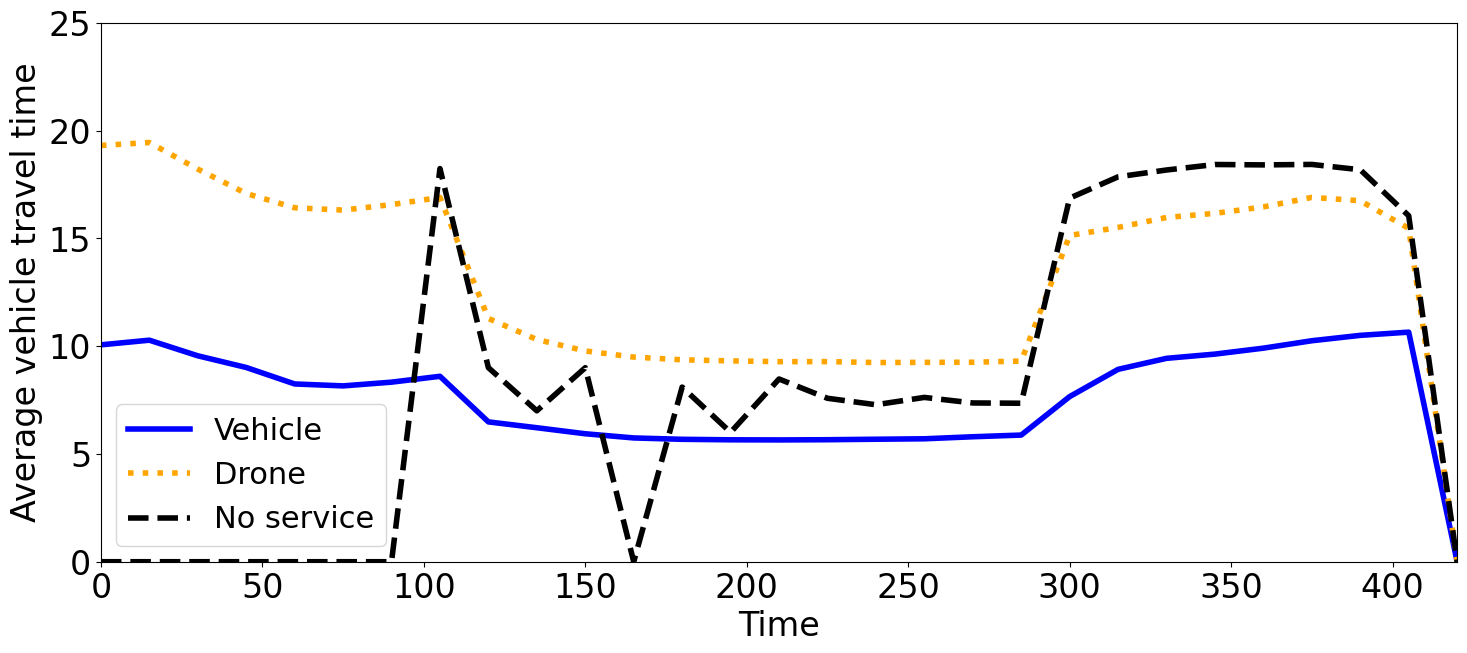}
	\caption{Average distance of requests vs. assignment decisions under $\pi\textsuperscript{Q}$ for spatially heterogeneous demand.}
	\label{H_dql_quantify}
\end{figure}

\clearpage

\subsection{Impact of Feature Selection}\label{appendix_feature_selection}

\begin{figure}[h]
	\begin{minipage}[b]{0.5\linewidth}
		\centering
		\includegraphics[width=.9\linewidth]{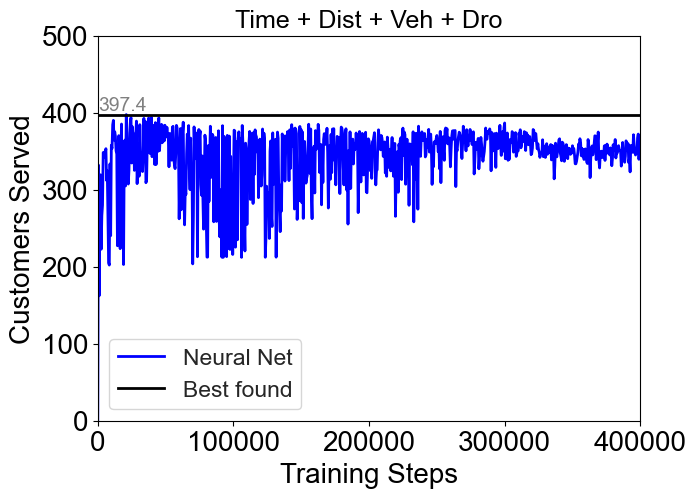} 
	\end{minipage}
	\begin{minipage}[b]{0.5\linewidth}
		\centering
		\includegraphics[width=.9\linewidth]{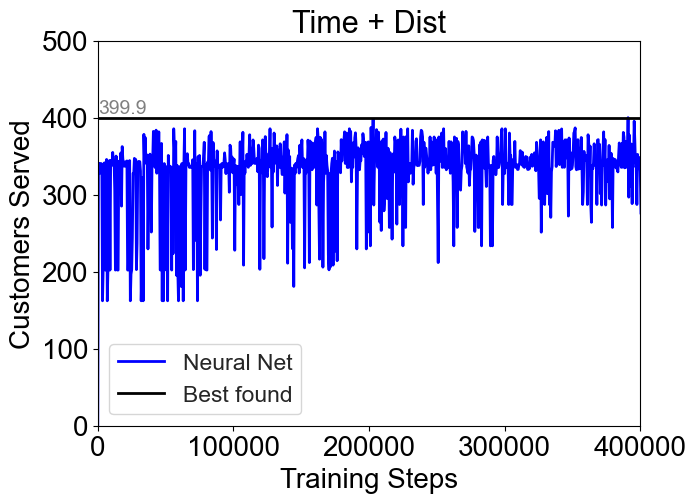} 
	\end{minipage}
	\caption{Solution quality curves with different features (3 vehicles 10 drones)}
	\label{3v10d_with_rej_A}
\end{figure}

In Figure~\ref{3v10d_with_rej_A}, we present two additional sets of features for our problem. 
On the left, the features time, distance to the customer, and availability times of the corresponding vehicle and drone are shown. On the right side, the set contains the time and the feature suggested in \cite{drones}, the distance to the customer. The results show that, the policies trained with these two sets of features still underperform our proposed policy after the same number of training epochs.

\subsection{The Value of Not Offering Service}\label{appendix_benchmark}

This section presents a computational comparison of a variant of $\pi$\textsuperscript{Q} that does not offer the opportunity to not offer service. Rather, the policy, denoted $\pi$\textsuperscript{Q\_no\_rej} uses logic of offering service similar to $\pi$\textsuperscript{PFA}. Table~\ref{table_pi_without} summarizes the logic.

\begin{table}[h!]
	\centering
	\begin{tabular}{ccc}
		\toprule   
		{Vehicle Feasibility} & {Drone Feasibility}  & {Accpt. and Assignment}\\
		\cmidrule{1-1} 
		\cmidrule{2-2} 
		\cmidrule{3-3} 
		$\times$ & $\times$                  & No service                      \\ 
		$\surd$ & $\times$                  & Vehicle                       \\ 
		$\times$& $\surd$                  & Drone                      \\
		$\surd$& $\surd$                  & $\phi(s)= \left\{\begin{array}{l}
		\textrm{Vehicle}\\
		\textrm{Drone}\end{array}\right. $\\
		\bottomrule
	\end{tabular}
	\caption{Logic of $\pi\textsuperscript{Q\_no\_rej}$ }
	\label{table_pi_without}
\end{table}

The policy $\pi\textsuperscript{Q\_no\_rej}$ requires only one NN. We use this NN to determine whether to serve a customer with vehicles or drones when both are feasible. We use the same features as those for $\pi$\textsuperscript{Q}. 

\begin{figure}[h]
	\centering
	\includegraphics[width=0.7\textwidth]{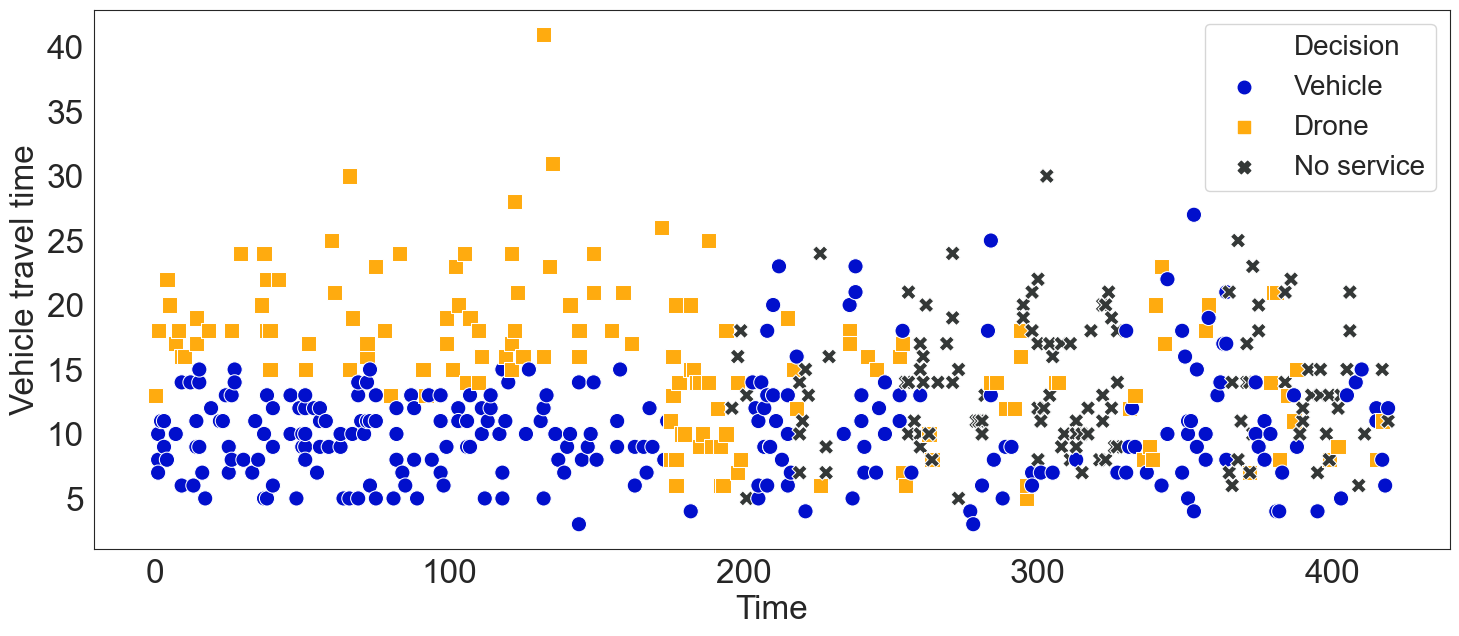}
	\caption{Time vs. travel time vs. decision under $\pi^{\textrm{Q\_no\_rej}}$ on the selected homogeneous instance,}
	\label{dql_without}.
\end{figure}

\begin{figure}[h]
	\centering
	\includegraphics[width=0.7\textwidth]{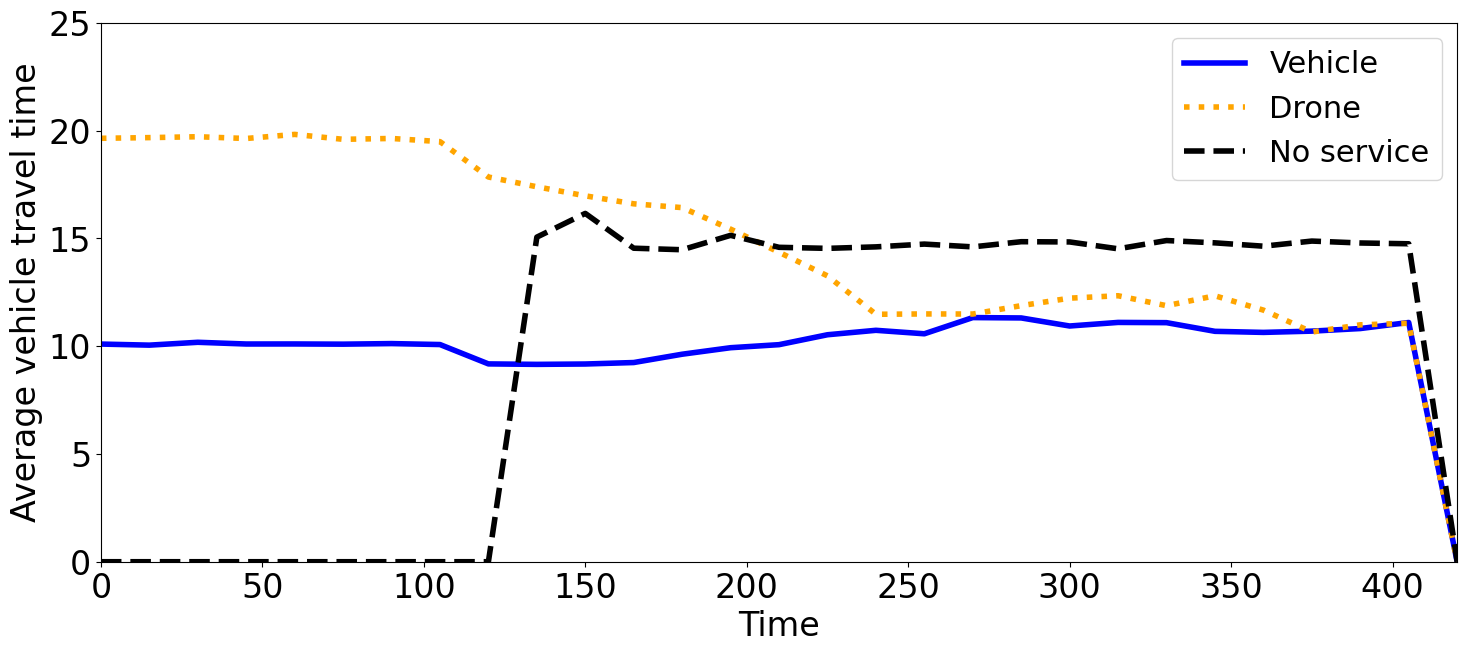}
	\caption{Average distance of requests vs. assignment decisions under $\pi^{\textrm{Q\_no\_rej}}$ for homogeneous demand.}
	\label{dql_no_rej_quantify}
\end{figure}

We take the homogeneous customer demand to illustrate the performance of $\pi\textsuperscript{Q\_no\_rej}$. Tables~\ref{app_normal_table} presents results analogous to those in Table~\ref{results} found in Section~\ref{s_q}. The results show that the policy $\pi\textsuperscript{Q\_no\_rej}$ is marginally better than the policy $\pi\textsuperscript{PFA}$. However, $\pi\textsuperscript{Q\_no\_rej}$ generally performs significantly worse than $\pi\textsuperscript{Q}$. The reason for this is illustrated in Figures~\ref{dql_without} and \ref{dql_no_rej_quantify}. These figures are analogous to Figures~\ref{dql} and Figure~\ref{dql_quantify}, respectively. 

\begin{table}[h!]
	\centering
	\small
	\begin{tabular}{ccccc}
		\toprule
		\multicolumn{1}{c}{\begin{tabular}[c]{@{}c@{}}Fleet Size\\
				(Veh, Drone)\end{tabular}} & \multicolumn{1}{c}{$\pi\textss{PFA}$} &  & \multicolumn{1}{c} {$\mathcal{P}(\pi\textss{Q\_no\_rej})$}  &
		\multicolumn{1}{c}{$\mathcal{P}(\pi\textss{Q})$} \\
		\cmidrule{1-5}
		2, 5                                                                                   & 227.6                      &                      &0.3 &  22.0\\ 
		2, 10                                                                                  & 312.8                      &                  & 0.1 & 10.7\\ 
		2, 15                                                                                  & 391.1                      &            &  0.2    & 6.8\\ 
		3, 5                                                                                   & 293.4                    &                        &  0.4 & 16.7\\ 
		3, 10                                                                                  & 376.2                      &                    &   0.4     &9.2\\ 
		3, 15                                                                                  & 460.3                      &                       &    0.6    &3.0\\ 
		4, 5                                                                                   & 354.7                     &                    &   0.2     & 11.0\\ 
		4, 10                                                                                  & 439.9                     &                       & 0.3      &3.6\\ 
		4, 15                                                                                  & 499.6                      &                       &     0.1     &0.0*\\ 
		\bottomrule
	\end{tabular}
	\caption{Improvements (\%) over $\pi\textsuperscript{PFA}$ (500 expected homogeneously distributed customers)}
	\label{app_normal_table}
\end{table}

\subsection{Expanding the Assignment Decisions}\label{sec:assignment}

For our main experiments, we use a cheapest inserting heuristic to determine the assignment, i.e., the vehicle the order is assigned to. This leads to a significant restriction of the action space. Such a restriction allows faster decision-making and learning, however, we may potentially ignore more valuable assignments. In the following, we analyze what happens if we learn the assignment decisions explicitly.

\subsubsection*{Extending Our Method}

We extend our method to allow explicit vehicle assignment decisions. Instead of learning the value of assigning to only the vehicle with the insertion that has the smallest $\Delta\textsubscript{Vehicle}$, the NNs learn the values for assignments to all the possible vehicles. Learning such assignments requires us to augment the set of features by incorporating $\Delta\textsubscript{Vehicle}$'s for all the vehicles. When the fleet of vehicles can feasibly serve the customer, we also augment the set of actions by having an output node for each of the $m$ vehicles. That is, we learn a state-action value for each vehicle. Note, different from learning individual values, the reward in this experiment is still calculated for the whole fleet.

\begin{figure}[h]
  \centering
    \includegraphics[width=0.45\textwidth]{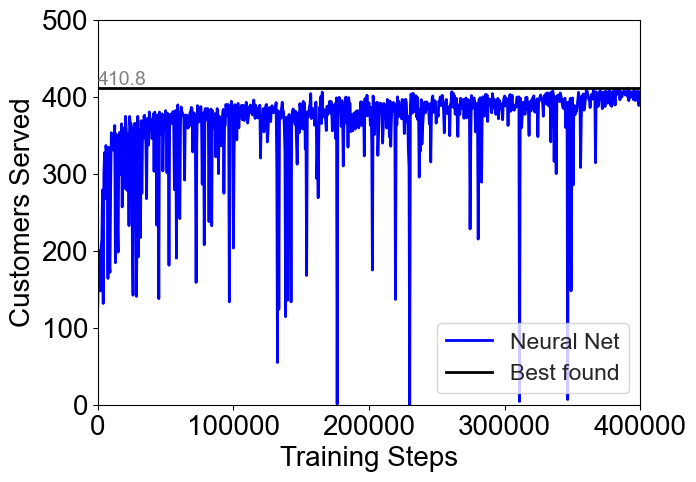}
    \caption{Extending our method: Learning assignments explicitly.}
    \label{fig:learn_routing}
\end{figure}

We train this extended method similar to $\pi\textsuperscript{Q}$. The learning process is shown in Figure \ref{fig:learn_routing}. We observe a slow and constant increase, but a large variation in the learning curve. One explanation is that, as the set of actions is augmented, the dispatcher has more choices when exploring. This leads to more variation and slower learning. As for the number of customers served on the test set, the extended method can eventually serve $409.8$ customers while $\pi\textsuperscript{Q}$ serves $410.7$ customers. However, the learning curve for the extended method indicates that the learning process has not yet finished after 400,000 training epochs. When running another 100,000 training epochs, we indeed observe the solution quality of the new policy further improves by $2.8\%$. This indicates that for our setup, the insertion heuristic performs well, but in cases where more training can be applied, our method should be extended to allow explicit assignment decisions.

\subsubsection*{Extending Method from Literature}

For a different dynamic routing problem, \citet{chen2019} implement a method whose features are not dependent on fleet information and can thus be applied to varying fleet sizes. Their work studies a courier dispatching problem by learning the probability of choosing an action on the individual level. In that work, the authors use one NN for all the individual couriers. The request is then assigned to the courier with smallest opportunity cost, the difference in expected number of customers the vehicle will serve in the future without and with the new customer. Thus, this policy is able to consider assignment decisions explicitly.

We adapt the work by \citet{chen2019} as follows. We use two NNs for the two fleets. Each NN represents the expected number of additional services a vehicle or drone can serve in a state. However, the NNs ignore the remaining setup of the fleet and only focus on the state information of individual vehicles and drones. Thus, the NN for the vehicles only uses the three features time $t_k$, the increase in route duration $\Delta\textsubscript{Vehicle}$, and the time the vehicle becomes available again $a(N_{2}^\theta)$. For the NN for the drones, the three features are time $t_k$, distance to the customer $d(C_k)$, and drone available time $a(N_{-1}^\theta)$.

Every time a customer makes a delivery request, the dispatcher checks the feasibility of serving it for all the vehicles and drones. Then, for each vehicle and drone that can feasibly serve the customer, the dispatcher use the corresponding NN to approximate the difference in the value of offering or not offering the service for the respective vehicle or drone, the opportunity cost. The customer is assigned to the vehicle or drone with the smallest opportunity cost or is not offered service if all opportunity cost are larger than one. 

\begin{figure}[h]
  \centering
    \includegraphics[width=0.45\textwidth]{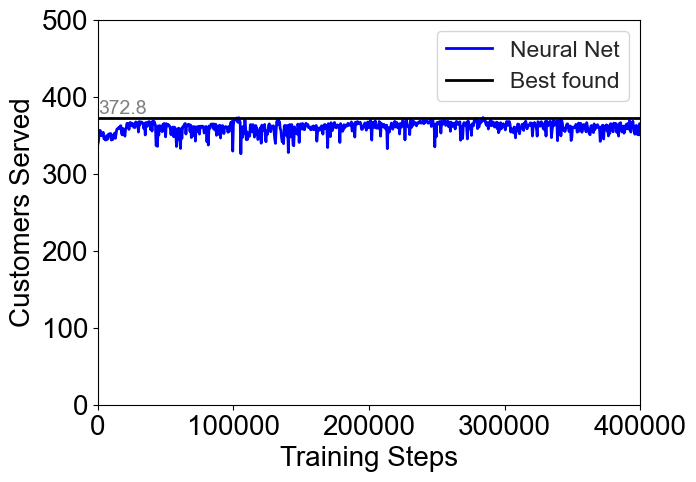}
    \caption{Learning values individually as suggested in \cite{chen2019}}
    \label{fig:learn_individual}
\end{figure}

In Figure \ref{fig:learn_individual}, we present the training curve for individual vehicles and drones. Although the curve shows small variance, the number of customers served barely improves in 400,000 steps. The best policy found can serve $372.8$ customers, while our policy $\pi^{\text{Q}}$ learning on fleet level serves $410.7$ customers. The result suggests that, for our problem, it is advantageous to learn the values on the fleet level than on the individual level. The reasons are twofold. First, the policy based on \citet{chen2019} ignores the fleet information and focuses only on individual vehicles and drones. Second, the problem at hand is substantially more complex compared to the problem addressed in \citet{chen2019}. Our problem is a pickup and delivery problem with many customers in the system and longer planned routes, while in \citet{chen2019}, no routes are considered and only the movements within a grid are planned.

\end{document}